\documentclass[acmtog]{acmart}
\acmSubmissionID{162}

\makeatletter
\DeclareRobustCommand\onedot{\futurelet\@let@token\@onedot}
\def\@onedot{\ifx\@let@token.\else.\null\fi\xspace}

\makeatother

\newcommand{\figref}[1]{Figure~\ref{fig:#1}}% type "\figref{}" to reference figure
\newcommand{\tabref}[1]{Table~\ref{tab:#1}} % type "\tabref{}" to reference table
\newcommand{\secref}[1]{Section~\ref{sec:#1}}

\newcommand{\red}[1]{{\color{red}#1}}

\newcommand{\ignore}[1]{}   % ignore this

\newcommand {\first}[1]{{\color{red}\textbf{#1}}}
\newcommand {\second}[1]{{\color{blue}\underline{#1}}}

\newlength\paramargin
\newlength\figmargin

\newlength\secmargin
\newlength\figcapmargin
\newlength\tabcapmargin

\newcommand{\mpage}[2]
{
\begin{minipage}{#1\linewidth}\centering
#2
\end{minipage}
}

\newcommand{\topic}[1]
{
\vspace{1mm}\noindent\textbf{#1}
}

\usepackage{enumitem}

\AtBeginDocument{%
  \providecommand\BibTeX{{%
    \normalfont B\kern-0.5em{\scshape i\kern-0.25em b}\kern-0.8em\TeX}}}

\copyrightyear{2024}
\acmYear{2024}
\usepackage{tikz}
\usepackage{stackengine}
\usepackage[percent]{overpic}
\usepackage{mathtools}
\usetikzlibrary{spy,backgrounds}
\pgfdeclarelayer{background}
\pgfdeclarelayer{foreground}
\pgfsetlayers{background,main,foreground}
\usepackage{float}
\begin{document}

\title{Planar Reflection-Aware Neural Radiance Fields}

% \affiliation{%

\author{Chen Gao}
\affiliation{%
  \institution{Meta}
  \city{Seattle}
  \state{Washington}
  \country{USA}
  }
\email{gaochen@meta.com}

\author{Yipeng Wang}
\affiliation{%
  \institution{Meta}
  \city{Seattle}
  \state{Washington}
  \country{USA}
  }
\email{yipeng.wang.99@outlook.com}

\author{Changil Kim}
\affiliation{%
  \institution{Meta}
  \city{Seattle}
  \state{Washington}
  \country{USA}
  }
\email{changil@meta.com}

\author{Jia-Bin Huang}
\affiliation{%
  \institution{University of Maryland}
  \city{College Park}
  \state{Maryland}
  \country{USA}
  }
\email{jbhuang@umd.edu}

\author{Johannes Kopf}
\affiliation{%
  \institution{Meta}
  \city{Seattle}
  \state{Washington}
  \country{USA}
  }
\email{jkopf@meta.com}


\begin{abstract}
Neural Radiance Fields (NeRF) have demonstrated exceptional capabilities in reconstructing complex scenes with high fidelity.
However, NeRF's view dependency can only handle low-frequency reflections. 
It falls short when dealing with complex planar reflections, often interpreting them as erroneous scene geometries and leading to duplicated and inaccurate scene representations.
To address this challenge, we introduce a planar reflection-aware NeRF that jointly models planar reflectors, such as windows, and explicitly casts reflected rays to capture the source of the high-frequency reflections.
We query a single radiance field to render the primary color and the source of the reflection.
We propose a sparse edge regularization to help utilize the true source of reflections for rendering planar reflections rather than creating a duplicate along the primary ray at the same depth.
As a result, we obtain accurate scene geometry.
Rendering along the primary ray results in a clean, reflection-free view, while explicitly rendering along the reflected ray allows us to reconstruct highly detailed reflections.
Our extensive quantitative and qualitative evaluations of real-world datasets demonstrate our method's performance in accurately handling reflections.
\end{abstract} 

%    <concept>
%        <concept_id>10010147.10010371.10010382.10010384</concept_id>
%        <concept_desc>Computing methodologies~Texturing</concept_desc>
%        <concept_significance>500</concept_significance>
%        </concept>
%  </ccs2012>



% \keywords{Digital humans, single-image 3D reconstruction, diffusion models}

\begin{teaserfigure}
\newlength\figwidthTeaser
\setlength\figwidthTeaser{0.198\linewidth}
% \begin{figure}[t]
%
\centering%
\parbox[t]{\figwidthTeaser}{\centering%
  \includegraphics[trim=0 0 0 0, clip=true, width=\figwidthTeaser]{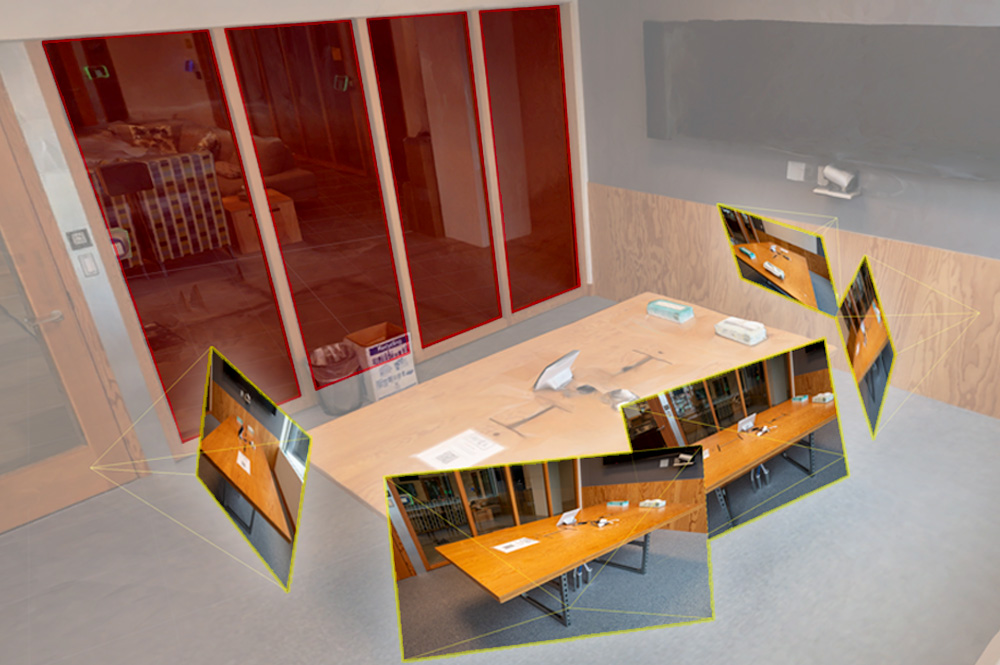}\\%
   \small Input}%
\hfill%
\parbox[t]{\figwidthTeaser}{\centering%
  \includegraphics[trim=300 200 0 0, clip=true, width=\figwidthTeaser]{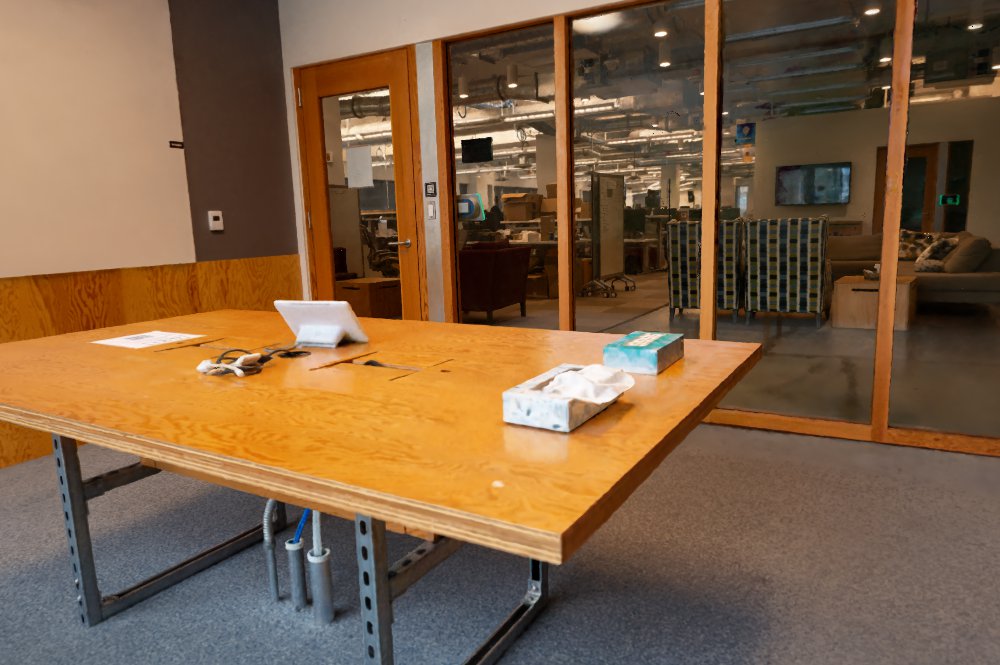}\\%
   \small Reflection-free rendering}%
\hfill
\parbox[t]{\figwidthTeaser}{\centering%
  \includegraphics[trim=300 200 0 0, clip=true, width=\figwidthTeaser]{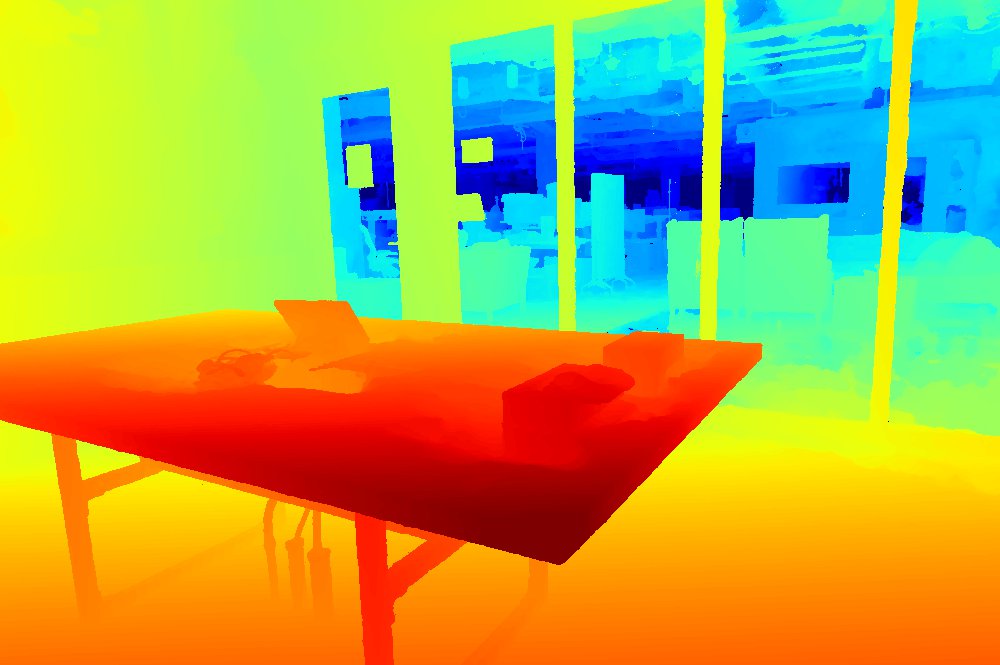}\\%
   \small Reflection-free depth}%
\hfill%
\parbox[t]{\figwidthTeaser}{\centering%
  \includegraphics[trim=300 200 0 0, clip=true, width=\figwidthTeaser]{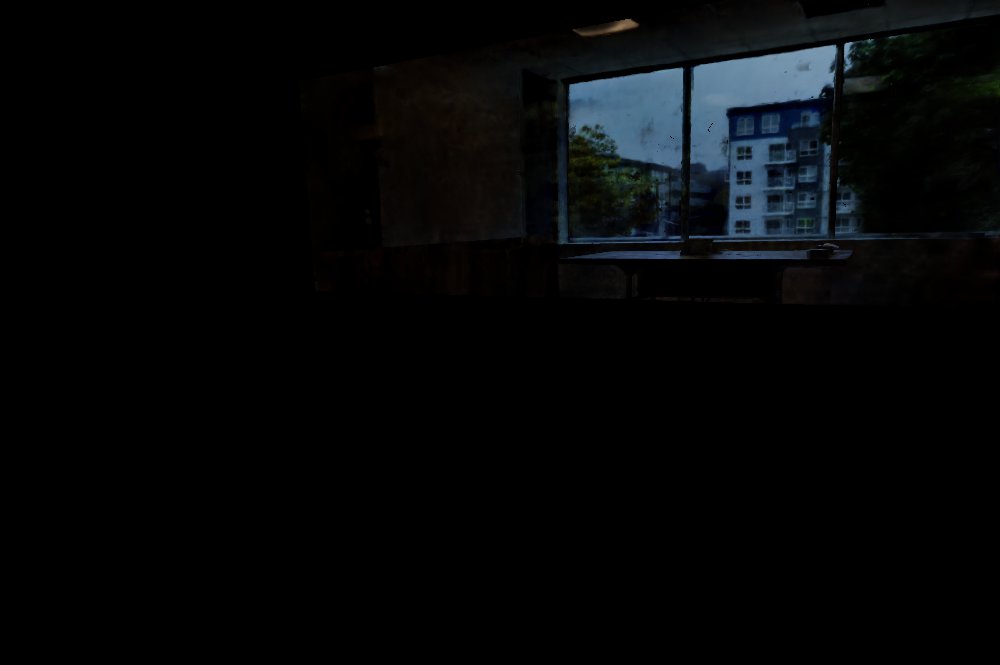}\\%
   \small Reflection}%
\hfill%
\parbox[t]{\figwidthTeaser}{\centering%
  \includegraphics[trim=300 200 0 0, clip=true, width=\figwidthTeaser]{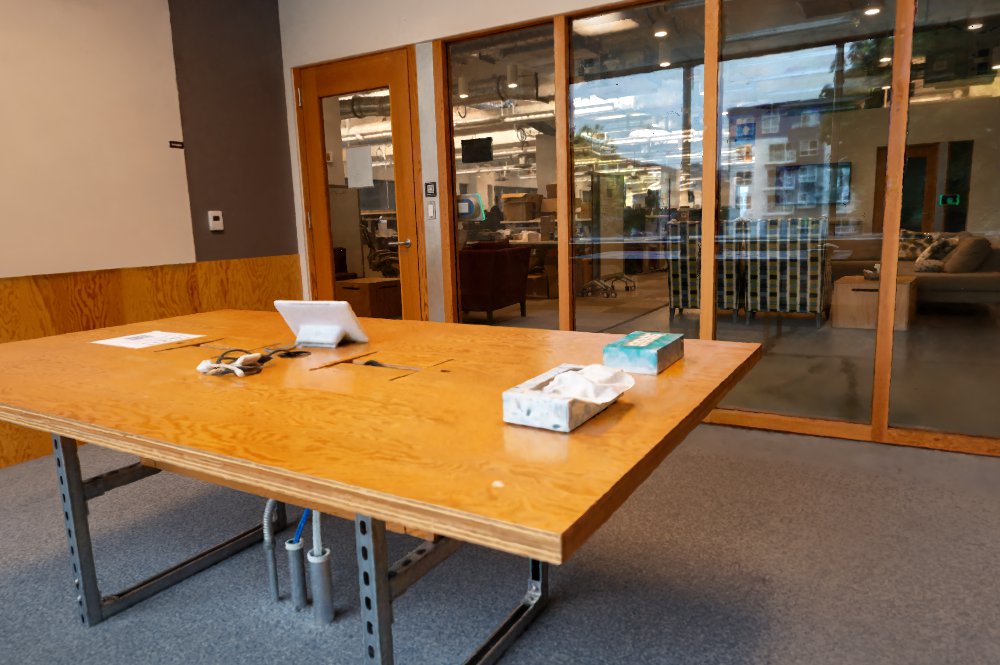}\\%
   \small Composed rendering}%
\hfill%
\caption{
\textbf{Planar reflection-aware novel view synthesis.}
Our method takes multi-view capture data as input to jointly model planar reflectors and reconstruct radiance fields free of false geometries caused by reflections.
This leads to accurate scene geometry, which benefits downstream applications, such as mesh extraction.
Rendering along the primary ray results in a clean, reflection-free view, while explicitly rendering along the reflected ray allows us to reconstruct highly detailed reflections.
By composing these two renderings, we achieve sharp and realistic novel view synthesis.
% \chen{Update the input image.}
}
\label{fig:teaser}
% \end{figure}
\end{teaserfigure}

\maketitle
\begin{figure*}[t]
\centering
\includegraphics[width=1.0\textwidth]{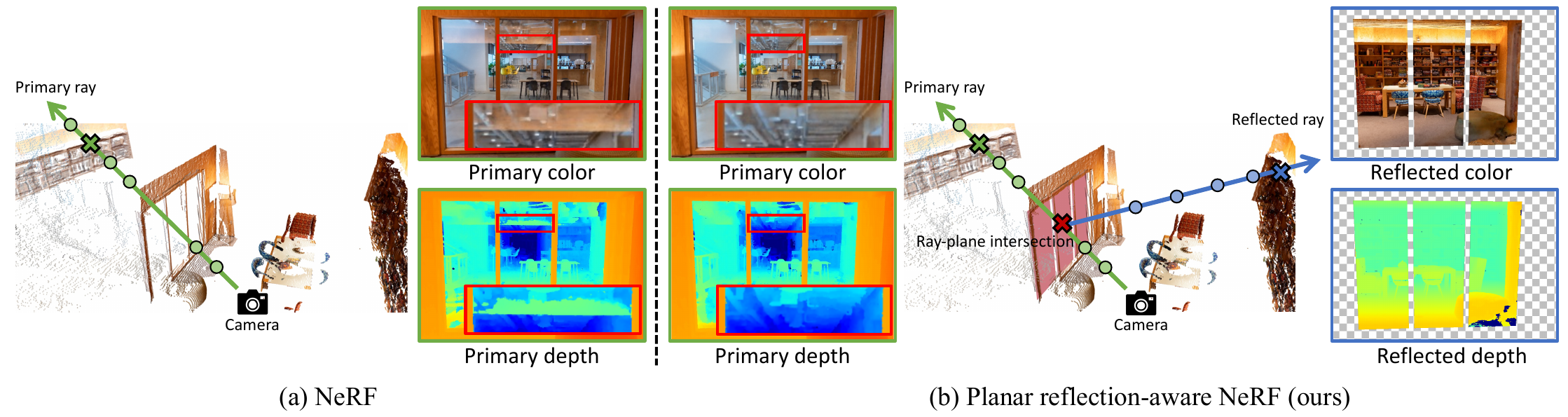}
\captionof{figure}{\textbf{Why NeRF generates false geometry?}
(a) NeRF only traces a single ray, thus incorrectly interpreting reflections as false geometry, leading to inaccurate scene representations.
(b) Our planar reflection-aware NeRF addresses this by employing the law of reflection during rendering, which involves explicitly casting a reflected ray to capture the \emph{source} of the reflections. 
Our method prevents the creation of false geometries along the primary ray at depths corresponding to real objects on the reflected ray.
}
\label{fig:motivation}
\end{figure*}

\section{Introduction}
\label{sec:introduction}
% Reflection is a common phenomenon.
Planar reflections are a ubiquitous and visually significant phenomenon in our everyday environments.
Windows, in particular, are common planar reflectors, making their proper handling a critical aspect of realistic scene reconstruction.
In this work, we focus on accurately modeling these planar reflections during scene reconstruction.

% Problem of existing methods
% - NeRF: only handle low-frequency reflections, does not work for high-frequency planar reflections.
% - NeRFren: decomposes the scene into two radiance fields --> bad decomposation and false geometries caused by reflections persist
% - MS-NeRF: decomposes the scene into multiple feature fields --> bad decomposation and false geometries caused by reflections persist
% - RefNeRF: can better handle reflection and could not separate reflections
Neural Radiance Fields (NeRF) \cite{mildenhall2020nerf} have made groundbreaking strides in rendering complex scenes with remarkable fidelity.
Recent advancements have further improved NeRF's capabilities, such as anti-aliasing \cite{barron2021mip, barron2022mip}, handling transient elements \cite{martin2021nerf, liu2023robust}, and addressing inconsistent camera exposure or illumination \cite{rematas2022urf, mildenhall2020nerf}.
However, a notable limitation of NeRF lies in its handling of reflections.
As illustrated in \figref{motivation}(a), NeRF traces only a single ray and relies on view-dependency or false geometry to account for reflections.
For low-frequency reflections like highlights, this is achieved by inputting 3D point positions and viewing directions into a small MLP, which predicts view-dependent emitted radiance.
However, this method falls short when dealing with complex, high-frequency reflections.
NeRF tends to create false geometry along the primary ray, incorrectly mirroring the actual object’s position along the reflected ray to represent planar reflections (\figref{motivation}a).
While this offers a plausible explanation for reflections, it fundamentally misinterprets them. 
NeRF cannot distinguish a reflected object from a duplicated one, as both appear visually similar.
This misrepresentation prevents us from creating correct novel views if we see the space \emph{behind} the reflector, not \emph{through} the reflector. It also leads to incorrect scene geometry, adversely affecting downstream applications, such as mesh extraction.

Several methods have been proposed to address reflections in NeRF \cite{kirillov2023segment,verbin2024nerf,liu2023nero,Guo_2022_CVPR,Yin_2023_CVPR,verbin2022refnerf}.
NeRFReN \cite{Guo_2022_CVPR} proposes to model the transmitted and reflected parts of a forward-facing scene with separate neural radiance fields.
MS-NeRF \cite{Yin_2023_CVPR} represents the scene using multiple feature fields decoded by small MLPs for rendering and blending.
These methods decompose the scene into several \emph{independent} components in a self-supervised manner, sometimes leading to sub-optimal decomposition where false geometries persist in the primary component.
Ref-NeRF \cite{verbin2022refnerf} addresses specular reflections by explicitly parameterizing outgoing radiance as a function of the reflection of the view direction about the local normal vector.
It shows promising results in representing specular reflections but lacks the ability to separate reflections.

% Core idea: \emph{single} radiance field --> a \emph{single} instance of geometry must be maintained. No duplication. If a real object is presented along the reflected ray, the model should not create an illusory duplicate along the primary ray trying to explain the reflection. It should always use the reflected ray (SOURCE of reflection) to explain the reflection. As a results, the primary is clean and accurate.
In this paper, we introduce a planar reflection-aware NeRF that jointly models planar surfaces and explicitly casts reflected rays to capture the \emph{source} of high-frequency reflections (\figref{motivation} b).
Our core idea is to maintain a \emph{single} geometry and appearance instance.
This principle dictates that if a real object is present along the reflected ray, our method should not create a false duplicate along the primary ray at the same depth to explain the reflection.
Instead, it should always use the reflected ray to explain the reflection, ensuring the primary view is clean and accurate.
To render novel views, we query a \emph{single} radiance field to render the primary color and the source of the reflection.
Then, we learn to attenuate the source of the reflection and compose it with the primary color to obtain the final color.
We introduce regularization to help utilize the true source of reflections for rendering planar reflections rather than creating a false duplicate.
As a result, we obtain accurate scene geometry.
Rendering along the primary ray results in a clean, reflection-free view, while explicitly rendering along the reflected ray allows us to reconstruct highly detailed reflections.
We validate our method's reconstruction performance on our real-world 360-degree dataset and demonstrate significant enhancements in NeRF's performance in handling reflections.
 
Our contributions are summarized as follows:
\begin{itemize}
\item We introduce a method for handling planar reflections by jointly modeling planar surfaces and casting reflected rays to accurately capture the source of high-frequency reflections.
\item We develop a regularization loss to discourage the creation of false geometries.
\item We provide a real-world 360-degree dataset that contains pronounced reflections, facilitating the quantitative evaluation of reflection-free reconstruction.
% Our model demonstrates favorable results compared to state-of-the-art algorithms on the dataset \johannes{The last sentence seems to belong to the first bullet about the method, rather than this bullet about the dataset?}.
\end{itemize}
\section{Related Work}
\label{sec:related_work}

\begin{figure*}[t]
\centering
\begin{overpic}[width=1\linewidth]{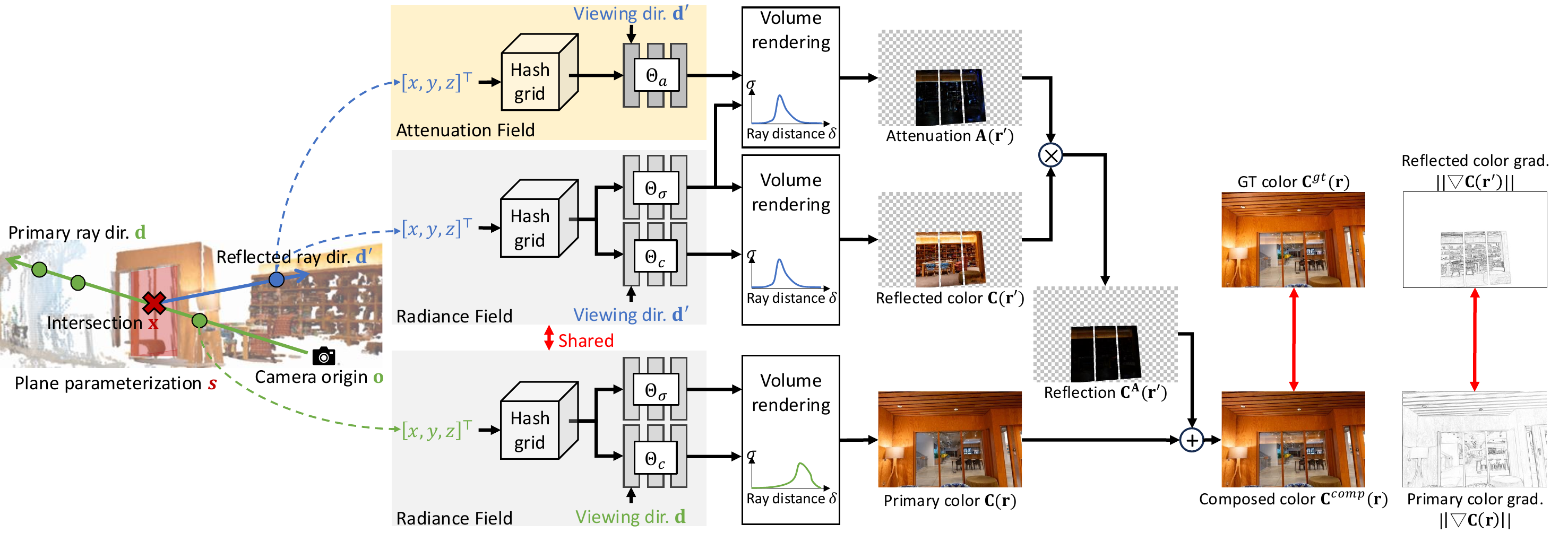}
\put (83, 12.5) {\scriptsize \red{$\mathcal{L}_\text{photo}$}}
\put (83, 11) {\scriptsize Eq. (\ref{eq:photometric_loss})}
\put (94.6, 12.5) {\scriptsize \red{$\mathcal{L}_\text{edge}$}}
\put (94.6, 11) {\scriptsize Eq. (\ref{eq:edge_loss})}
\end{overpic}
\caption{
\textbf{Overall framework.}
Our method takes a multi-view capture as input.
We first detect potential planar reflectors (e.g., window planes).
Then for each camera ray, denoted by $(\mathbf{o}, \mathbf{d})$ where $\mathbf{o}$ is the origin and $\mathbf{d}$ the viewing direction, the 3D coordinates of sampled points and their viewing directions serve as inputs to predict volumetric density $\sigma$ and color $\mathbf{c}$.
% \changil{A bit too detailed to my taste. Most people will know how NeRF rendering works.}
Upon intersecting with a plane, a reflected ray $\mathbf{r}'$ emanates from the point of intersection $\mathbf{x}$, with its new direction $\mathbf{d}'$ computed as per Eq. (\ref{eq:reflected_viewing_direction}). This reflected ray probes the same radiance field to obtain the density and color.
% Additionally, the transmittance $\mathrm{T}$ at the intersection is acquired via Eq. (\ref{eq:transmittance}). We omit $\mathrm{T}$ in this figure for simplicity.
% \changil{I'd be careful and rather not omit anything that's important for the sake of simplicity.}
% \chen{Right. But I couldn't find a good place to insert the transmittance map.}
Through volume rendering, colors $\mathbf{C}(\mathbf{r})$ and $\mathbf{C}(\mathbf{r}')$ are computed for the primary and reflected rays, respectively.
% We learn an attenuation field to model the Fresnel law and HDR tonemapping, and absorbs errors due to intensity clipping.
% The attenuation $\mathrm{A}$, predicted by the attenuation field, integrates with $\mathrm{T}$ to form the composited color: $\mathbf{C}^\textit{comp}(\mathbf{r}) = \mathbf{C}(\mathbf{r}) + \mathrm{T} \cdot \mathrm{A} \cdot \mathbf{C}(\mathbf{r}')$.
To model the Fresnel effect, account for HDR tonemapping, and absorb errors due to intensity clipping, we introduce an attenuation field to adjust the reflected color.
We add the reflection $\mathbf{C^A}(\mathbf{r}')$ to the primary color $\mathbf{C}(\mathbf{r})$ to yield the composited color.
% $\mathbf{C}^\textit{comp}(\mathbf{r}) = \mathbf{C}(\mathbf{r}) + \mathrm{T} \cdot \mathrm{A} \cdot \mathbf{C}(\mathbf{r}')$.
The reconstruction loss is calculated between the composed color $\mathbf{C}^\textit{comp}$ and the ground truth $\mathbf{C}^\textit{gt}$.
We also apply the sparse edge regularization between the primary color gradient and the reflected color gradient.
The loss is backpropagated to update the radiance field, the attenuation field, and the plane parameters.
% \changil{Can you give an intuitive explanation about what the attenuation field models? It's an important piece in our model -- it models the frenel law (we don't use the physics formula here, do we?) and HDR tonemapping, and absorbs errors due to intensity clipping, etc. In the ``volume rendering'' box of the attenuation field, should $\sigma$ on the y-axis be $\alpha$?}
}
\label{fig:overview}
\end{figure*}
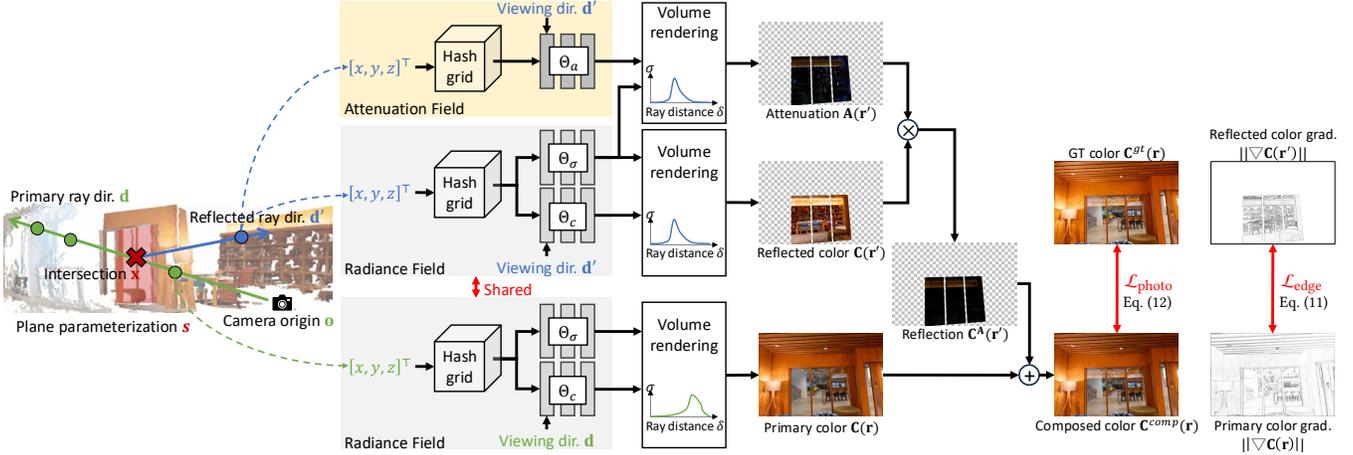

\topic{Novel view synthesis.} 
Novel view synthesis aims to generate new views from a set of posed images \cite{shum2000review, chaurasia2013depth}.
Techniques such as light fields \cite{levoy1996light} and Lumigraph \cite{gortler1996lumigraph} leverage implicit scene geometry to synthesize realistic appearances, although they require densely captured images.
High-quality synthesis with fewer input images is achievable using explicit geometric proxies \cite{buehler2001unstructured}, yet accurately estimating scene geometry remains challenging due to issues like highlights and reflections.
Prior work has addressed these challenges through learning-based dense depth maps \cite{flynn2016deepstereo}, MPIs \cite{flynn2019deepview, choi2019extreme, srinivasan2019pushing}, the incorporation of learned deep features \cite{hedman2018deep, riegler2020free}, or voxel-based implicit scene representations \cite{sitzmann2019scene}, utilizing proxy scene geometry for enhanced rendering.
Recently, neural implicit representations, particularly NeRF \cite{mildenhall2020nerf}, have shown promising results in this domain.
However, achieving high-quality, artifact-free rendering remains a substantial challenge.
Advancements in NeRF technology, such as anti-aliasing \cite{barron2021mip, barron2022mip, barron2023zip}, handling of transient elements \cite{martin2021nerf, li2021neural, gao2021dynamic, liu2023robust}, adjustments for inconsistent camera exposure or illumination \cite{rematas2022urf, mildenhall2020nerf}, and improvements in training efficiency \cite{fridovich2022plenoxels, muller2022instant, sun2022direct, chen2022tensorf}, have further refined the quality of rendered outputs.
While these methods yield high-quality results, a notable limitation lies in their handling of reflections.
NeRF is effective for low-frequency reflections but struggles with more complex, high-frequency planar reflections.
In this work, we focus on handling planar reflections to facilitate reflection-free novel view synthesis.

\topic{Reflection modeling.}
In computer vision and graphics, reflections are a significant challenge.
Traditional graphics techniques utilize ray tracing to simulate reflections with high accuracy but at a high computational cost.
More recent works have explored incorporating reflection models into neural rendering pipelines, allowing for more efficient and realistic rendering of reflective surfaces.
Ref-NeRF \cite{verbin2022refnerf} addresses specular reflections by explicitly parameterizing outgoing radiance as a function of the reflection of the view direction about the local normal vector. It shows promising results in representing specular reflections but cannot separate reflections.
TraM-NeRF \cite{van2023tram} models mirror-like reflection behavior but overlooks transmission and does not focus on separating reflections.
Mirror-NeRF \cite{zeng2023mirror-nerf} has a similar reflected ray tracing idea to ours but focuses on mirrors. It does not consider transmission and thus is not able to handle transparent reflectors.

\topic{Scene decomposition.}
Recent efforts have attempted to extend NeRF to handle reflections by introducing additional network outputs.
NeRFReN \cite{Guo_2022_CVPR} proposes modeling the transmitted and reflected parts of the forward-facing scene with separate neural radiance fields. The final image is obtained by composing the results from these two radiance fields.
MS-NeRF \cite{Yin_2023_CVPR} represents the scene using multiple feature fields decoded by small MLPs for rendering and blending.
These methods decompose the scene into independent components in a self-supervised manner, sometimes leading to suboptimal decomposition where false geometries caused by reflections persist.
In contrast, our reflection-aware NeRF employs a \emph{single} radiance field while explicitly casting reflected rays to capture the source of reflections.
It adheres to the principle that if an actual object is observed along the reflected ray, no false duplicate should be created along the primary ray at the same depth.
Instead, the reflected ray explains the reflection, ensuring a clean and accurate primary view.

\section{Method}
% \vspace{-2.5mm}
\label{sec:method}
This section first recapitulates neural radiance fields in~\secref{preliminaries}. 
We then provide an overview of our method in~\secref{overview}. 
Subsequently, we delve into the details of our reflection-aware NeRF in~\secref{ranerf}. 
We discuss the sparse edge regularization and training strategies in~\secref{training}.
Finally, we outline the implementation details in~\secref{details}.

\subsection{Neural Radiance Fields}
\label{sec:preliminaries}
Neural Radiance Fields (NeRF)~\cite{mildenhall2020nerf} employ MLPs parameterized by $\Theta$ to map the 3D position $\mathbf{x} = (x, y, z)$ and the normalized viewing direction $\mathbf{d} = (d_x, d_y, d_z) \in \mathbb{R}^3$ to the corresponding volume density $\sigma$ and color $\mathbf{c}$.
\begin{equation}
(\sigma, \mathbf{c}) = \text{MLP}_{\Theta}(\mathbf{x}, \mathbf{d}). 
\label{eq:nerf}
\end{equation}
The color of a pixel can be computed by performing volume rendering along the ray $\mathbf{r}$, emitted from the camera origin~\cite{drebin1988volume}:
\begin{align}
\mathbf{C}(\mathbf{r}) &= \sum_{k=1}^{K} T_k \left( 1-e^{-\sigma_k\delta_k} \right) \mathbf{c}_k, \label{eq:rendering}\\
T_k &= e^{-\sum_{{k'}=1}^{k-1}\sigma_{k'}\delta_{k'}} \label{eq:transmittance},
\end{align}
where $\delta(k)$ is the distance between two consecutive points along the ray, $K$ denotes the number of samples along each ray, and $T_k$ indicates the accumulated transmittance.

To optimize the weights $\Theta$ of the NeRF model, we first construct the camera ray $\mathbf{r} = (\mathbf{o}, \mathbf{d})$, where $\mathbf{o}$ is the camera origin and $\mathbf{d}$ is the viewing direction. We can then optimize $\Theta_s$ by minimizing the photometric loss between the rendered color $\mathbf{C}$ and the ground truth color $\mathbf{C}^\textit{gt}$:
\begin{equation}
\mathcal{L} = \left \| \mathbf{C}(\mathbf{r}) - \mathbf{C}^\textit{gt}(\mathbf{r}) \right \|_{2}^{2}.
\end{equation}

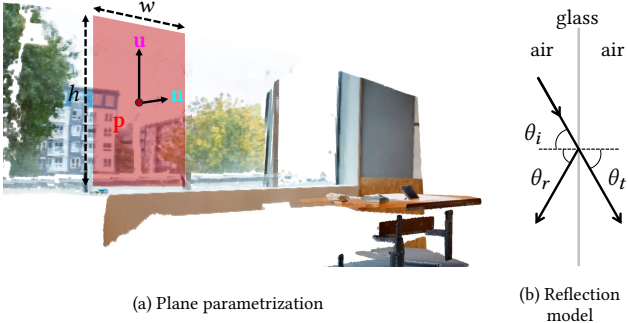
\begin{figure}[t]
\begin{overpic}[width=1\linewidth]{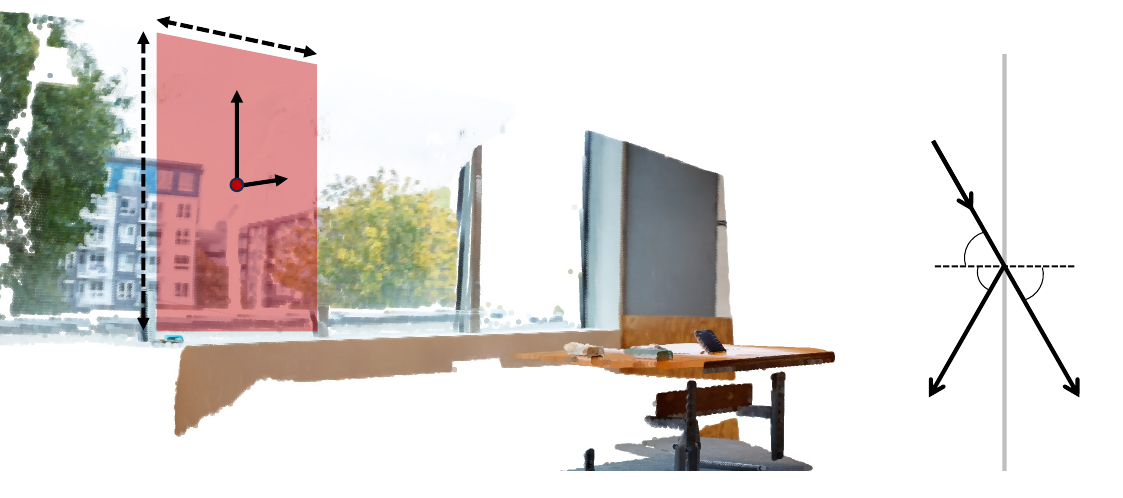}
% \begin{overpic}[width=1\linewidth,grid,tics=5]{figures/plane_parameterization.pdf}
\put (17, 24) {\textcolor[HTML]{FF0000}{$\mathbf{p}$}}
\put (20.2, 36.5) {\textcolor[HTML]{FF00FF}{$\mathbf{u}$}}
\put (26, 28) {\textcolor[HTML]{00FFFF}{$\mathbf{n}$}}

\put (21, 42) {$w$}
\put (10, 28) {$h$}

\put (81, 22) {$\theta_i$}
\put (82, 15) {$\theta_r$}
\put (94, 15) {$\theta_t$}

\put (82, 35) {\small air}
\put (93, 35) {\small air}
\put (86, 40) {\small glass}

\end{overpic}
\mpage{0.7}{\footnotesize (a) Plane parametrization}
\hfill
\mpage{0.2}{\footnotesize (b) Reflection model}
\vspace{\figcapmargin}
\caption{
\textbf{Reflection model.} (a) Our plane is parameterized by 
a center \textcolor[HTML]{FF0000}{$\mathbf{p}$}, 
a normal \textcolor[HTML]{00FFFF}{$\mathbf{n}$},
an up vector \textcolor[HTML]{FF00FF}{$\mathbf{u}$},
and width and height $(w, h)$.
(b) We adopt a simple reflection model to conceptualize the reflection process as involving solely a primary incident ray and its consequent reflected ray, effectively neglecting any refraction effects that would arise from a finite glass thickness. 
}
\label{fig:plane_parameterization}
\end{figure}

\subsection{Method Overview}
\label{sec:overview}
We show our proposed framework in~\figref{overview}. 
We tackle the challenging problem of accurately modeling planar reflections. 
NeRF only traces a single ray, thus tends to create false geometry along the primary ray, incorrectly mirroring the actual object’s position along the reflected ray to represent planar reflections. \figref{motivation} (a) illustrates this common pitfall with the reflection of a shelf erroneously modeled as a solid object. 
Thus, in this paper, we introduce a reflection-aware NeRF that jointly models planar surfaces and explicitly casts reflected rays to capture the \emph{source} of the high-frequency reflections (\figref{motivation} b). 
Given a multi-view capture of a scene as input, we aim to improve the rendering quality in the reflection regions and obtain accurate geometry.

\subsection{Plane Annotation and Parameterization}
\label{sec:plane}
\begin{figure}[t]
\newlength\figwidthRefine
\setlength\figwidthRefine{0.494\linewidth}
\parbox[t]{\figwidthRefine}{\centering%
  \fbox{\includegraphics[trim=482 350 0 0, clip=true, width=\figwidthRefine]{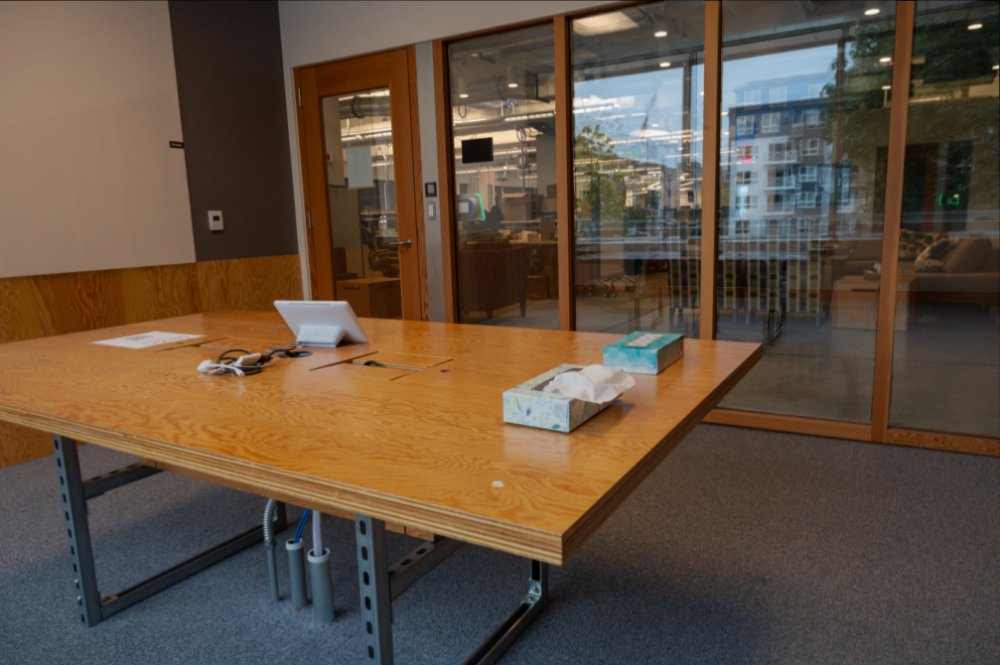}}\\
  \small (a) Before plane refinement}%
\hfill
\parbox[t]{\figwidthRefine}{\centering%
  \fbox{\includegraphics[trim=482 350 0 0, clip=true, width=\figwidthRefine]{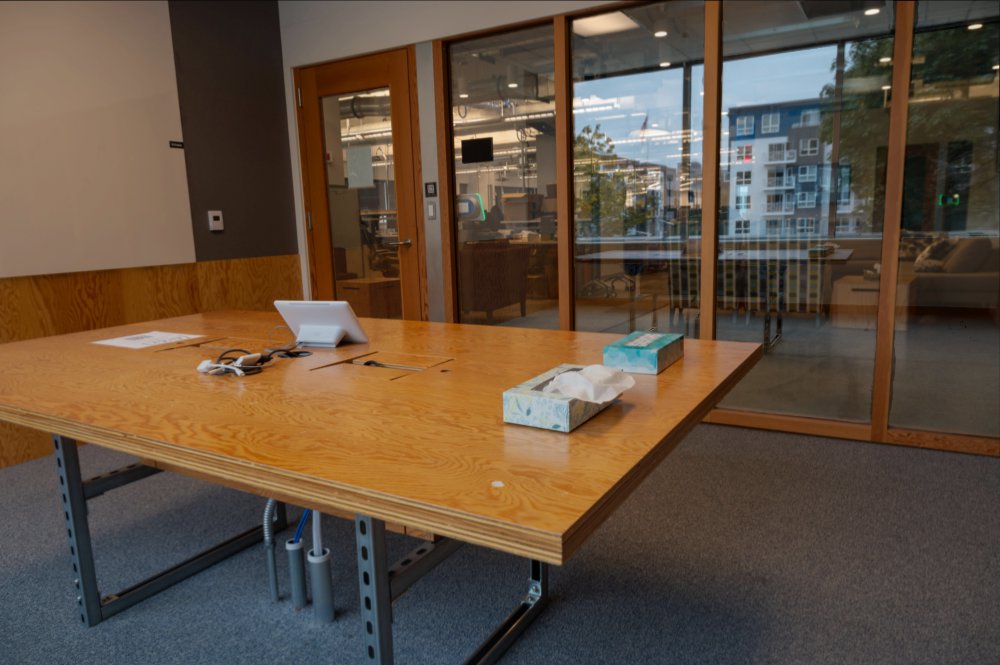}}\\
  \small (b) After plane refinement}%
% left, bottom, right and top
\vspace{\figcapmargin}
\caption{\textbf{Plane refinement.}
With jointly optimization, our rendered reflection aligns with the observations.
}
\label{fig:plane_refinement}
\end{figure}

% We propose a simple method to eliminate the need for manual plane annotation.
We propose a simple method to detect and annotate planes.
We first train an InstantNGP and render the depth maps.
We use Segment Anything (SAM) \cite{kirillov2023segment} to segment the windows in 2D and unproject the four dilated corners into 3D space using the depth map.
A least square fit to these points gives us a plane parameter initialization.
Following NeurMiPs \cite{lin2022neurmips}, a plane segment is parameterized by $\mathbf{s} = (\mathbf{p}, \mathbf{n}, \mathbf{u}, w, h)$, where $\mathbf{p}$ is the plane center; $\mathbf{n}$ is the plane normal; $\mathbf{u}$ is the normalized up vector defining the y-axis direction of the plane; and $(w, h)$ are the dimensions of the plane.
See \figref{plane_parameterization} (a) for an illustration of the plane parameterization.
Our model does not require accurate annotation; these coarsely annotated planes serve as initialization, and the plane parameterization is jointly refined during training (\figref{plane_refinement}).
% As shown in \figref{decomposition} first row, SAM groups all windows together, and our method can still refine the plane segment accurately.

\subsection{Reflection-aware NeRF}
\label{sec:ranerf}

% \topic{Primer on Snell's Law.}
% Snell's Law, also known as the law of refraction, describes the relationship between the angles of incidence and refraction for light or other waves passing through the boundary between two different isotropic media, such as air and glass. It is given by:
% \begin{equation}
% n_1 \sin(\theta_i) = n_2 \sin(\theta_t),
% \end{equation}
% where $n_1$ and $n_2$ are the refractive indices of the media (air and glass, respectively), $\theta_i$ is the angle of incidence, and $\theta_t$ is the angle of refraction.

% left, bottom, right and top

\topic{Our reflection model.}
We illustrate our reflection model in~\figref{plane_parameterization}(b).
Applying Snell's Law at each interface, the outgoing angle of refraction $\theta_{t}$ equals the incoming angle of incidence $\theta_i$. 
% The glass sheet causes an offset. In practice, we find it unnecessary to model glass thickness. 
We adopt the following assumptions: 
1) the glass is impurity-free and does not scatter light diffusely; 
2) the glass is infinitely thin. 
Based on these assumptions, our model simplifies by considering only the primary and reflected incident ray. The primary ray remains unaltered in its path.
% \johannes{It's hard to understand what you're trying to convey here. I think you want to say something about the fact that glass sheets don't bend the rays because there are two medium transitions that cancel each other? There's an offset, though, which is worth mentioning? Even if you then say that it's not significant... if you think about it, it really wouldn't be hard to model glass thickness in our model, right? Why didn't we do it? Probably because it wasn't necessary. So maybe just mention that?}.

\topic{Ray-plane intersection.} 
Given a plane $\mathbf{s}$ and a primary ray $(\mathbf{o}, \mathbf{d})$, we can determine whether the ray intersects the plane. First, we find the intersection point $\mathbf{x}$ between the infinite plane $(\mathbf{p}, \mathbf{n})$ and the ray:
\begin{equation}
\mathbf{x} = \mathbf{o} + t \cdot \mathbf{d}, \quad t = \frac{(\mathbf{p} - \mathbf{o}) \cdot \mathbf{n}}{\mathbf{d} \cdot \mathbf{n}}.
\end{equation}
We discard rays if the plane is behind the ray's origin (i.e., when $t<0$). Additionally, we discard rays whose intersection points do not lie within the finite plane segment. When a ray intersects multiple planes, we select the intersection with the nearest plane.

\begin{figure}[t]
\centering%
\parbox[t]{1\linewidth}{\centering%
 \includegraphics[width=1\linewidth]{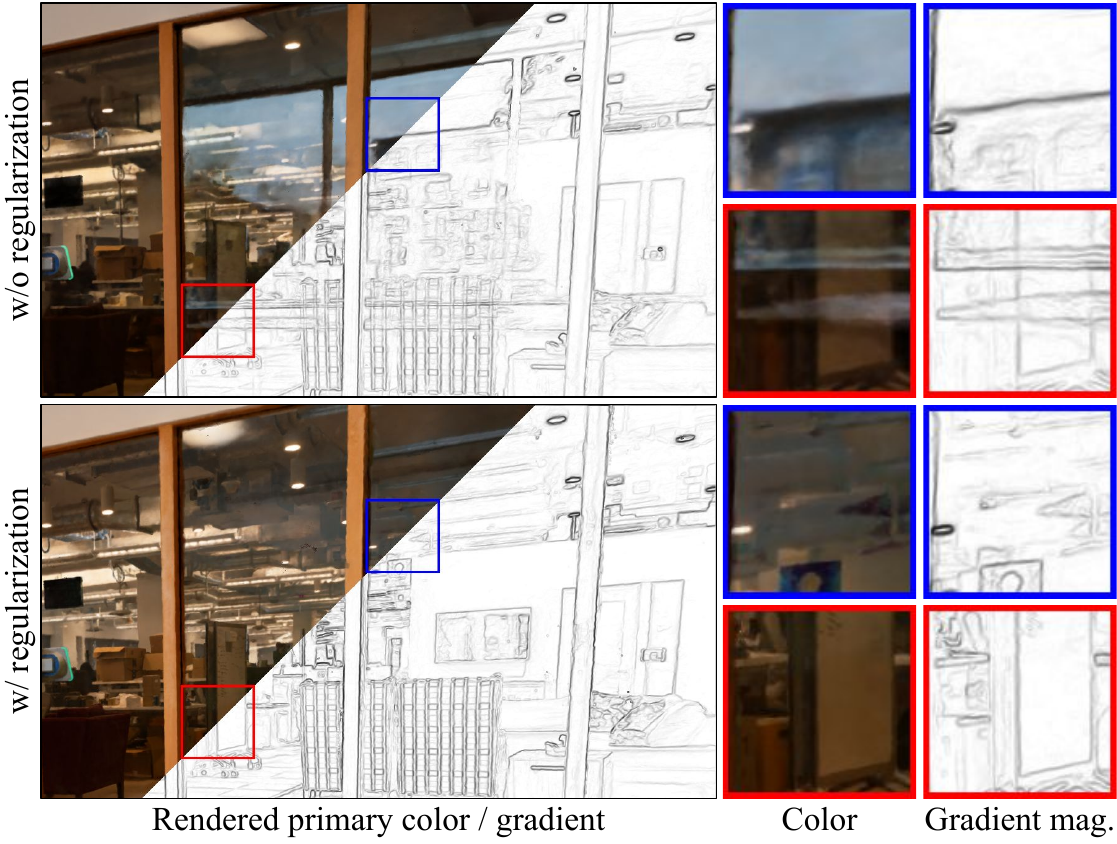}}
% \hfill%
\caption{\textbf{Sparse edge regularization.}
By applying the sparse edge regularization, the primary view is clean and free of false geometries.
% \changil{Redundant to Figure 4. I think this figure is better than Figure 4. Maybe use a different color coding than grayscale? Also show RGB images too?}
}
\label{fig:effectiveness_edge_loss}
\end{figure}
\topic{Casting reflected ray.}
If a ray intersects a plane, we cast a reflected ray. 
The new origin of the reflected ray is $\mathbf{x}$, and the new viewing direction is given by:
\begin{equation}
\mathbf{d}' = \mathbf{d} - 2 (\mathbf{d} \cdot \mathbf{n}) \cdot \mathbf{n}.
\label{eq:reflected_viewing_direction}
\end{equation}
We query the same NeRF model to obtain the volume density $\sigma$ and color $\mathbf{c}$ for the 3D points along the reflected ray $\mathbf{r}' = (\mathbf{x}, \mathbf{d}')$.
Now, if a ray intersects the plane, we can perform volume rendering along each ray to obtain the primary color $\mathbf{C}(\mathbf{r})$ and the source of reflection $\mathbf{C}(\mathbf{r}')$. 
Next, we discuss how we compose these two components to calculate the final color.

\topic{Composing primary and reflected ray.}
We introduce a dual-function attenuation field to model both actual attenuation (such as Fresnel effects) and compositing, diverging from conventional physics-based models that rely solely on Fresnel equations to determine reflection and transmission coefficients based on the viewing direction.
Fresnel equations rely on the assumption of a linear color space. 
However, following tone mapping, RGB values no longer reside in a linear space, thus invalidating the direct application of the Fresnel equation. 
Therefore, we propose optimizing an adaptable attenuation field that compensates for the non-linearity introduced by HDR tone mapping, models the Fresnel effect, and absorbs errors due to intensity clipping. 
Since attenuation is associated with the intensity of the reflection, our attenuation field maps the 3D position $(x, y, z)$ along the reflected ray $\mathbf{r}'$ and the normalized reflected viewing direction $\mathbf{d}'$ to the corresponding attenuation $a$, rather than using the primary ray to predict the attenuation as previous work did~\cite{Guo_2022_CVPR}:
\begin{align}
a &= \text{MLP}_{a}(x, y, z, \mathbf{d}') \\
\mathbf{A}(\mathbf{r}') &= \sum_{k=1}^{K} T_k \left( 1-e^{-\sigma_k\delta_k} \right) a_k
\label{eq:attenuation_field}
\end{align}
To calculate the final color, we additionally obtain the transmittance $\mathrm{T}$ of each ray at the intersection using Eq. (\ref{eq:transmittance}).
The final color is composited by:
\begin{equation}
\mathbf{C}^\textit{comp}(\mathbf{r}) = \mathbf{C}(\mathbf{r}) + T \cdot \mathbf{C^A}(\mathbf{r}'),
\label{eq:rendering_composed}
\end{equation}
where the attenuated reflection is calculated by
\begin{equation}
\mathbf{C^A}(\mathbf{r}') = \sum_{k=1}^{K} T_k \left( 1-e^{-\sigma_k\delta_k} \right) (a_k \cdot c_k)
\label{eq:attenuated_color}
\end{equation}
We demonstrate in the experimental section that our attenuation field consistently learns attenuation with robustness in real-world scenarios.

\begin{figure*}[t]
\newlength\figwidthNovel
\setlength\figwidthNovel{0.162\linewidth}
\centering%
\parbox[t]{\figwidthNovel}{\centering%
  \fbox{\includegraphics[trim=0 0 0 0, clip=true, width=\figwidthNovel]{fig/results/meeting_room/rgb_000001.jpg}}\\%
  \fbox{\includegraphics[trim=0 0 0 0, clip=true, width=\figwidthNovel]{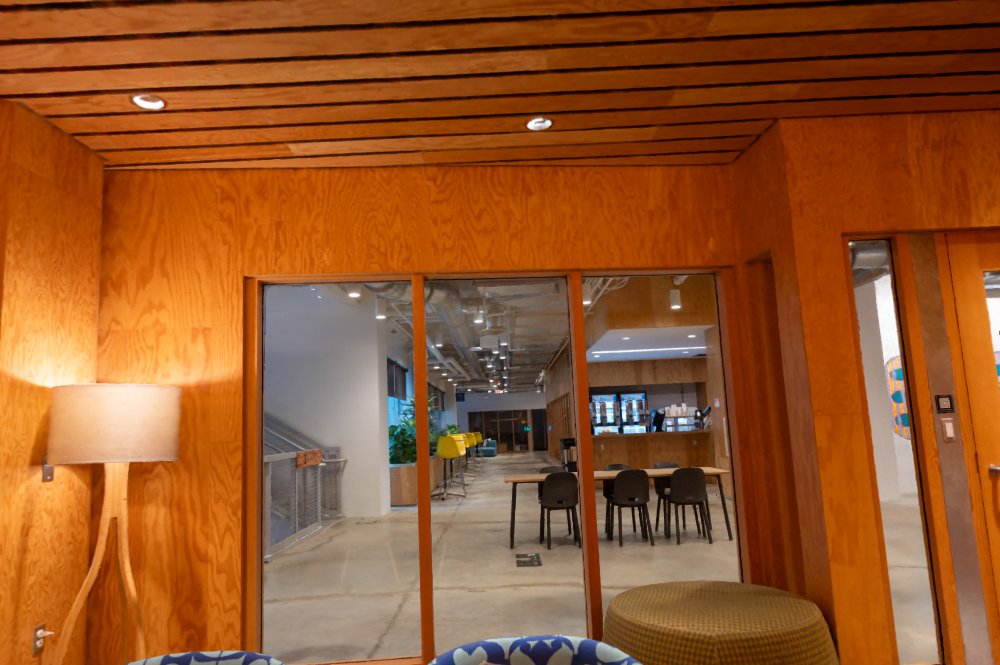}}\\%
  \fbox{\includegraphics[trim=0 0 0 0, clip=true, width=\figwidthNovel]{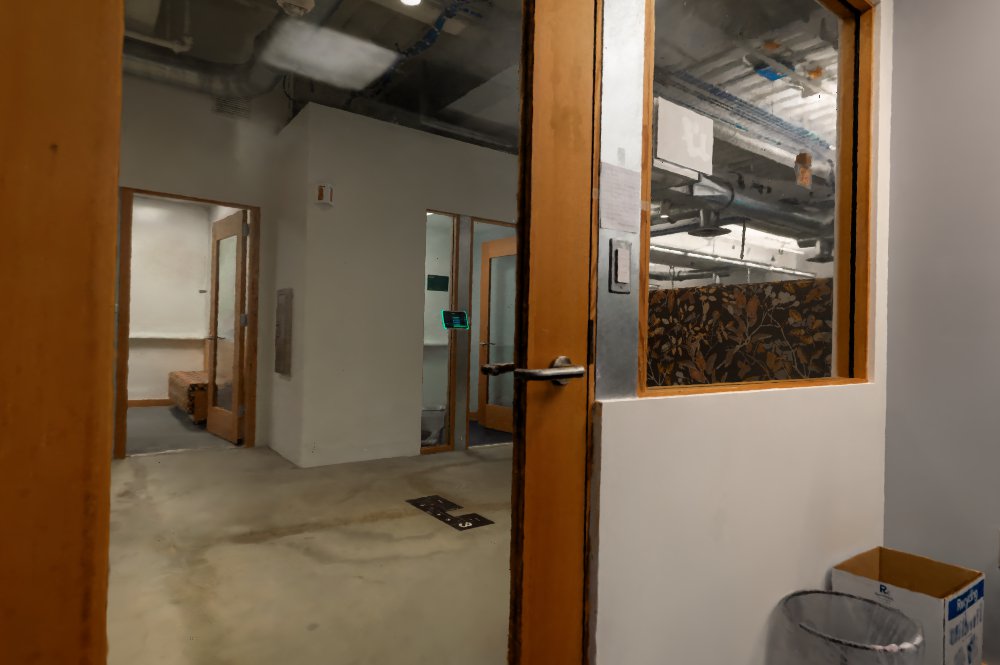}}\\%
  \fbox{\includegraphics[trim=0 0 0 0, clip=true, width=\figwidthNovel]{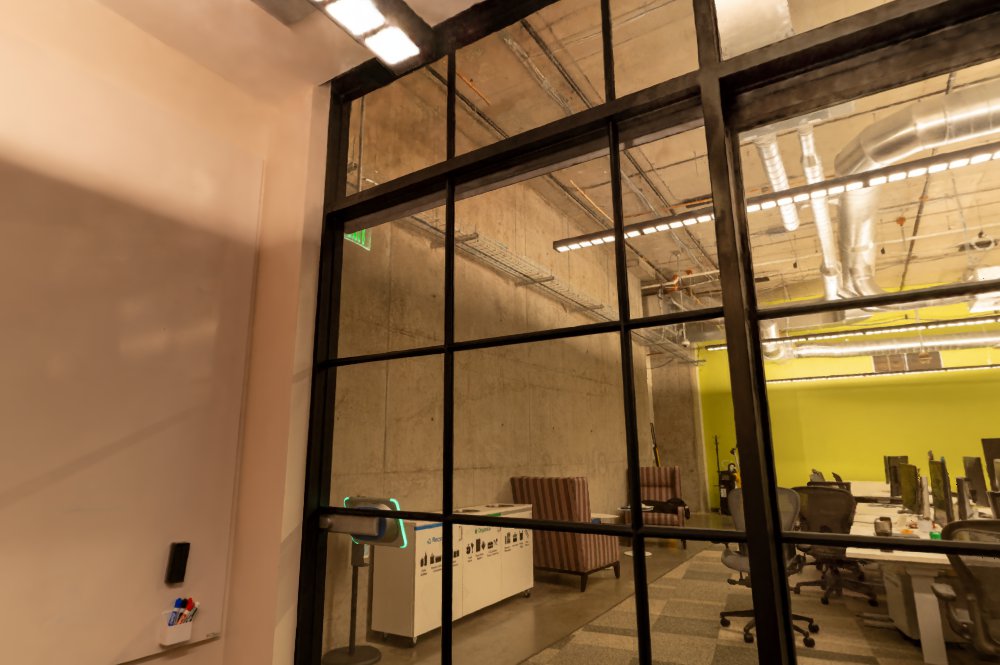}}\\%
  \fbox{\includegraphics[trim=0 0 0 0, clip=true, width=\figwidthNovel]{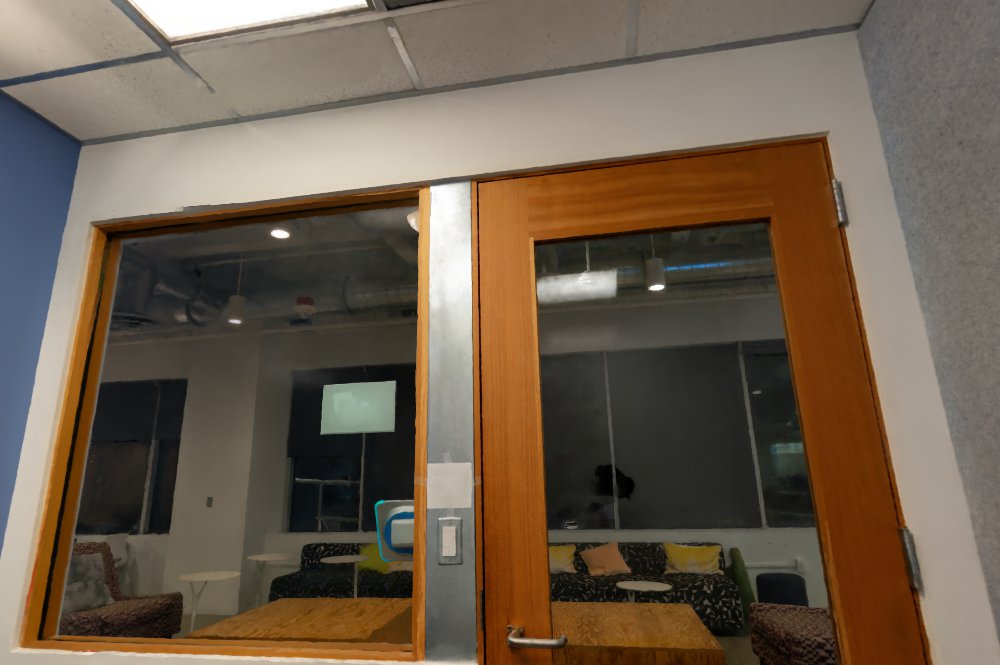}}\\%
  \fbox{\includegraphics[trim=0 0 0 0, clip=true, width=\figwidthNovel]{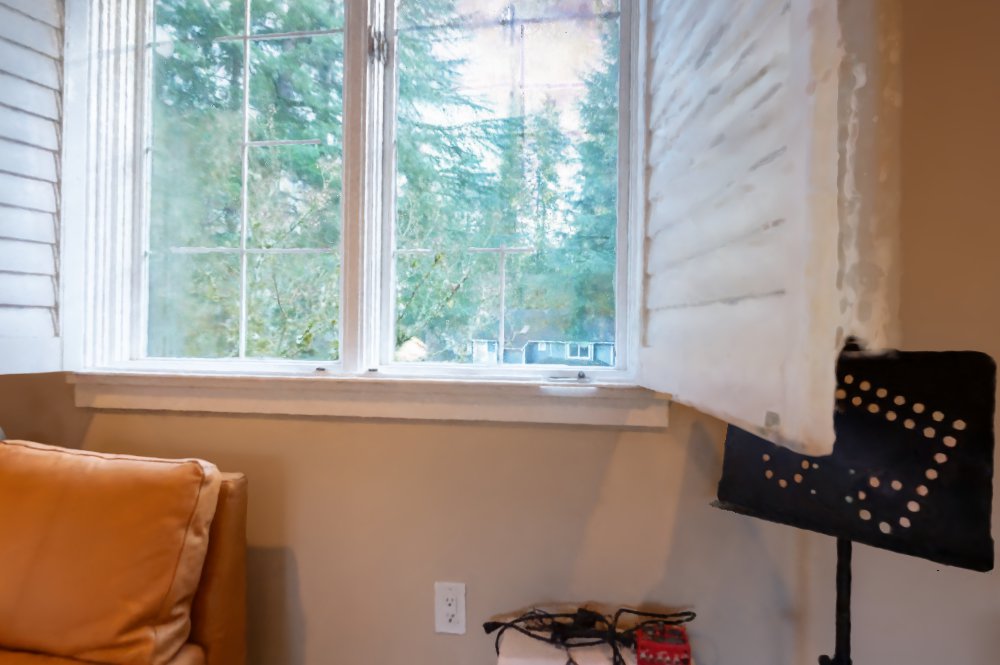}}\\%
   \footnotesize Primary rendering}%
\hfill%
\parbox[t]{\figwidthNovel}{\centering%
  \fbox{\includegraphics[trim=0 0 0 0, clip=true, width=\figwidthNovel]{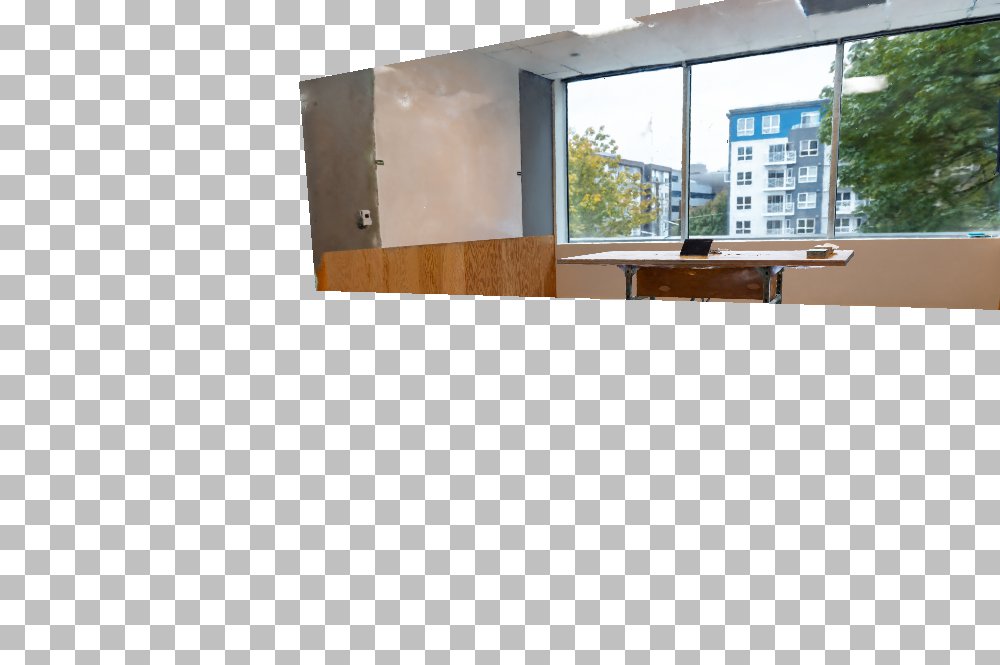}}\\%
  \fbox{\includegraphics[trim=0 0 0 0, clip=true, width=\figwidthNovel]{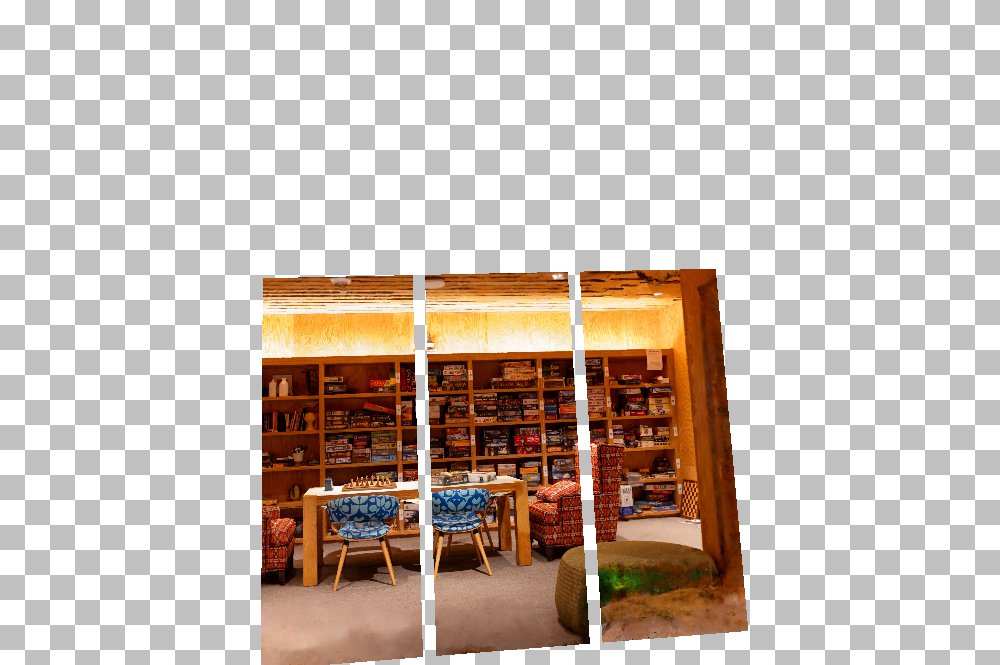}}\\%
  \fbox{\includegraphics[trim=0 0 0 0, clip=true, width=\figwidthNovel]{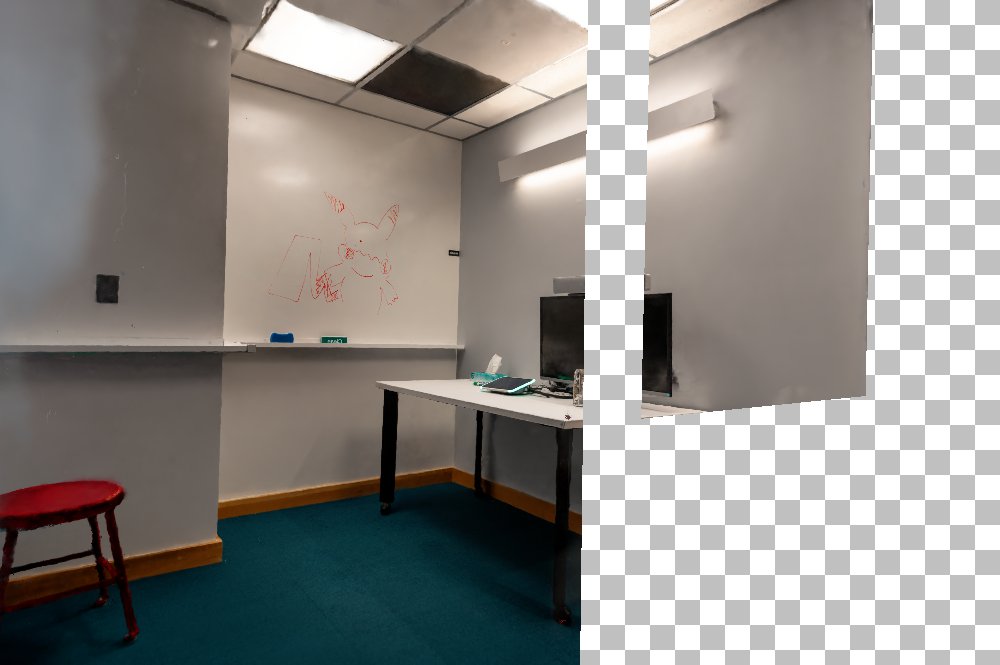}}\\%
  \fbox{\includegraphics[trim=0 0 0 0, clip=true, width=\figwidthNovel]{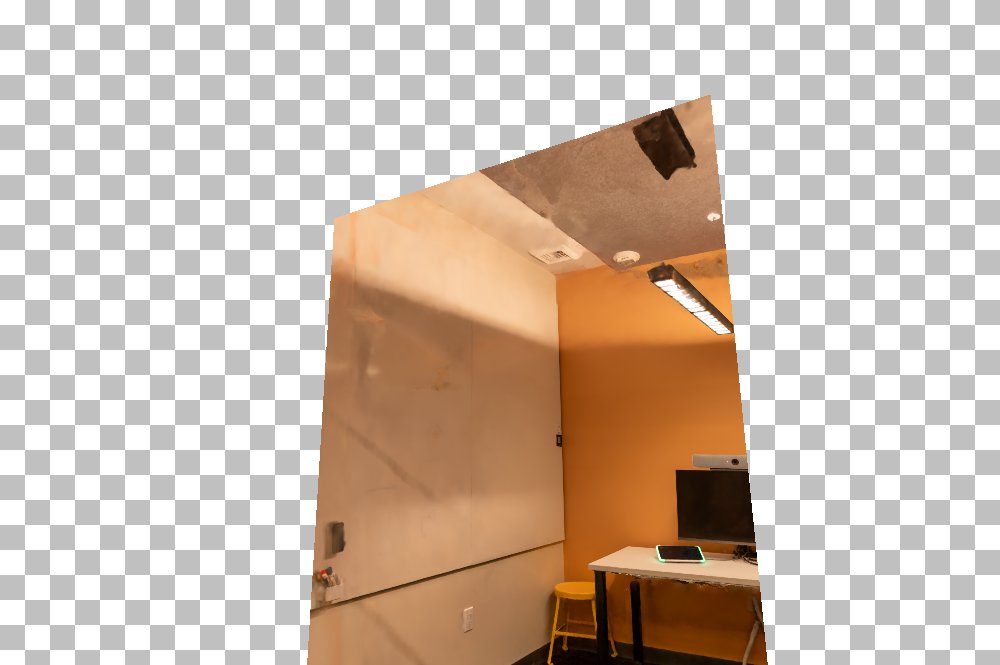}}\\%
  \fbox{\includegraphics[trim=0 0 0 0, clip=true, width=\figwidthNovel]{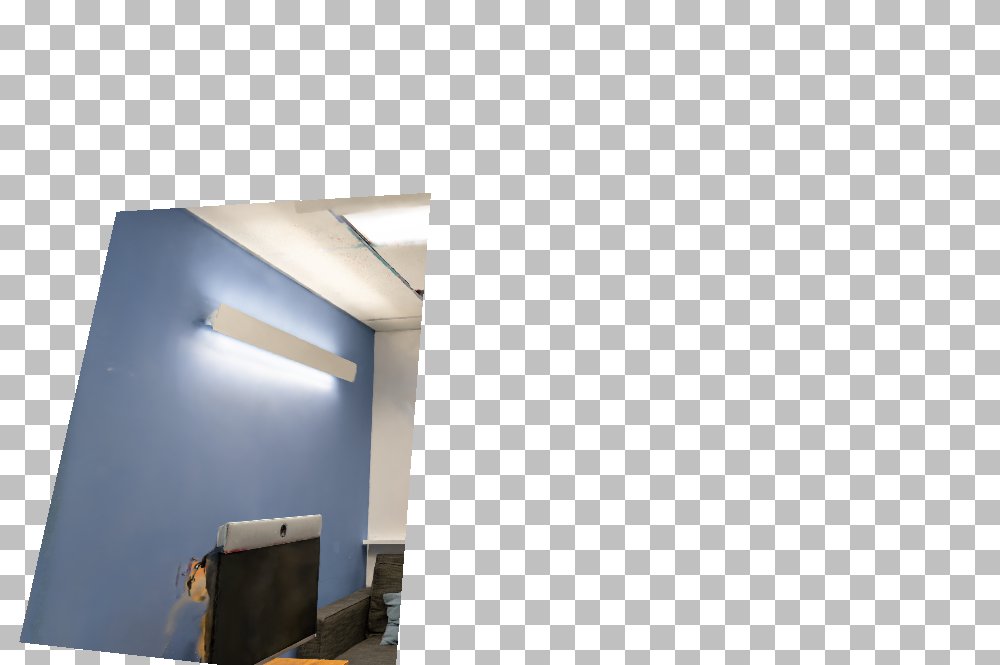}}\\%
  \fbox{\includegraphics[trim=0 0 0 0, clip=true, width=\figwidthNovel]{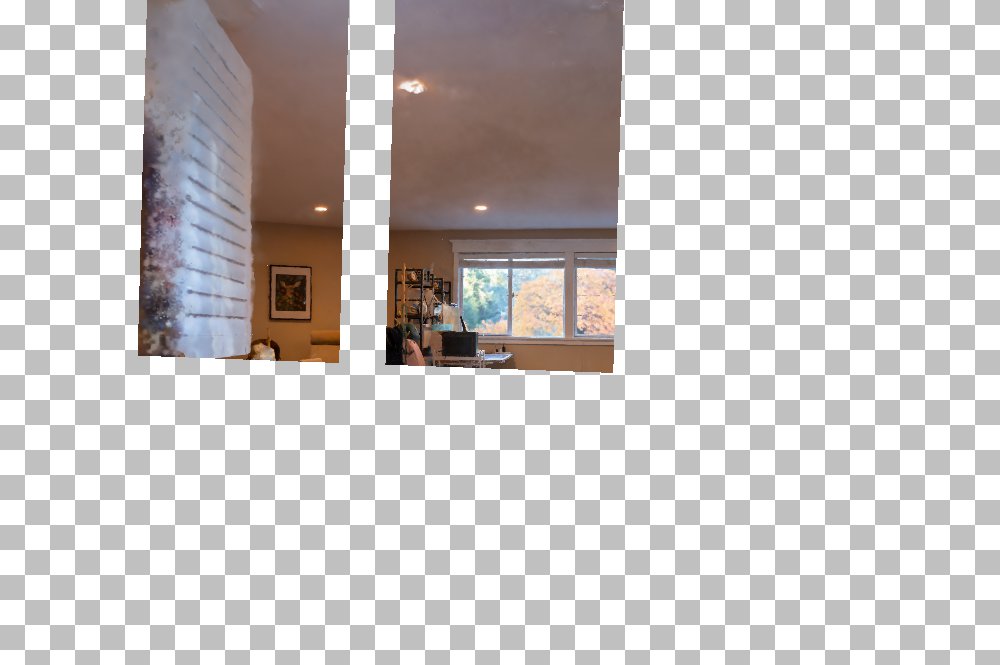}}\\%
   \footnotesize Source of reflection}%
\hfill%
\parbox[t]{\figwidthNovel}{\centering%
  \fbox{\includegraphics[trim=0 0 0 0, clip=true, width=\figwidthNovel]{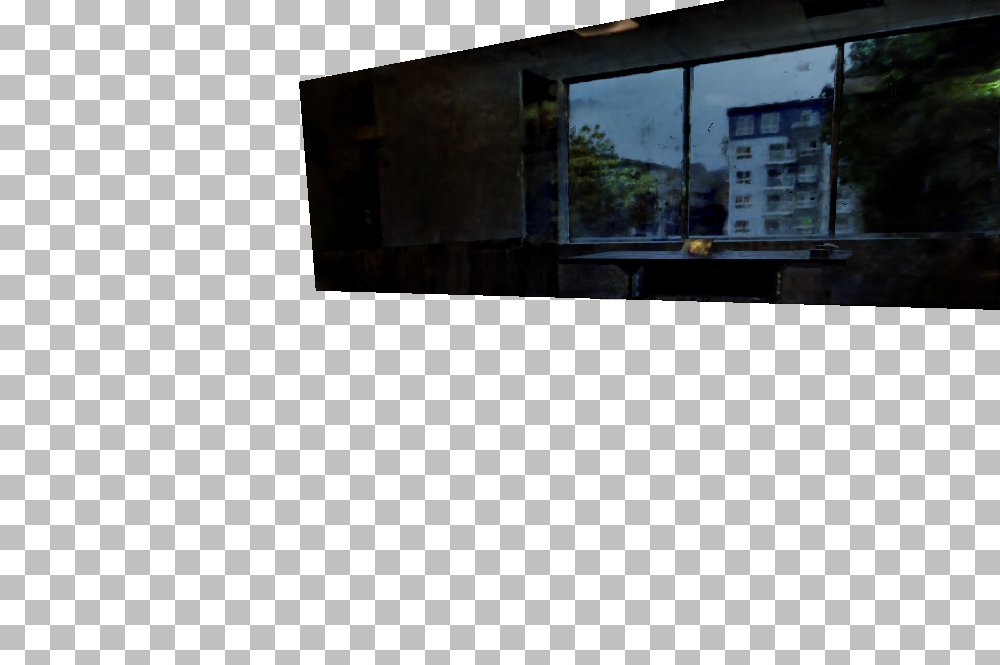}}\\%
  \fbox{\includegraphics[trim=0 0 0 0, clip=true, width=\figwidthNovel]{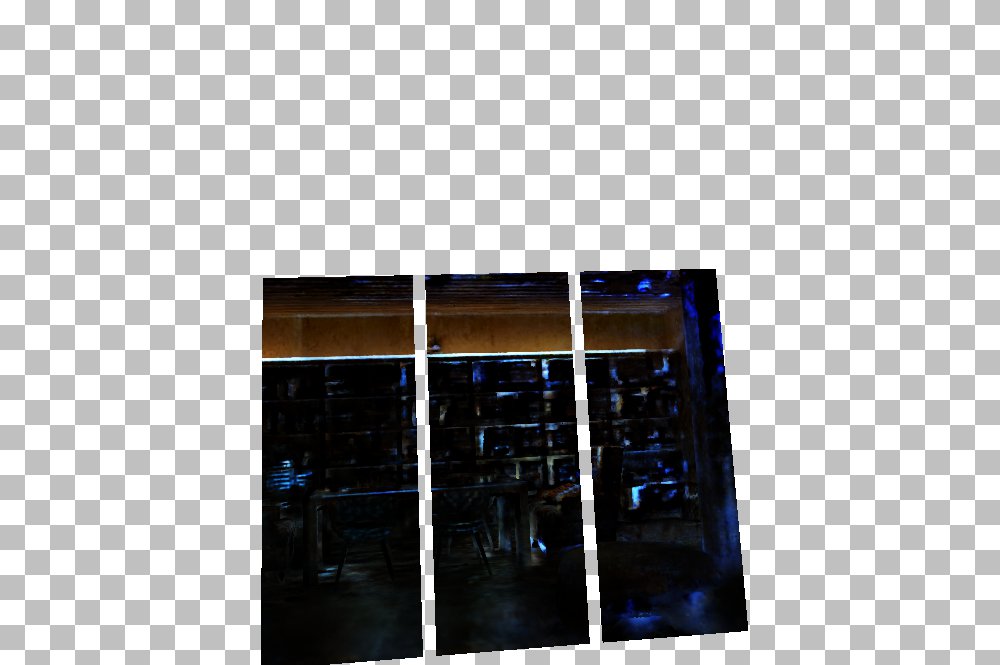}}\\%
  \fbox{\includegraphics[trim=0 0 0 0, clip=true, width=\figwidthNovel]{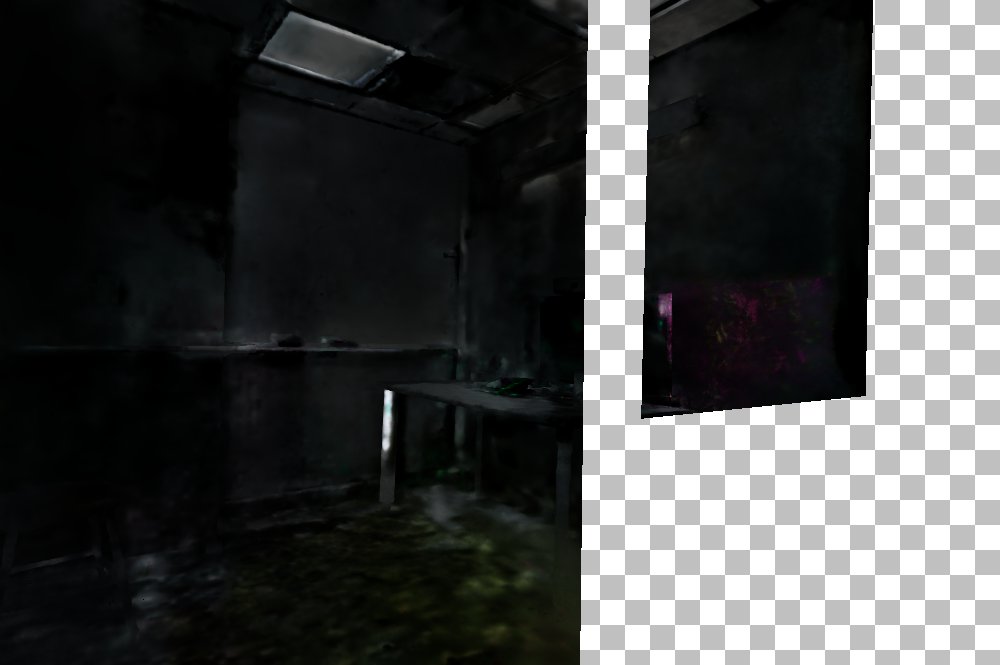}}\\%
  \fbox{\includegraphics[trim=0 0 0 0, clip=true, width=\figwidthNovel]{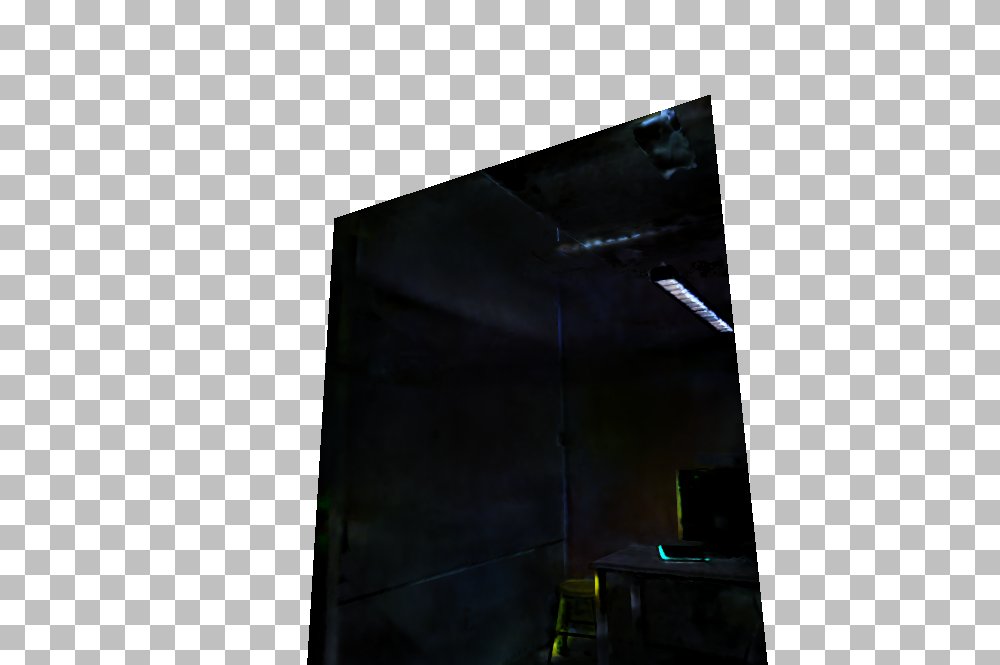}}\\%
  \fbox{\includegraphics[trim=0 0 0 0, clip=true, width=\figwidthNovel]{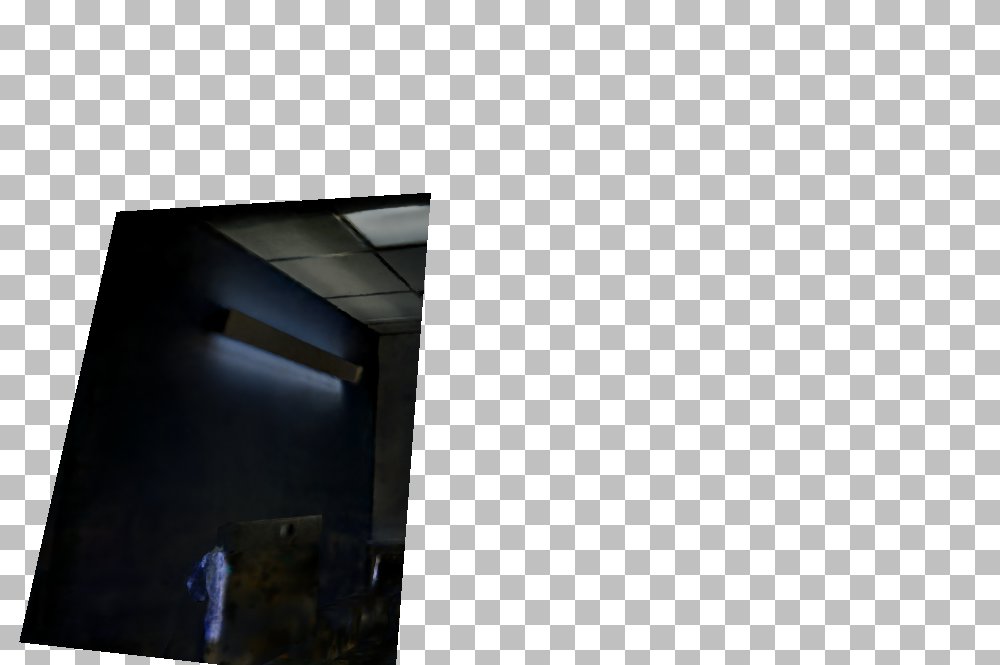}}\\%
  \fbox{\includegraphics[trim=0 0 0 0, clip=true, width=\figwidthNovel]{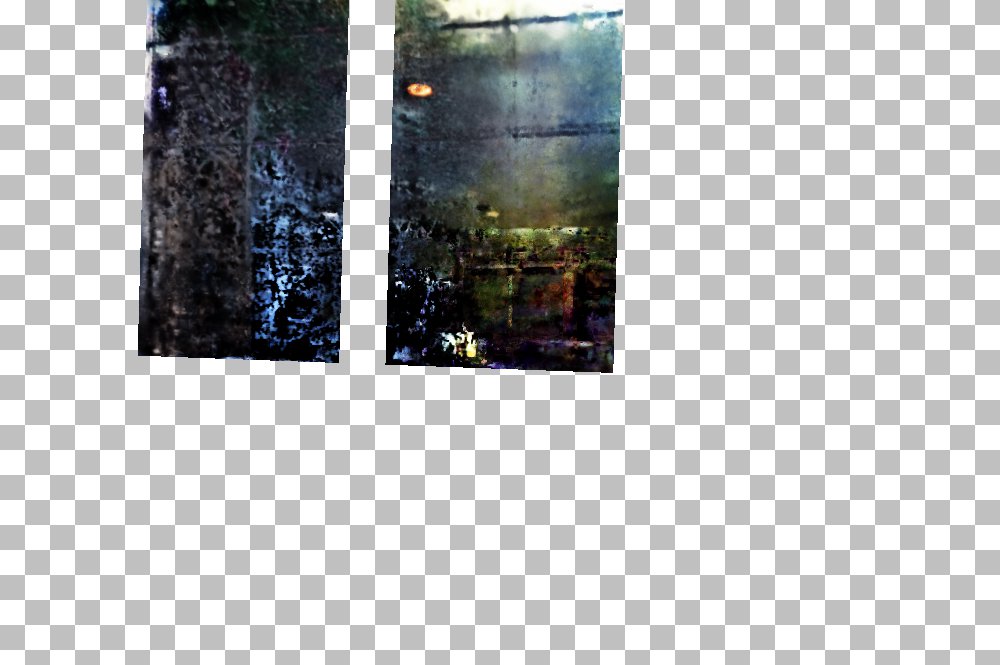}}\\%
   \footnotesize Attenuation}%
\hfill%
\parbox[t]{\figwidthNovel}{\centering%
  \fbox{\includegraphics[trim=0 0 0 0, clip=true, width=\figwidthNovel]{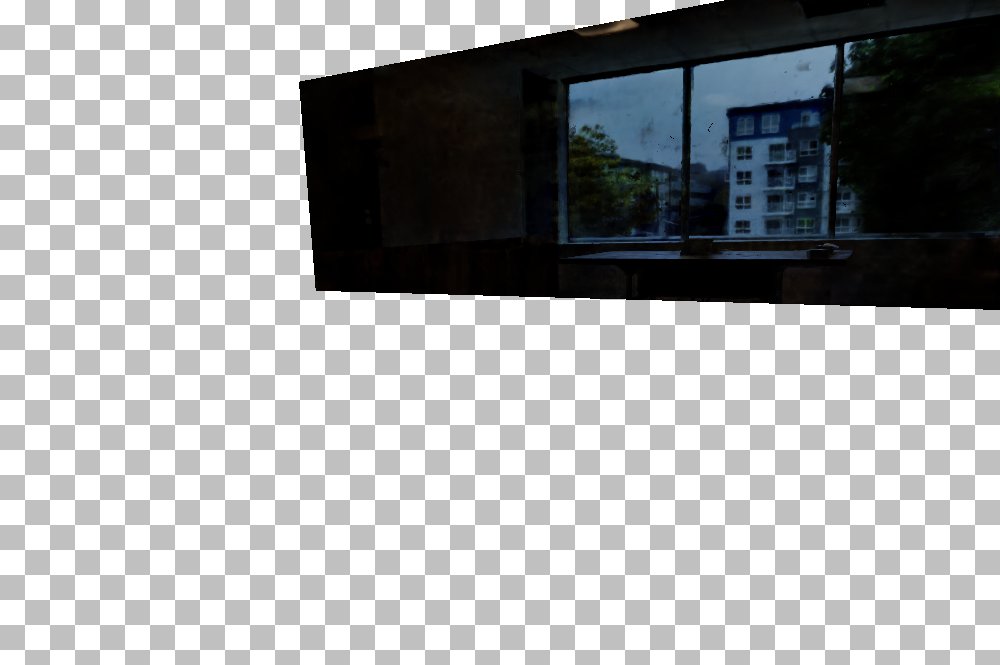}}\\%
  \fbox{\includegraphics[trim=0 0 0 0, clip=true, width=\figwidthNovel]{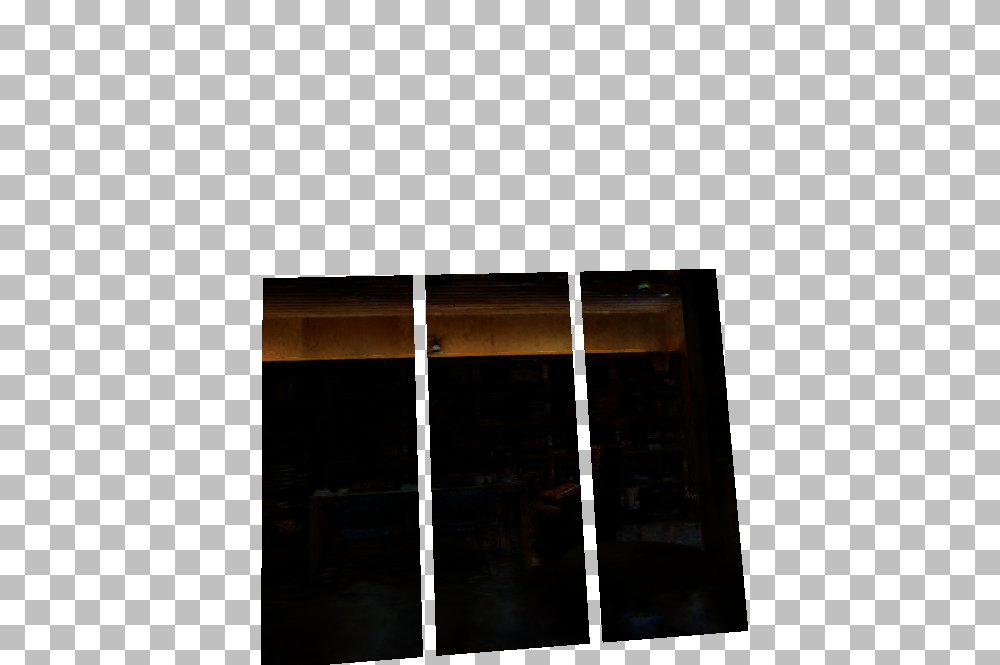}}\\%
  \fbox{\includegraphics[trim=0 0 0 0, clip=true, width=\figwidthNovel]{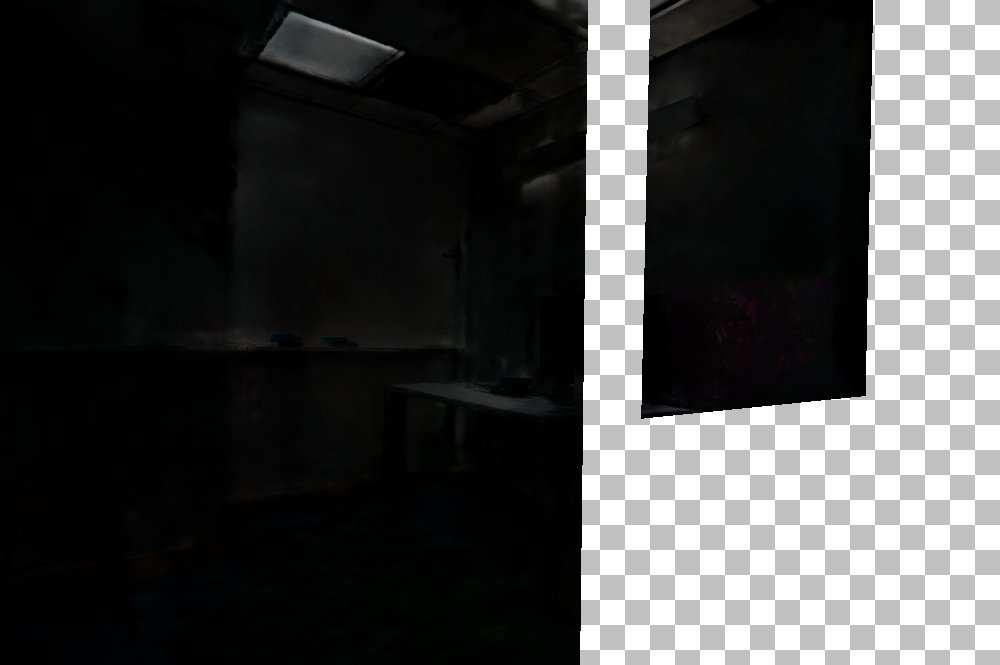}}\\%
  \fbox{\includegraphics[trim=0 0 0 0, clip=true, width=\figwidthNovel]{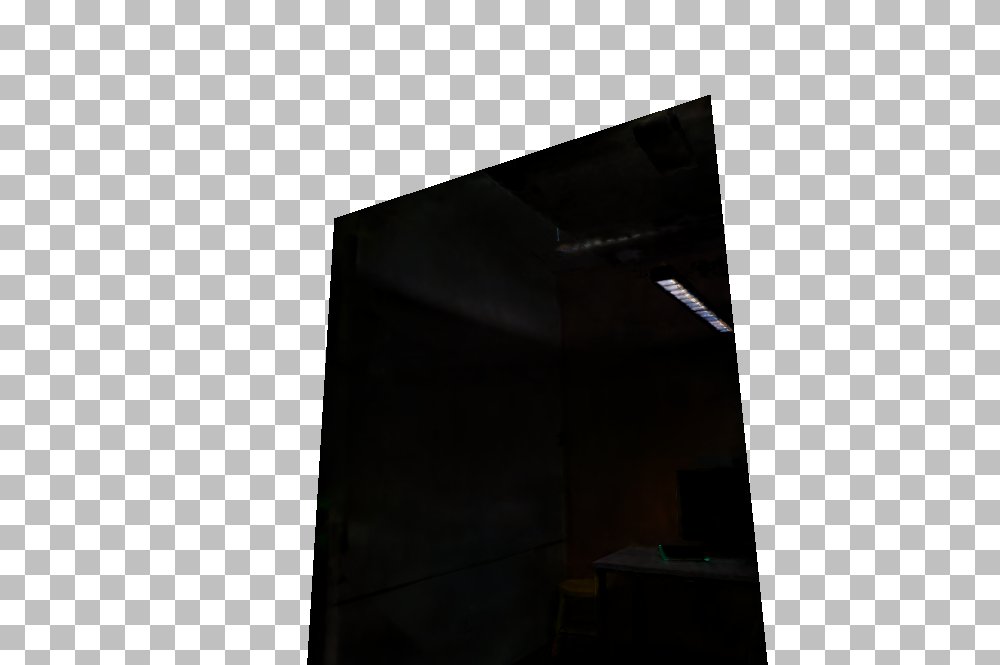}}\\%
  \fbox{\includegraphics[trim=0 0 0 0, clip=true, width=\figwidthNovel]{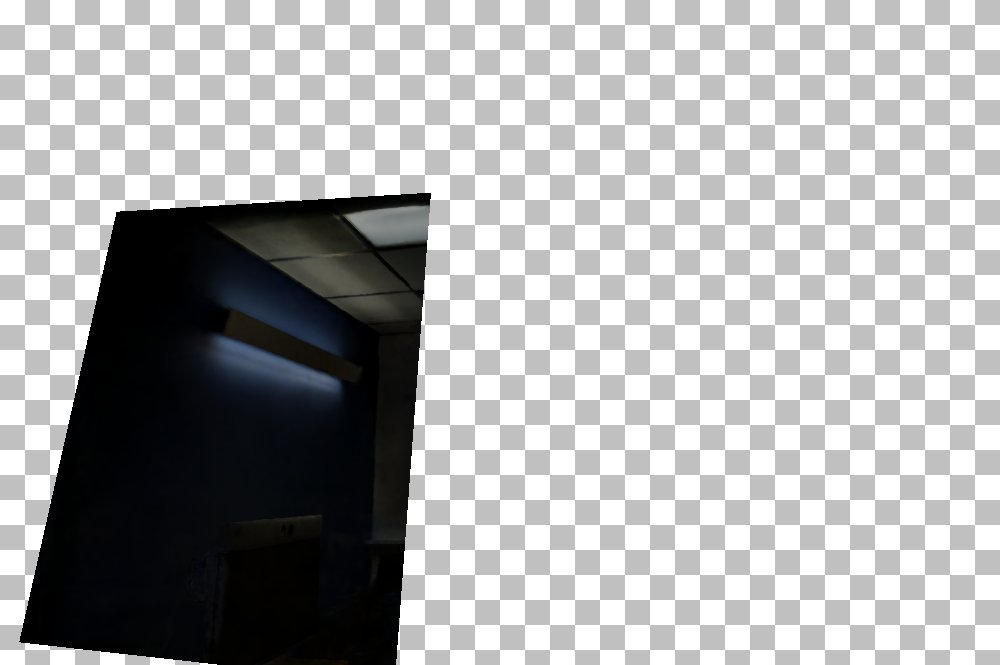}}\\%
  \fbox{\includegraphics[trim=0 0 0 0, clip=true, width=\figwidthNovel]{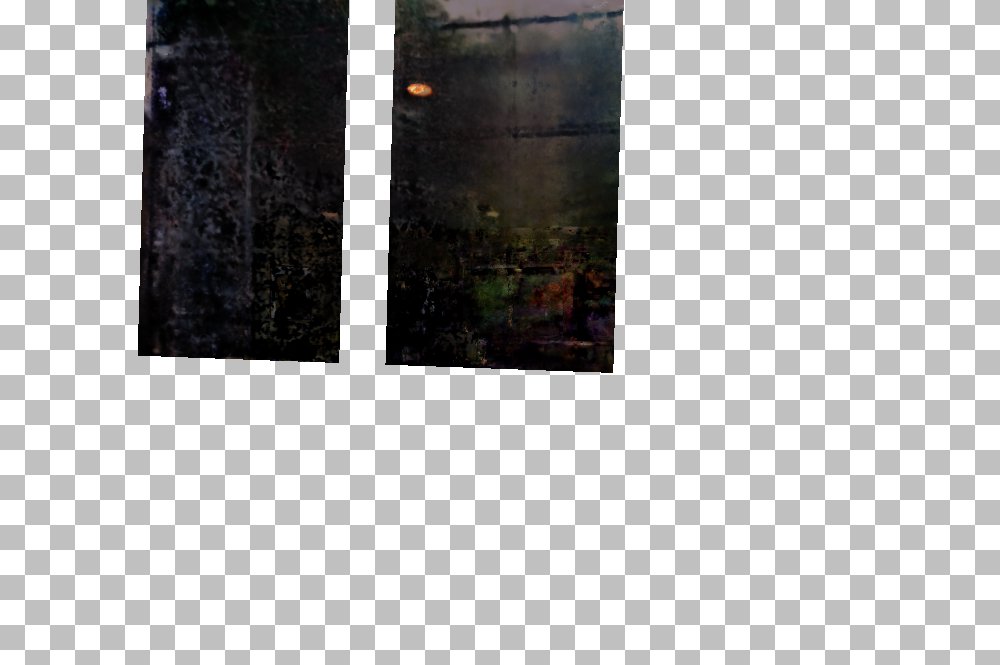}}\\%
   \footnotesize Reflection rendering}%
\hfill%
\parbox[t]{\figwidthNovel}{\centering%
  \fbox{\includegraphics[trim=0 0 0 0, clip=true, width=\figwidthNovel]{fig/results/meeting_room/rgb_composite_000001.jpg}}\\%
  \fbox{\includegraphics[trim=0 0 0 0, clip=true, width=\figwidthNovel]{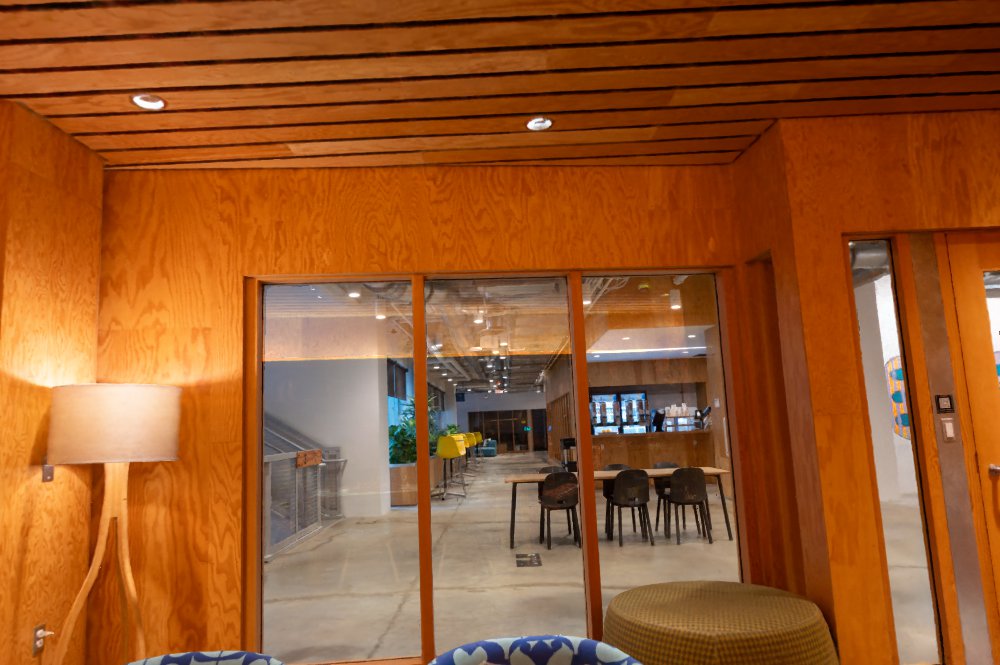}}\\%
  \fbox{\includegraphics[trim=0 0 0 0, clip=true, width=\figwidthNovel]{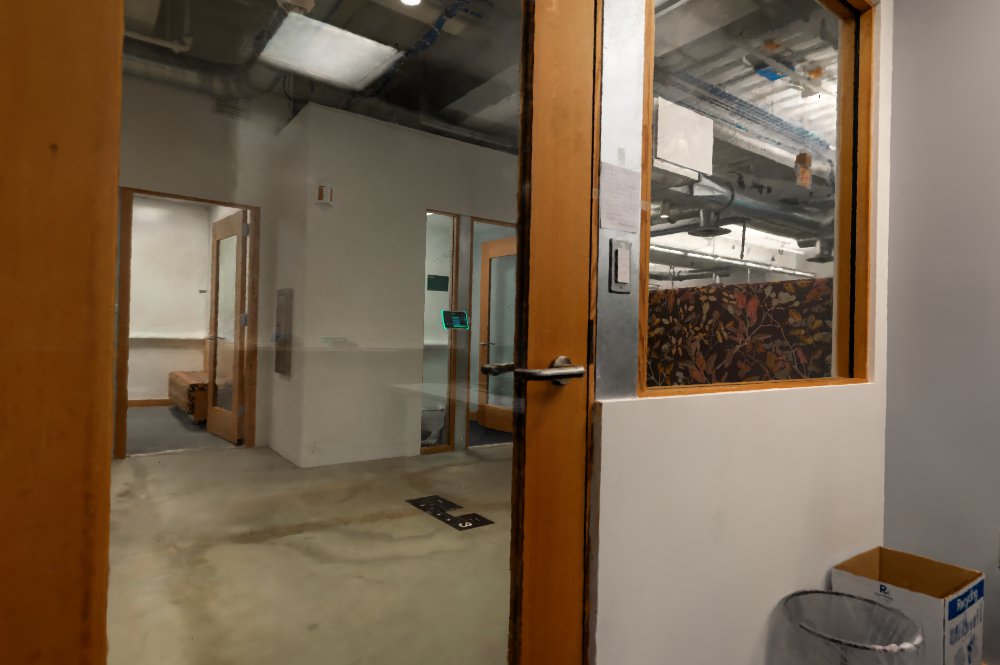}}\\%
  \fbox{\includegraphics[trim=0 0 0 0, clip=true, width=\figwidthNovel]{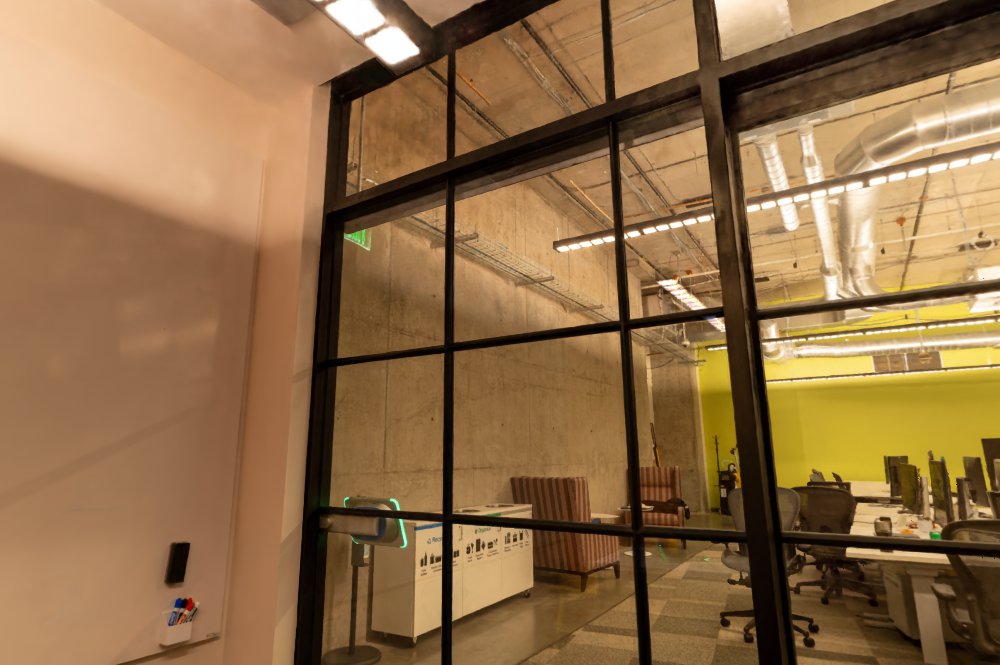}}\\%
  \fbox{\includegraphics[trim=0 0 0 0, clip=true, width=\figwidthNovel]{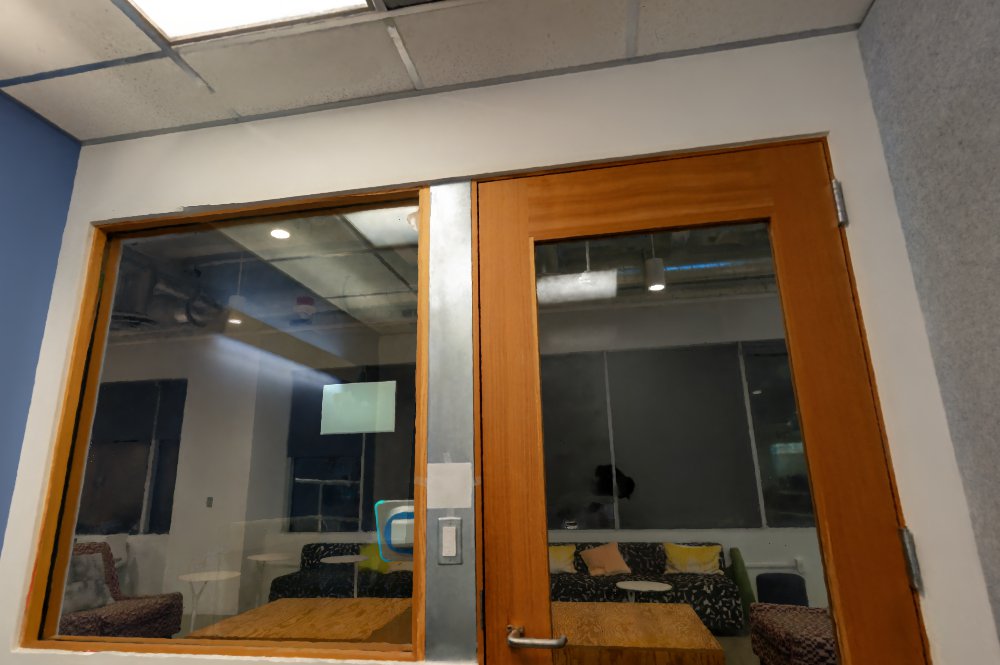}}\\%
  \fbox{\includegraphics[trim=0 0 0 0, clip=true, width=\figwidthNovel]{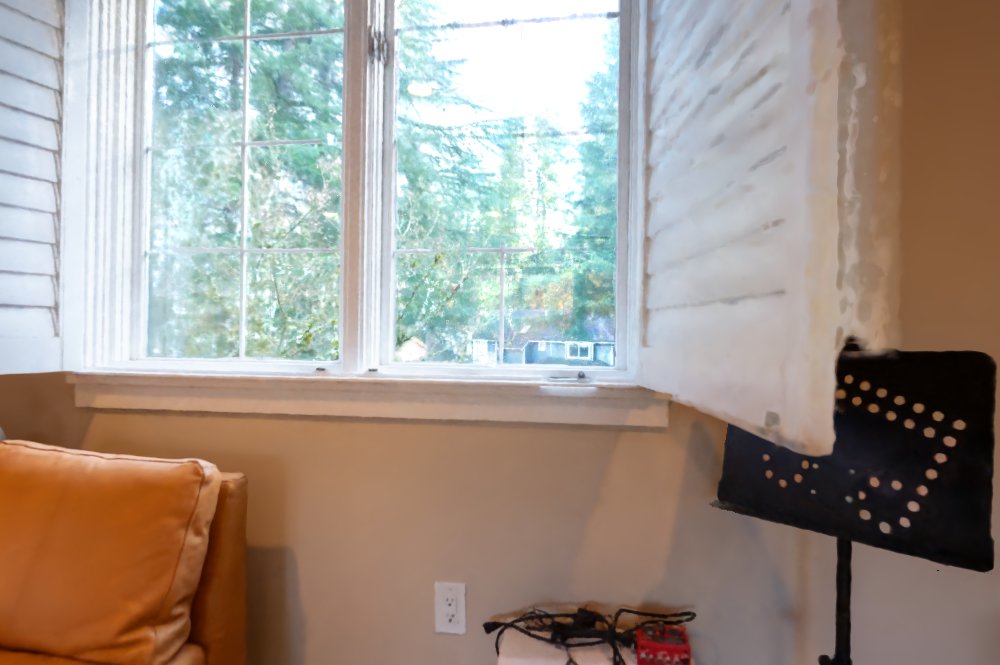}}\\%
   \footnotesize Composed rendering}%
\hfill%
\parbox[t]{\figwidthNovel}{\centering%
  \fbox{\includegraphics[trim=0 0 0 0, clip=true, width=\figwidthNovel]{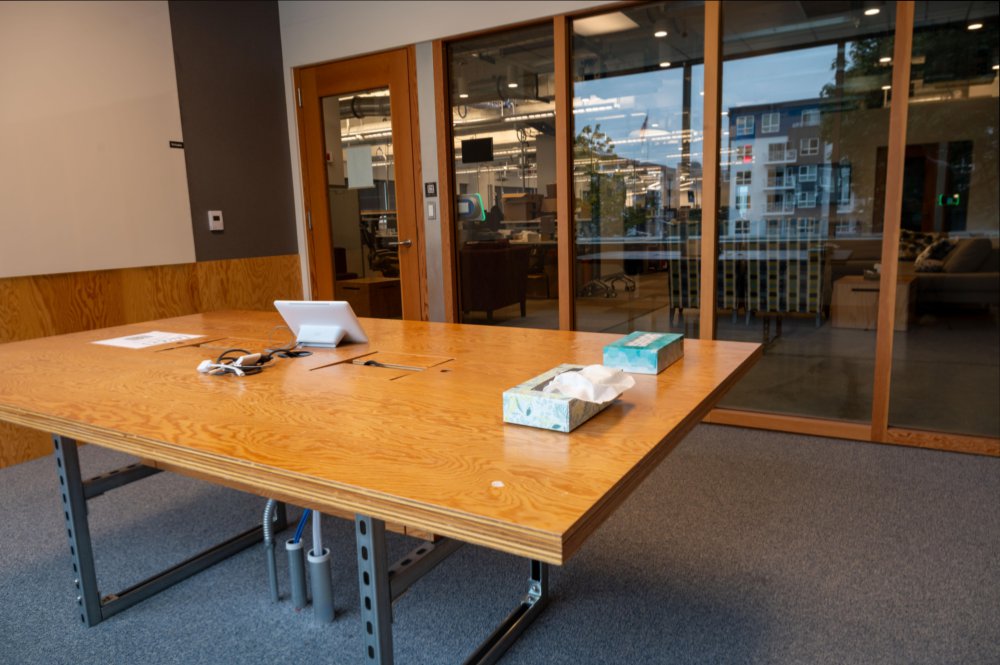}}\\%
  \fbox{\includegraphics[trim=0 0 0 0, clip=true, width=\figwidthNovel]{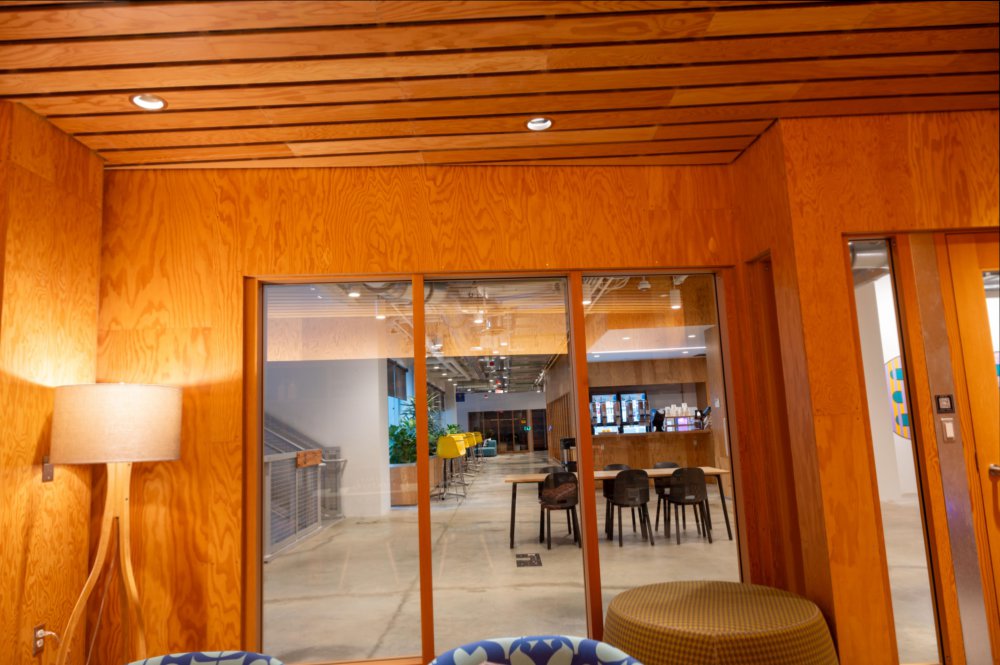}}\\%
  \fbox{\includegraphics[trim=0 0 0 0, clip=true, width=\figwidthNovel]{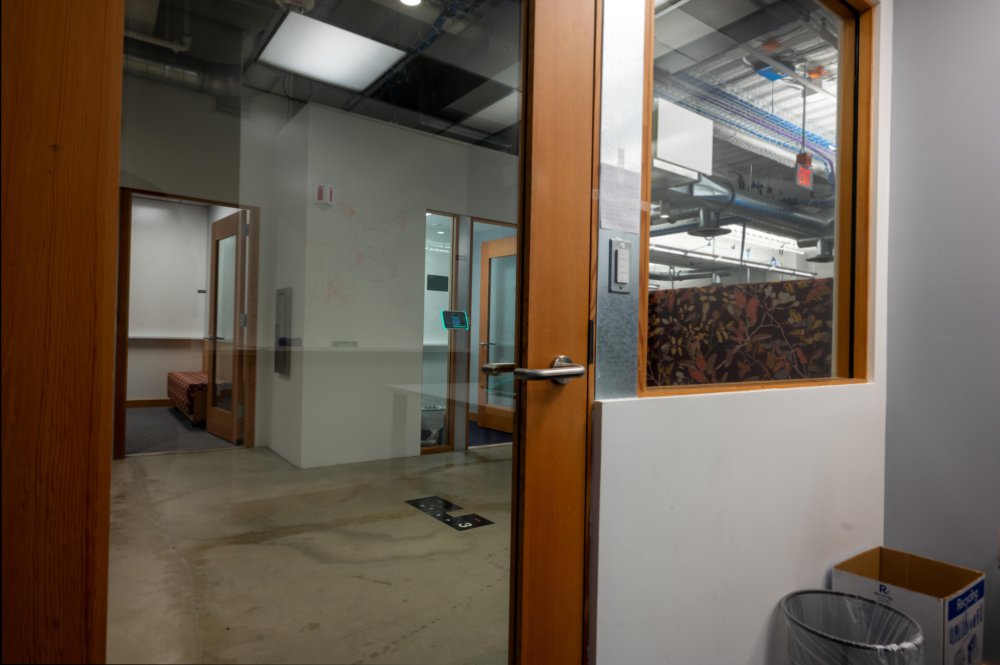}}\\%
  \fbox{\includegraphics[trim=0 0 0 0, clip=true, width=\figwidthNovel]{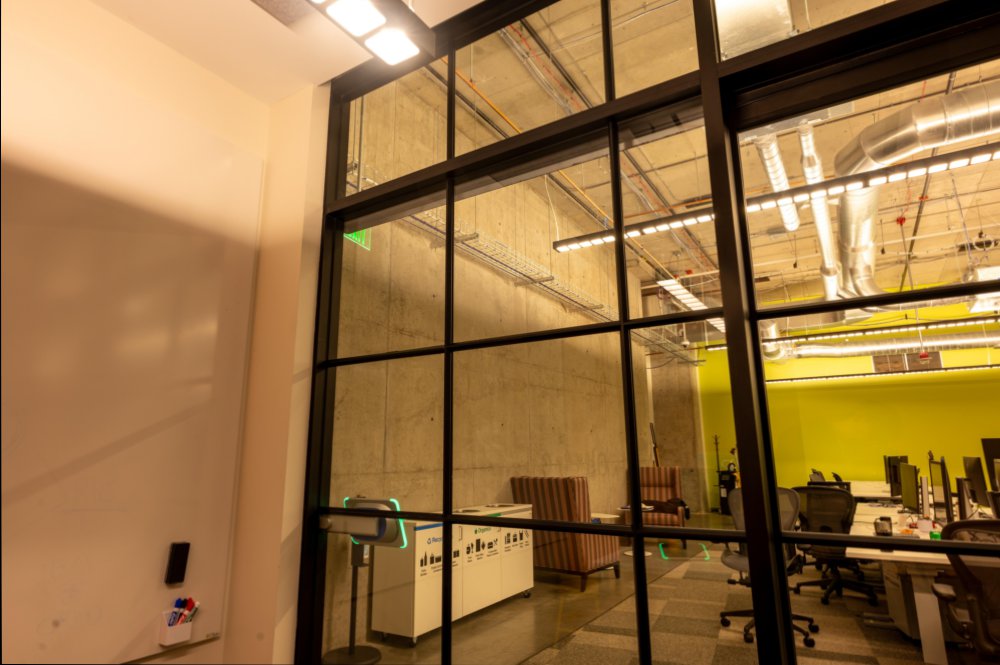}}\\%
  \fbox{\includegraphics[trim=0 0 0 0, clip=true, width=\figwidthNovel]{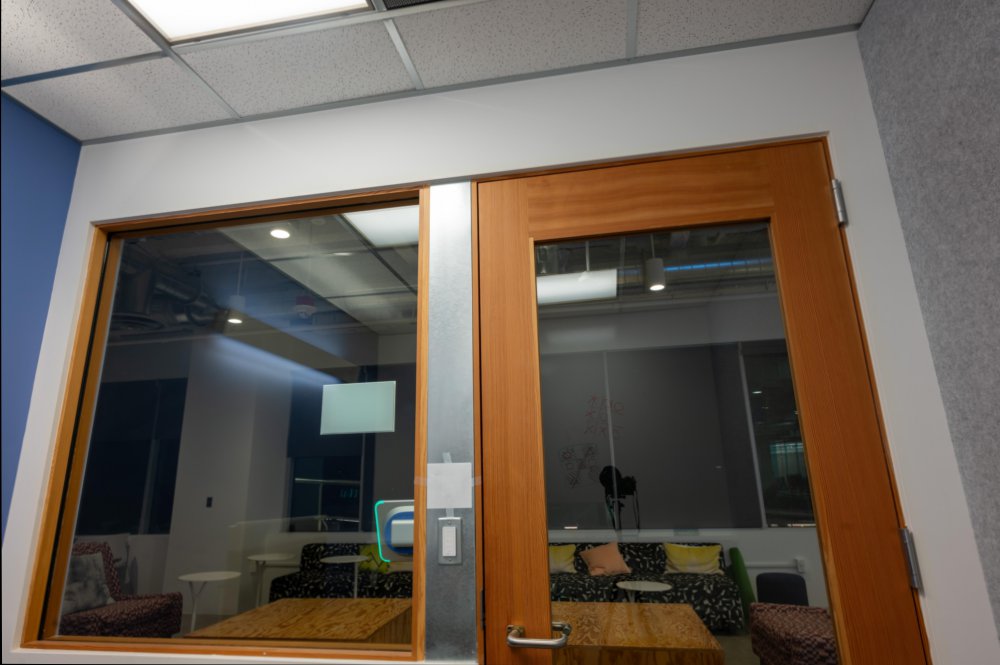}}\\%
  \fbox{\includegraphics[trim=0 0 0 0, clip=true, width=\figwidthNovel]{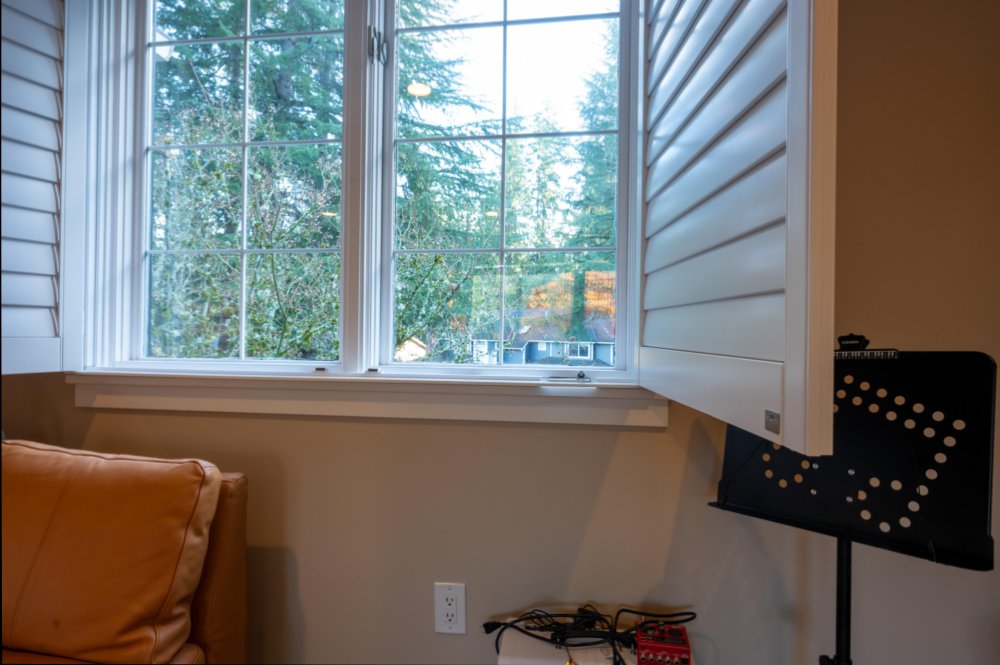}}\\%
   \footnotesize Ground truth}%
% left, bottom, right and top
% \vspace{\figcapmargin}
\caption{\textbf{Individual component.}
Rendering along the primary ray yields a reflection-free image. 
Rendering along the reflected ray captures the source of the reflection. The final image is composed by combining the reflection-free image with the attenuated source of the reflection.
Our method ensures that reflections are effectively isolated and can be manipulated independently from the reflection-free scene reconstruction.
% \changil{Can we come up with an interesting usecase/application for manipulating isolated reflections?}
}
\label{fig:decomposition}
\end{figure*}

% \changil{You should rather show the geometry when you say this. E.g., compare the NeRF depth you get with your reflection-free model vs the depth of the vanila NeRF? Can we also show some areas that are reconstructed better with explicit reflection modeling? See my comments in Figure 1 as well.}

% \tiny	F-tiny.jpg
% \scriptsize	F-scriptsize.jpg
% \footnotesize	F-footnotesize.jpg
% \small	F-small.jpg
% \normalsize	F-normalsize.jpg
% \large	F-large.jpg
% \Large	F-large2.jpg
% \LARGE	F-large3.jpg
% \huge	F-huge.jpg
% \Huge	F-huge2.jpg

\subsection{Training}
\label{sec:training}
Training our reflection-aware NeRF requires jointly optimizing the plane geometry $\mathbf{s}$, the radiance field, and the attenuation field.
In the following, we discuss two key training strategies: 1) sparse edge regularization; 2) scheduling.

\topic{Sparse edge regularization.}
Even with our proposed reflection model, the network still tends to reject the model by learning a small attenuation number. It has a propensity to introduce false geometry along the primary ray $\mathbf{C}(\mathbf{r})$, situated at the same depth as the actual object along the reflected ray $\mathbf{C}(\mathbf{r}')$.
% as shown in~\figref{edge_loss_calculation}.
To mitigate this issue of duplication, we leverage a sparse edge regularization.
Utilizing the Sobel operator $\nabla$, we compute the gradients to identify edges within the image.
Sparse edge regularization is applied by minimizing element-wise multiplication between the gradient of the primary colors and those of the reflected colors:
\begin{equation}
\mathcal{L}_\text{edge} = \left \| \nabla(\mathbf{C}(\mathbf{r})) * \nabla(\mathbf{C}(\mathbf{r}')) \right \|_{1}.
\label{eq:edge_loss}
\end{equation}
We detach the gradient flow from $\mathcal{L}_\text{edge}$ to $\mathbf{C}(\mathbf{r}')$ to constrain the regularization strictly to the primary colors. The underlying principle is that if an edge is discernible in the reflected color image (indicative of the source of reflection), such an edge should not concurrently exist in the primary color image. This regularization effectively prevents the primary color from incorporating duplicated objects, as shown in \figref{effectiveness_edge_loss}.

\topic{Training scheduling.}
We first initialize the plane geometric parameters $\mathbf{s}$ using the annotation.
Then, we jointly optimize the plane geometry $\mathbf{s}$, the radiance field, and the attenuation field using photometric loss:
\begin{equation}
\mathcal{L}_\text{photo} = \left \| \mathbf{C}^\textit{comp}(\mathbf{r}) - \mathbf{C}^\textit{gt}(\mathbf{r}) \right \|_{2}^{2}.
\label{eq:photometric_loss}
\end{equation}
We show in~\figref{plane_refinement} that with joint optimization, our rendered reflection aligns perfectly with the observations.
Upon reaching the 50K iteration, we cease adjustments to $\mathbf{s}$ and proceed to refine the model with the sparse edge regularization loss for an additional 50K iterations:
\begin{equation}
\mathcal{L} = \mathcal{L}_\text{photo} + 0.5\mathcal{L}_\text{edge}.
\label{eq:full_loss}
\end{equation}
The sparse edge regularization helps eliminate false geometries but also diminishes the reconstruction sharpness. We fix the attenuation field to restore detail and gradually decay the weight for sparse edge regularization for another $100K$ iterations.

\subsection{Implementation Details}
\label{sec:details}
We implement our approach using a single NVIDIA V100 GPU.
We use one set of hyperparameters for all experiments.
Our network follows the default InstantNGP design (hash grid + MLP) to predict color and density and uses an additional NGP to predict the attenuation.
We parameterize the scenes using the contraction parameterization \cite{barron2022mip}.
Compared to the vanilla InstantNGP framework which requires 5 hours to complete 150k iterations, our RA-NeRF requires only an additional 2 hours to complete the same number of iterations.
\section{Experimental Results}
\label{sec:experimental_results}
\begin{table}[t]
\caption{
\textbf{Quantitative evaluation on reflection removal.}
We report the average PSNR, SSIM and LPIPS results with comparisons to existing methods on the held-out reflection-free views captured \emph{outside} the rooms.
% reflection-free \emph{outside} split.
% \changil{Use a more descriptive phrase instead of the ``outside subset''? E.g., held-out views captured outside of rooms, hence reflection-free.}
}
\label{tab:quantitative_reflection_removal}
\centering
\resizebox{0.7\linewidth}{!} 
{
\begin{tabular}{l | ccc}
\toprule
& PSNR $\uparrow$ & SSIM $\uparrow$ & LPIPS $\downarrow$ \\
\midrule
NeRFReN~\cite{Guo_2022_CVPR} &  12.27 & 0.3786 & 0.760 \\
MS-NeRF~\cite{Yin_2023_CVPR} &  13.28 & 0.4580 & 0.657 \\
% 
% \midrule
Ours & 
\textbf{15.04} &  \textbf{0.5885} & \textbf{0.446} \\
\bottomrule
\end{tabular}
}
\vspace{-3mm}
\end{table}

\subsection{Dataset}
To address the scarcity of real-world 360-degree datasets that contain pronounced reflections, we present the \emph{Office dataset}.
This dataset contains six distinct 360-degree scenes captured in office environments where reflections are prominently caused by window glass.
We capture each scene from both interior and exterior perspectives.
The dataset is organized into three splits: \emph{inside-train}, \emph{inside-val}, and \emph{outside}.
The \emph{inside-train} and \emph{inside-val} splits feature images with reflections, while the \emph{outside} split consists of images without reflections.

\begin{table*}[t]
\caption{
\textbf{Quantitative evaluation on reconstruction.}
We report the average PSNR, SSIM and LPIPS results with comparisons to existing methods on Office dataset \emph{inside-val} split. The best performance is in \first{bold} and the second best is \second{underscored}.
% We omit PSNR because it is not exposure-agnostic.
}
\label{tab:quantitative_reconstruction}
\centering
\resizebox{\linewidth}{!} 
{
\begin{tabular}{l cccccc|c}
\toprule
PSNR $\uparrow$ / SSIM $\uparrow$ / LPIPS $\downarrow$ & Game Room & Meeting Room 1 & Meeting Room 2 & Meeting Room 3 & Meeting Room 4 & Home Office & Average \\
\midrule
NeRFReN~\cite{Guo_2022_CVPR} & 
 17.81 / 0.5035 / 0.337 &
 19.76 / 0.5871 / 0.431 &
 22.67 / 0.8497 / 0.218 &
 21.30 / 0.8072 / 0.253 &
 \first{24.96} / \second{0.8660} / \second{0.222} &
 14.78 / 0.4465 / 0.592 &
 20.43 / 0.6924 / 0.343 \\
MS-NeRF~\cite{Yin_2023_CVPR} & 
 \second{23.68} / \second{0.8281} / \second{0.173} &
 \first{28.21} / \second{0.8450} / \second{0.196} &
 \second{24.54} / \second{0.8688} / \second{0.189} &
 \first{24.91} / \second{0.8439} / \second{0.213} &
 22.18 / 0.7940 / 0.342 &
 15.89 / \second{0.6396} / \second{0.454} &
 \first{23.20} / \second{0.8010} / \second{0.269} \\
Ref-NeRF~\cite{verbin2022refnerf} & 
 22.46 / 0.7152 / 0.318 &
 \second{25.86} / 0.7572 / 0.307 &
 22.30 / 0.7599 / 0.389 &
 \second{23.28} / 0.7205 / 0.365 &
 24.20 / 0.8054 / 0.314 &
 \second{16.26} / 0.5466 / 0.710 &
 22.39 / 0.7177 / 0.408 \\
Ours & 
 \first{24.45} / \first{0.9140} / \first{0.076} &
 23.14 / \first{0.8847} / \first{0.125} &
 \first{24.69} / \first{0.9218} / \first{0.132} &
 21.99 / \first{0.9106} / \first{0.096} &
 \second{24.66} / \first{0.9351} / \first{0.109} &
 \first{17.47} / \first{0.8196} / \first{0.232} &
 \second{22.58} / \first{0.8962} / \first{0.133} \\
\bottomrule
\end{tabular}
}
\end{table*}
% \begin{figure}[h]
% %
% \newlength\figwidtfGeo
% \setlength\figwidtfGeo{0.248\linewidth}
% \centering%
% \includegraphics[trim=0 0 0 0, clip=true, width=1\linewidth]{fig/accurate_geometry.pdf}\\%
% \vspace{-3.8mm}
% $\underbracket[1pt][2.0mm]{\hspace{0.49\linewidth}}_%
%     {\substack{\vspace{-3.0mm}\\\colorbox{white}
%     {Ours}}}$\vspace{1mm} \hfill
% $\underbracket[1pt][2.0mm]{\hspace{0.49\linewidth}}_%
%     {\substack{\vspace{-3.0mm}\\\colorbox{white}
%     {NeRF}}}$\vspace{1mm}
% \caption{\textbf{Accurate geometry.}
% By explicitly casting reflected rays, our method ensures accurate geometry representation, which is not degraded by the false geometries introduced by reflections.
% }
% \label{fig:better_geometry}
% \end{figure}

\begin{figure}[h]
\centering%
\includegraphics[trim=0 0 0 0, clip=true, width=1\linewidth]{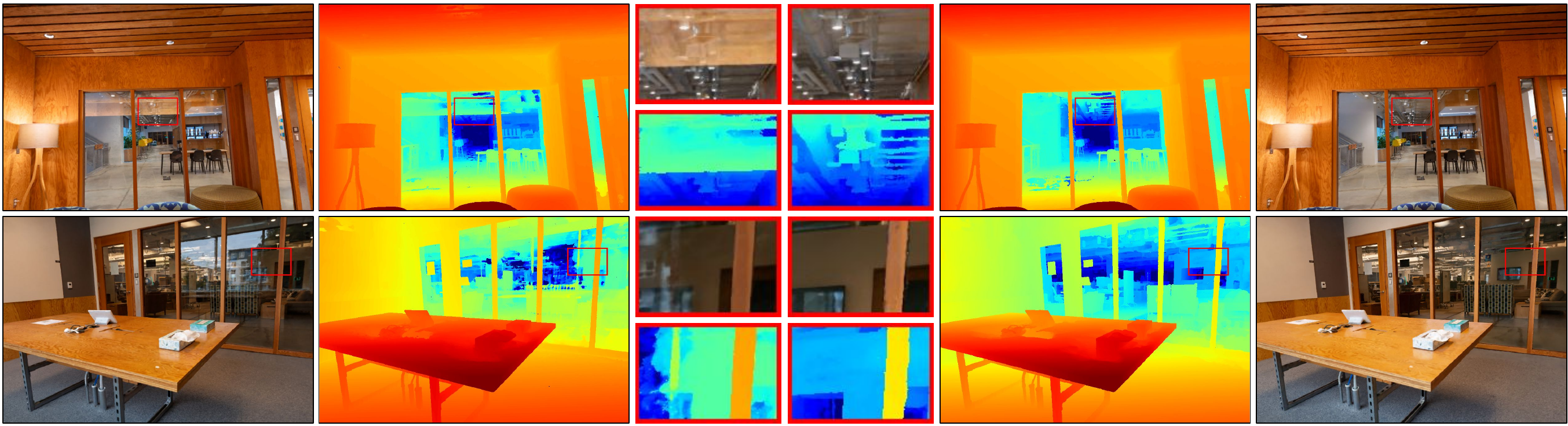}\\%
\mpage{0.48}{$\underbrace{\hspace{\textwidth}}_{\substack{\vspace{-5.0mm}\\\colorbox{white}{~~NeRF~~}}}$}\hfill
\mpage{0.48}{$\underbrace{\hspace{\textwidth}}_{\substack{\vspace{-5.0mm}\\\colorbox{white}{~~Ours~~}}}$}
\vspace{-2mm}
\caption{\textbf{Accurate geometry.}
By explicitly casting reflected rays, our method ensures accurate geometry representation, which is not degraded by the false geometries introduced by reflections.
}
\label{fig:better_geometry}
\end{figure}
\begin{figure}[t]
\newlength\figwidtfComparison
\setlength\figwidtfComparison{0.244\linewidth}
\centering%
\parbox[t]{\figwidtfComparison}{\centering%
  % \fbox{\includegraphics[trim=0 222 400 0, clip=true, width=\figwidtfComparison]{fig/comparison/meeting_room3/NeRFRen_3.jpg}}\\%
  \fbox{\includegraphics[trim=0 0 270 150, clip=true, width=\figwidtfComparison]{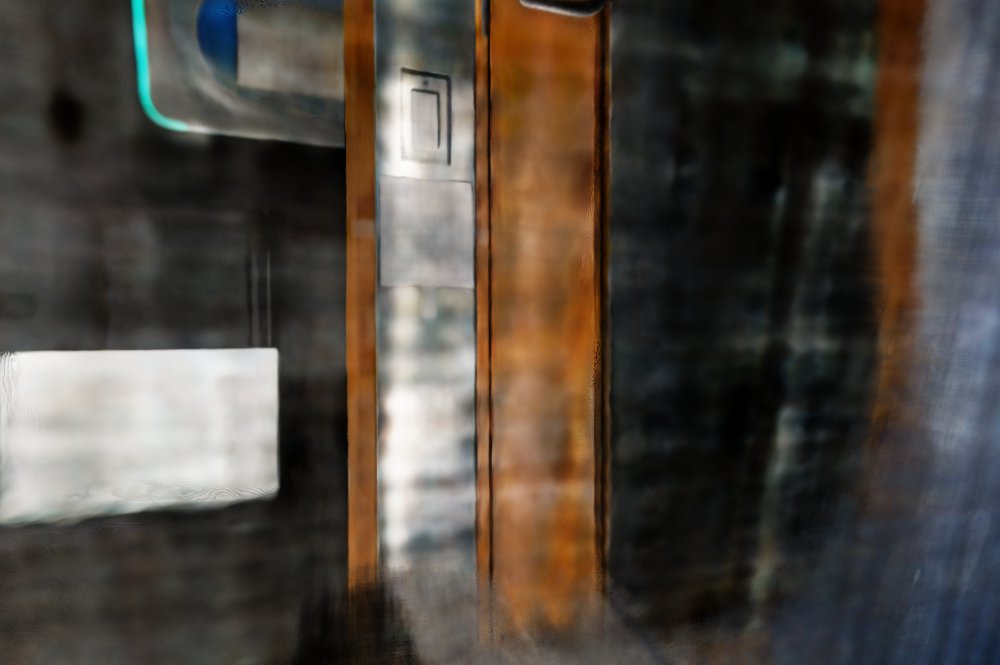}}\\%
  \fbox{\includegraphics[trim=0 0 0 0, clip=true, width=\figwidtfComparison]{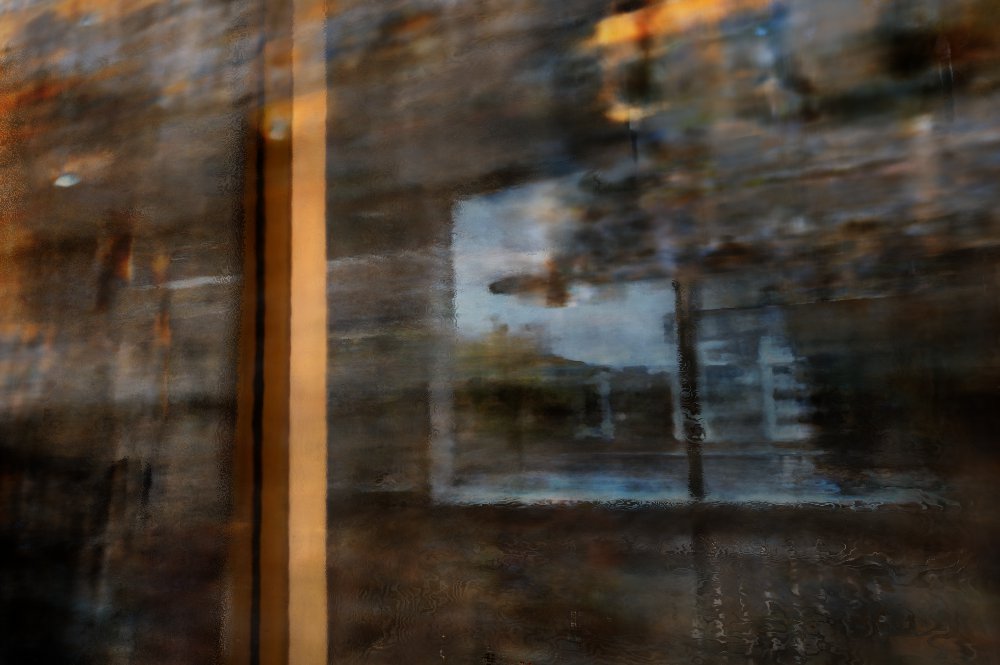}}\\%
  \fbox{\includegraphics[trim=0 0 0 0, clip=true, width=\figwidtfComparison]{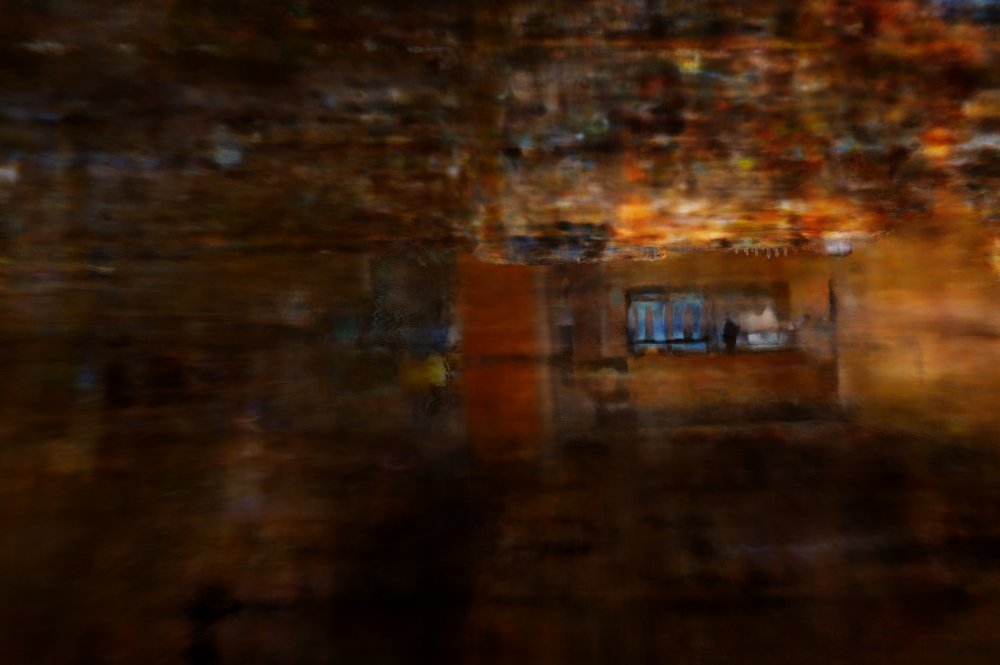}}\\%
   \footnotesize NeRFReN}%
\hfill%
\parbox[t]{\figwidtfComparison}{\centering%
  % \fbox{\includegraphics[trim=0 222 400 0, clip=true, width=\figwidtfComparison]{fig/comparison/meeting_room3/MSNeRF_3.jpg}}\\%
  \fbox{\includegraphics[trim=0 0 270 150, clip=true, width=\figwidtfComparison]{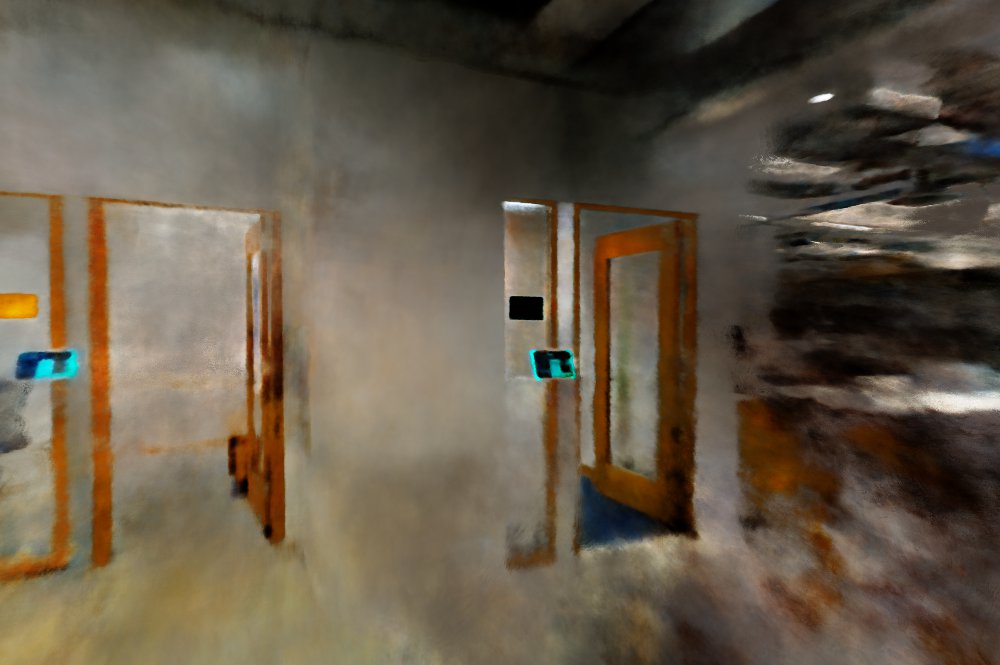}}\\%
  \fbox{\includegraphics[trim=0 0 0 0, clip=true, width=\figwidtfComparison]{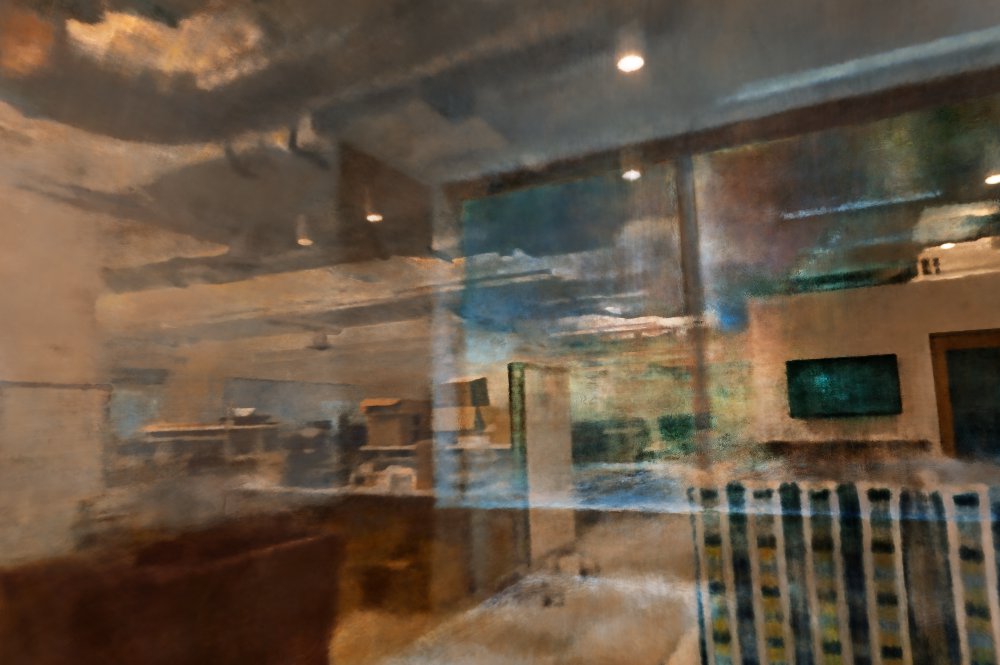}}\\%
  \fbox{\includegraphics[trim=0 0 0 0, clip=true, width=\figwidtfComparison]{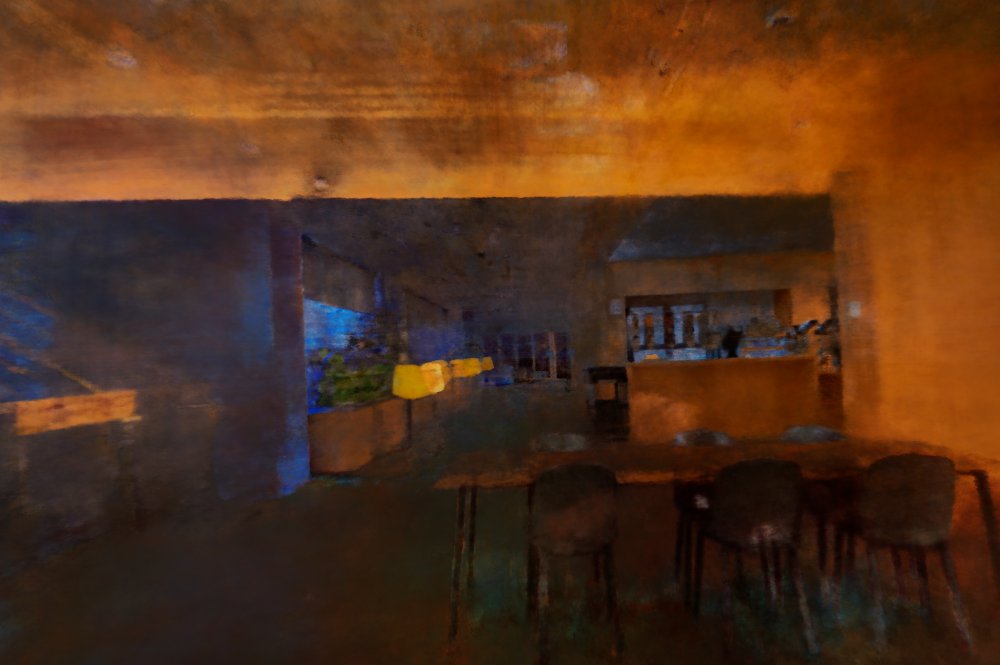}}\\%
   \footnotesize MS-NeRF}%
\hfill%
\parbox[t]{\figwidtfComparison}{\centering%
  % \fbox{\includegraphics[trim=0 222 400 0, clip=true, width=\figwidtfComparison]{fig/comparison/meeting_room3/Ours_3.jpg}}\\%
  \fbox{\includegraphics[trim=0 0 270 150, clip=true, width=\figwidtfComparison]{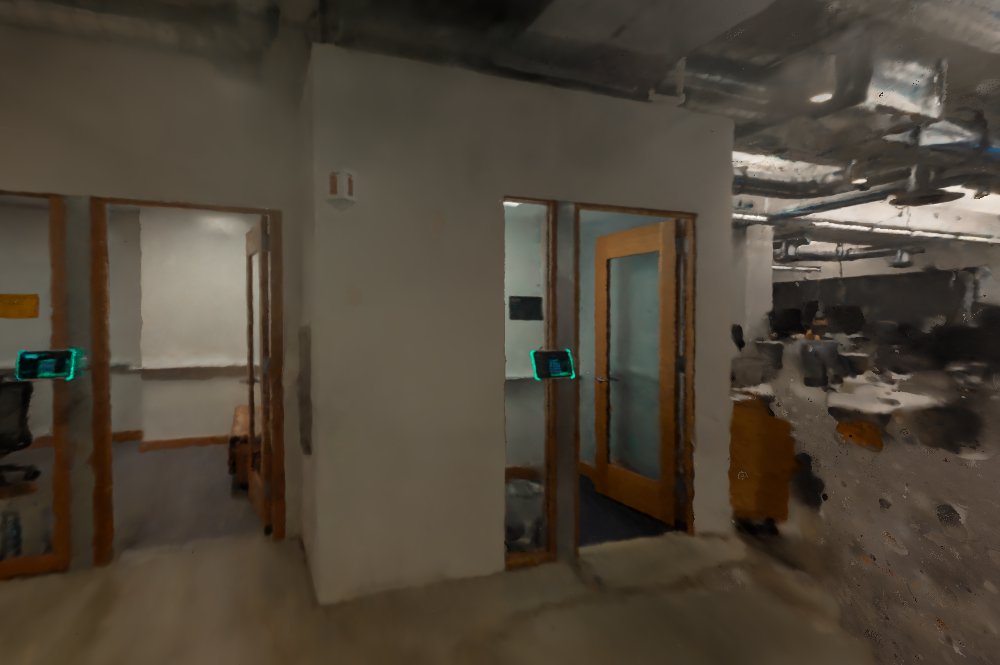}}\\%
  \fbox{\includegraphics[trim=0 0 0 0, clip=true, width=\figwidtfComparison]{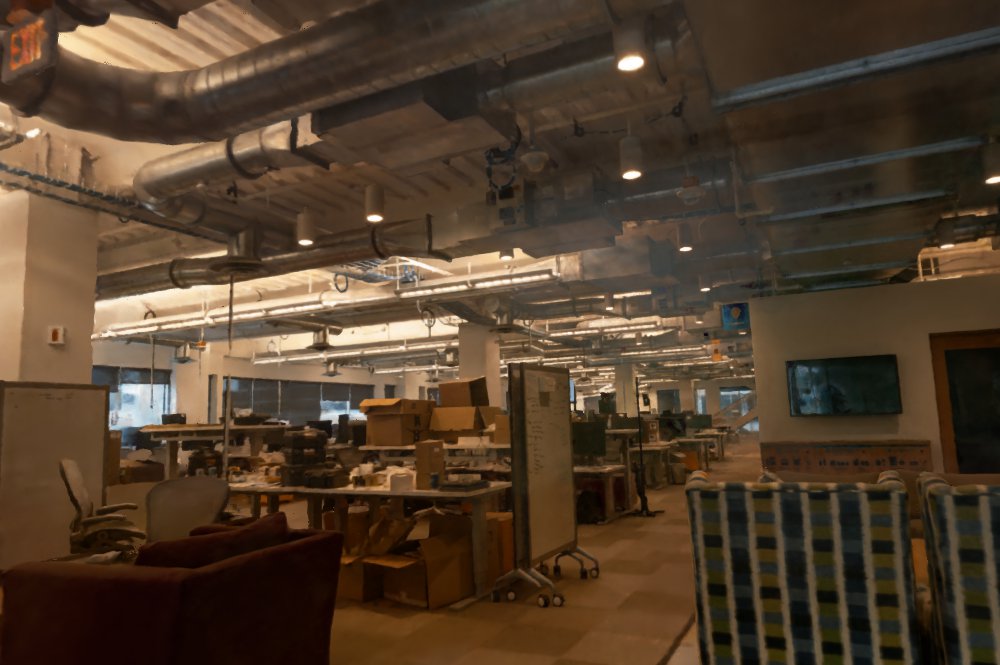}}\\%
  \fbox{\includegraphics[trim=0 0 0 0, clip=true, width=\figwidtfComparison]{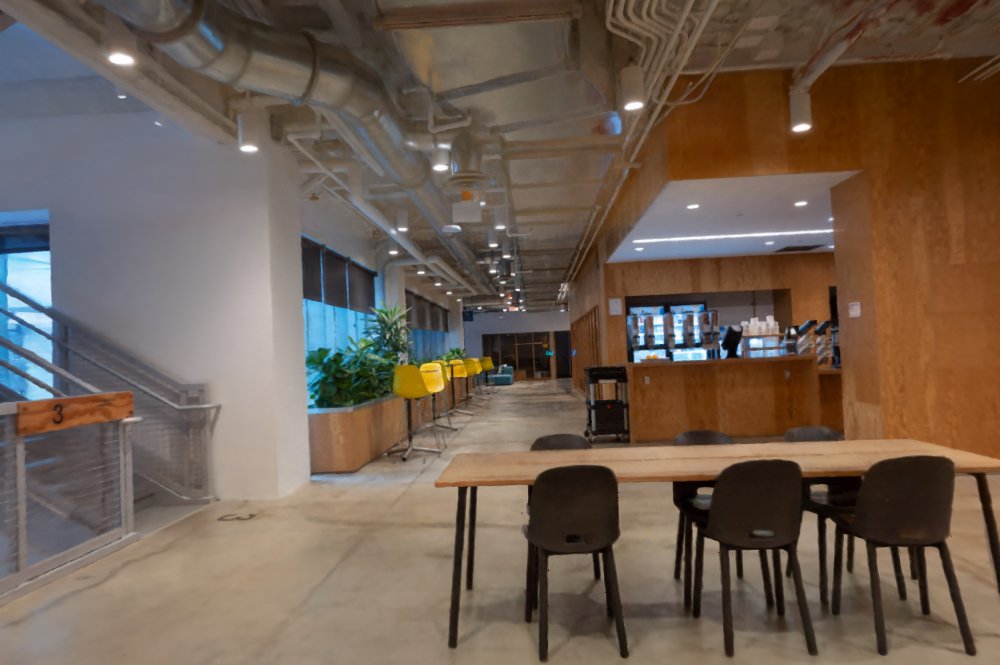}}\\%
   \footnotesize Ours}%
\hfill%
\parbox[t]{\figwidtfComparison}{\centering%
  % \fbox{\includegraphics[trim=0 222 400 0, clip=true, width=\figwidtfComparison]{fig/comparison/meeting_room3/GT_3.jpg}}\\%
  \fbox{\includegraphics[trim=0 0 270 150, clip=true, width=\figwidtfComparison]{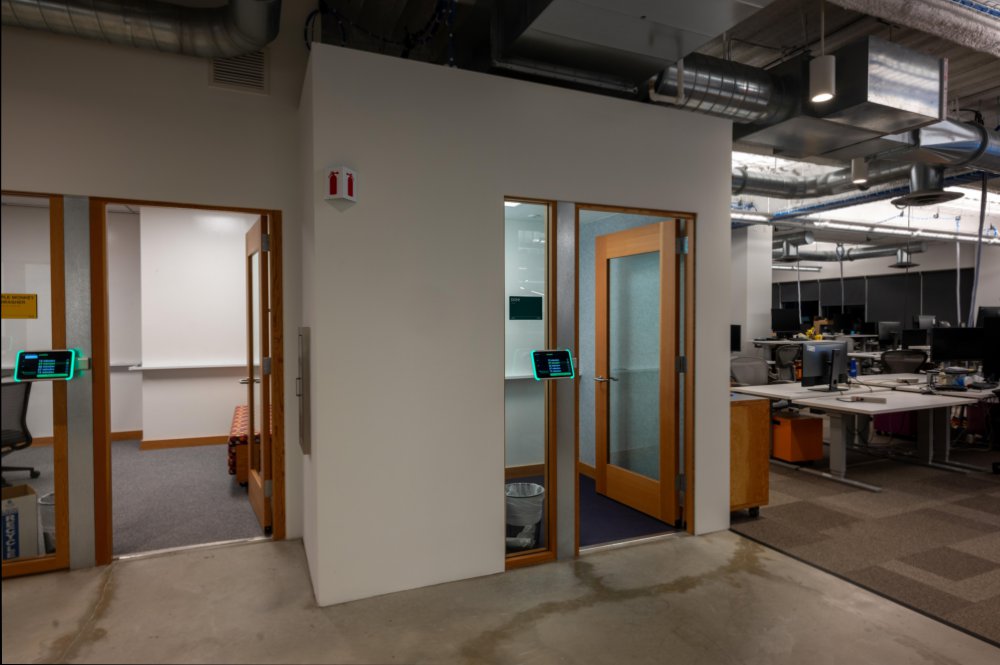}}\\%
  \fbox{\includegraphics[trim=0 0 0 0, clip=true, width=\figwidtfComparison]{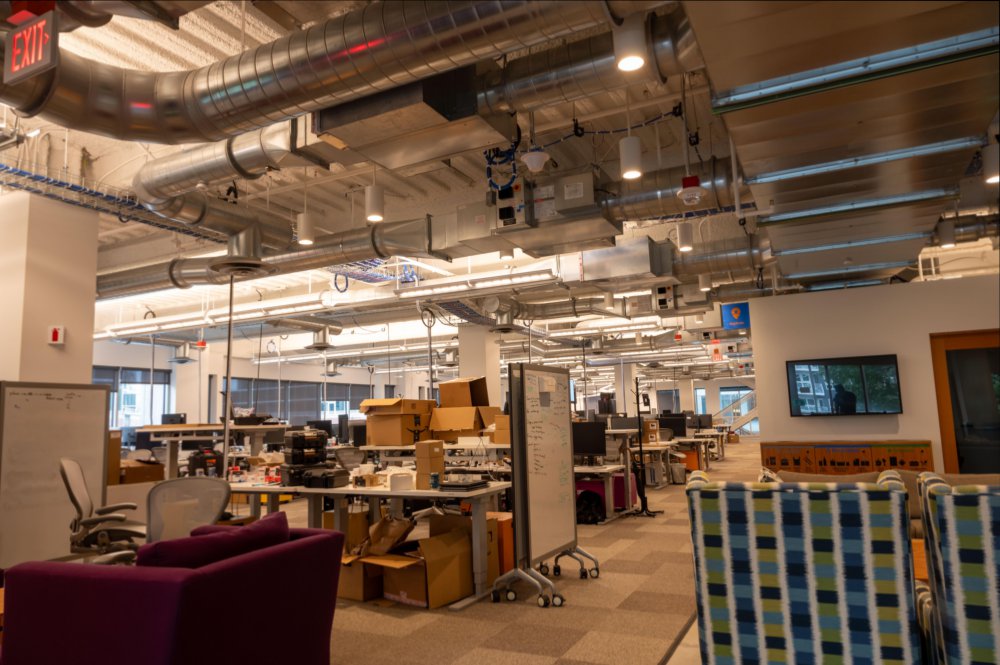}}\\%
  \fbox{\includegraphics[trim=0 0 0 0, clip=true, width=\figwidtfComparison]{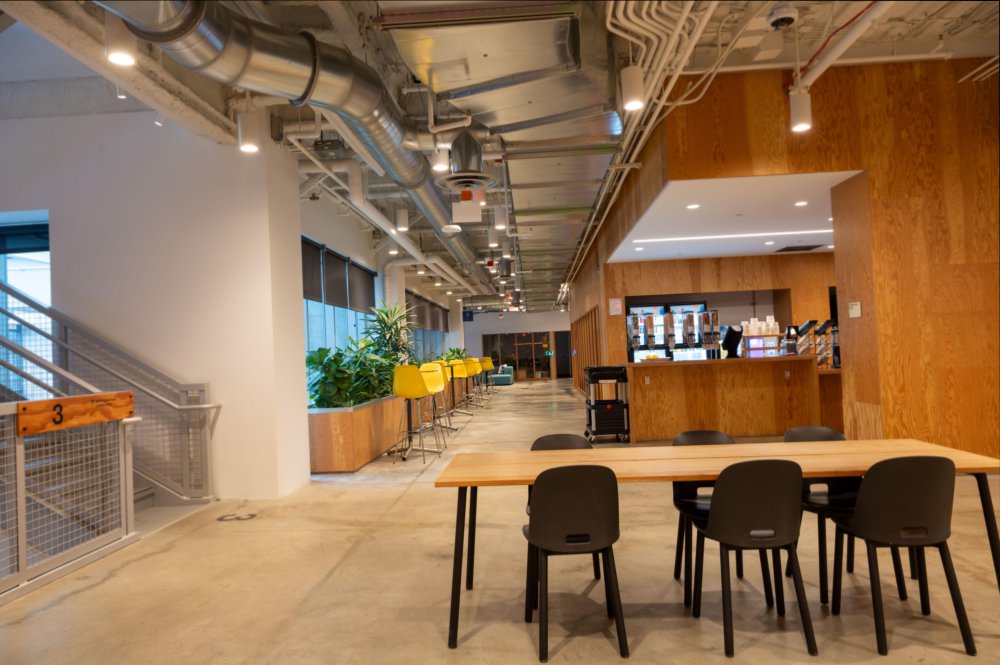}}\\%
   \footnotesize Ground truth}%
% left, bottom, right and top
\vspace{\figcapmargin}
\caption{\textbf{Reflection-removal comparison on the outside split.}
We evaluate the reflection-removal performance of each method using held-out views captured \emph{outside} the rooms.
As the query viewpoints outside the room are far from the training views (inside the room), artifacts are thus visible.
% Notably, all methods are trained on interior views, presenting challenges due to dis-occlusion.
% \changil{What are the occlusion challenges? Add citations to all method names in the figure.}
}
\label{fig:reflection_removal_comparison}
\end{figure}

% \tiny	F-tiny.jpg
% \scriptsize	F-scriptsize.jpg
% \footnotesize	F-footnotesize.jpg
% \small	F-small.jpg
% \normalsize	F-normalsize.jpg
% \large	F-large.jpg
% \Large	F-large2.jpg
% \LARGE	F-large3.jpg
% \huge	F-huge.jpg
% \Huge	F-huge2.jpg
\subsection{Experimental setup}
Our evaluation focuses on two key scenarios: 
1) reflection-free reconstruction and 
2) improved rendering quality (with reflection). 
For evaluating reflection-free reconstruction, we train our model on the \emph{inside-train} split and test on the \emph{outside} split.
For evaluating improved rendering quality, we train on the \emph{inside-train} split and validate on the \emph{inside-val} split.

\begin{figure}
\newlength\figwidtfComparisonIndoor
\setlength\figwidtfComparisonIndoor{0.328\linewidth}
\centering%
\parbox[t]{\figwidtfComparisonIndoor}{\centering%
  \fbox{\includegraphics[trim=0 0 0 0, clip=true, width=\figwidtfComparisonIndoor]{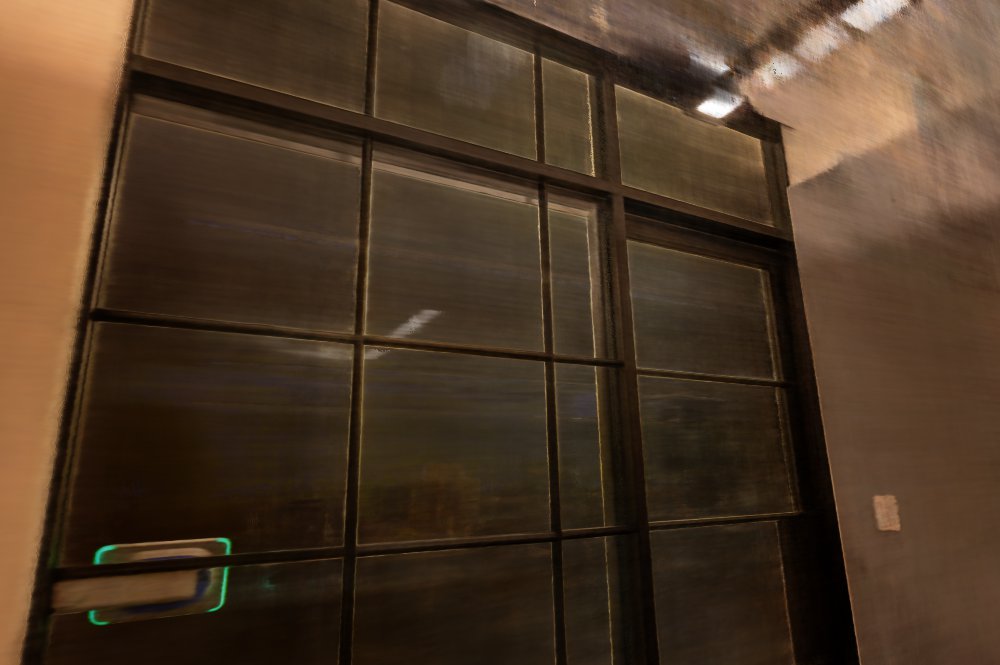}}\\%
  \fbox{\includegraphics[trim=0 0 0 0, clip=true, width=\figwidtfComparisonIndoor]{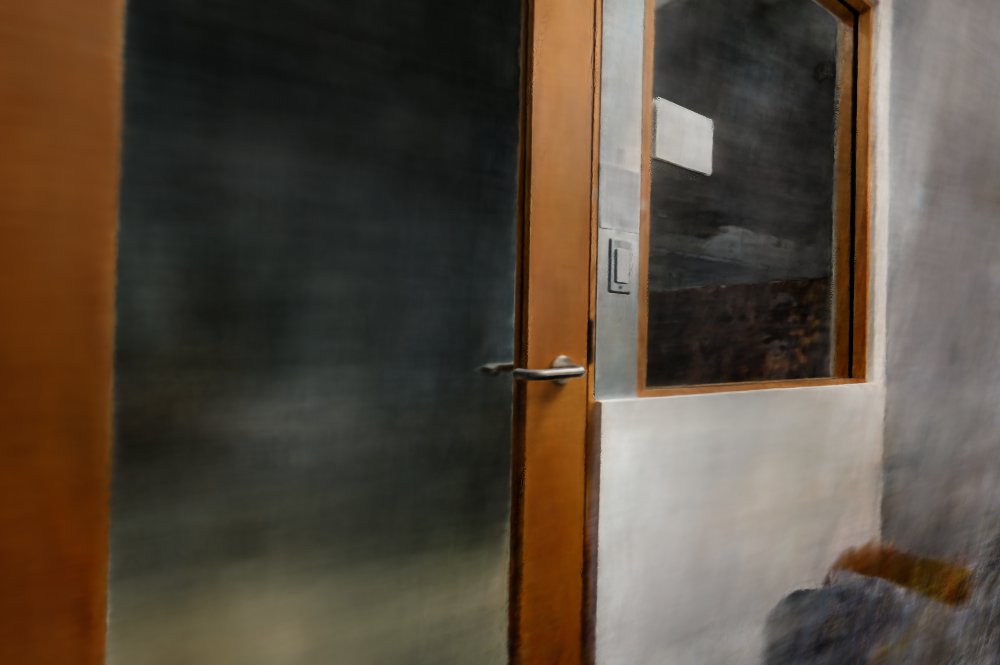}}\\%
  \fbox{\includegraphics[trim=0 0 0 0, clip=true, width=\figwidtfComparisonIndoor]{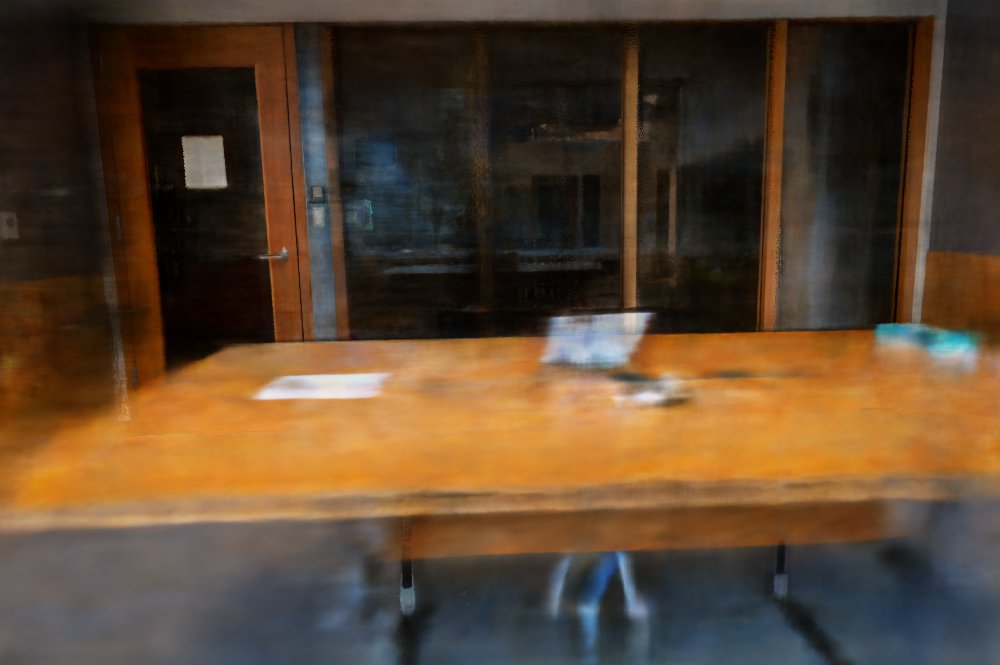}}\\%
  % \fbox{\includegraphics[trim=0 0 0 0, clip=true, width=\figwidtfComparisonIndoor]{fig/comparison/game_room/NeRFRen_3.jpg}}\\%
  \footnotesize NeRFReN\cite{Guo_2022_CVPR}}%
\hfill%
\parbox[t]{\figwidtfComparisonIndoor}{\centering%
  \fbox{\includegraphics[trim=0 0 0 0, clip=true, width=\figwidtfComparisonIndoor]{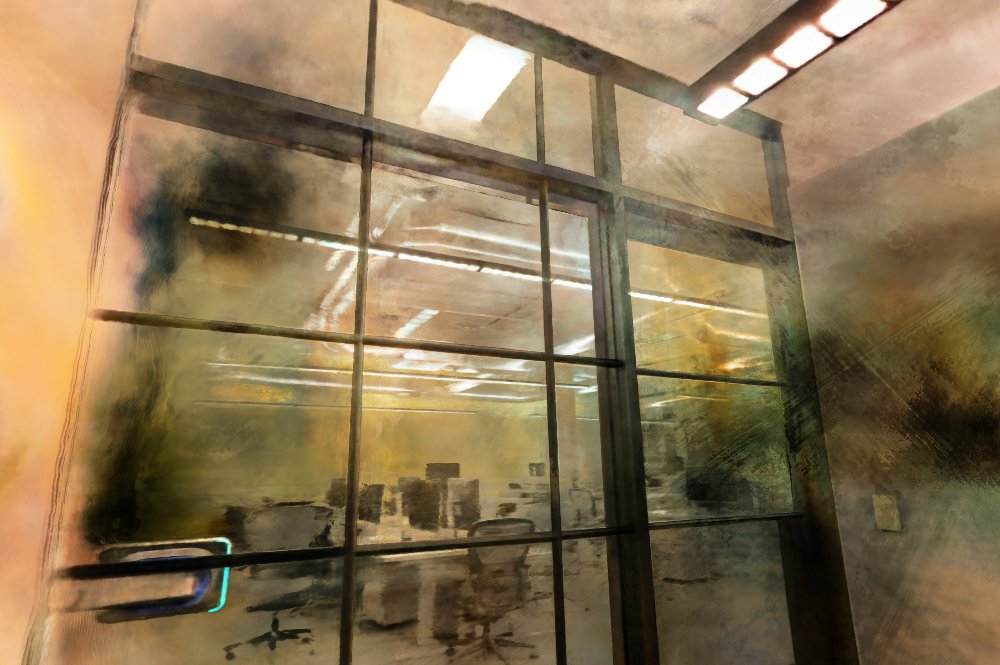}}\\%
  \fbox{\includegraphics[trim=0 0 0 0, clip=true, width=\figwidtfComparisonIndoor]{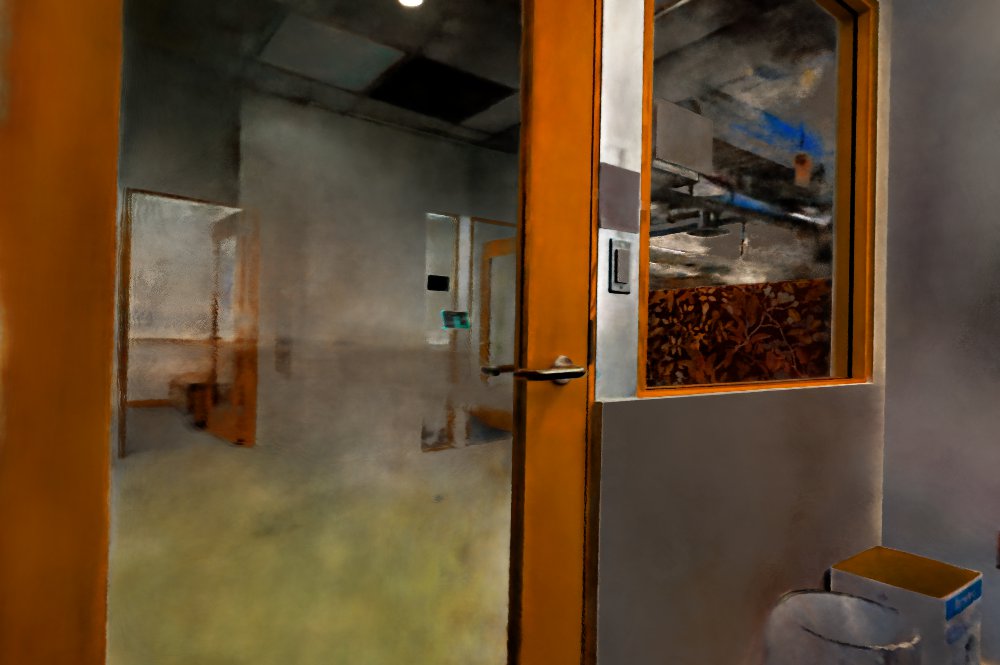}}\\%
  \fbox{\includegraphics[trim=0 0 0 0, clip=true, width=\figwidtfComparisonIndoor]{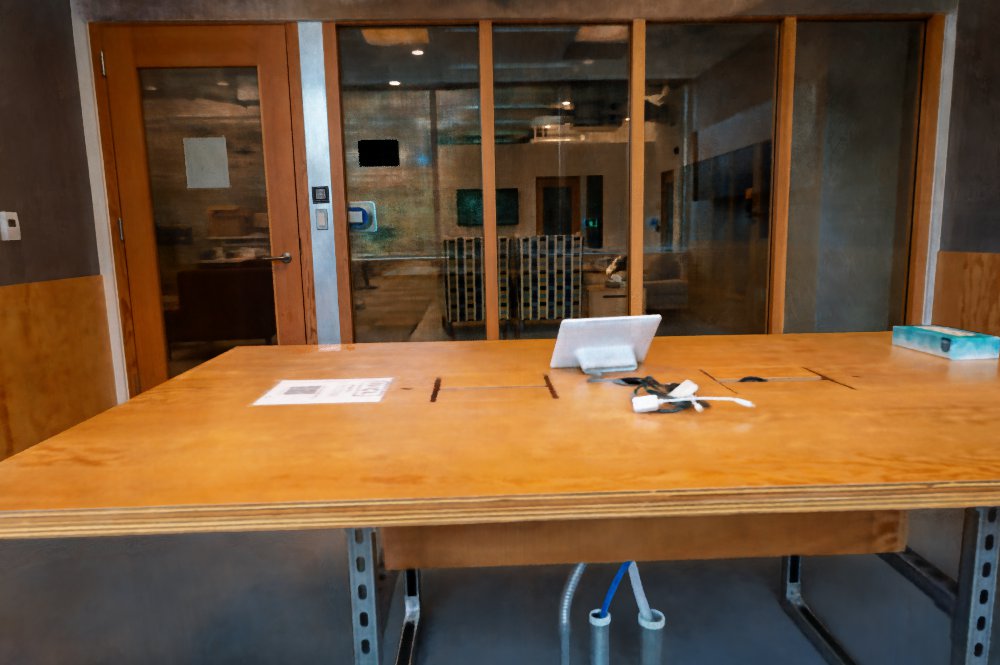}}\\%
  % \fbox{\includegraphics[trim=0 0 0 0, clip=true, width=\figwidtfComparisonIndoor]{fig/comparison/game_room/MSNeRF_3.jpg}}\\%
  \footnotesize MS-NeRF\cite{Yin_2023_CVPR}}%
\hfill%
\parbox[t]{\figwidtfComparisonIndoor}{\centering%
  \fbox{\includegraphics[trim=0 0 0 0, clip=true, width=\figwidtfComparisonIndoor]{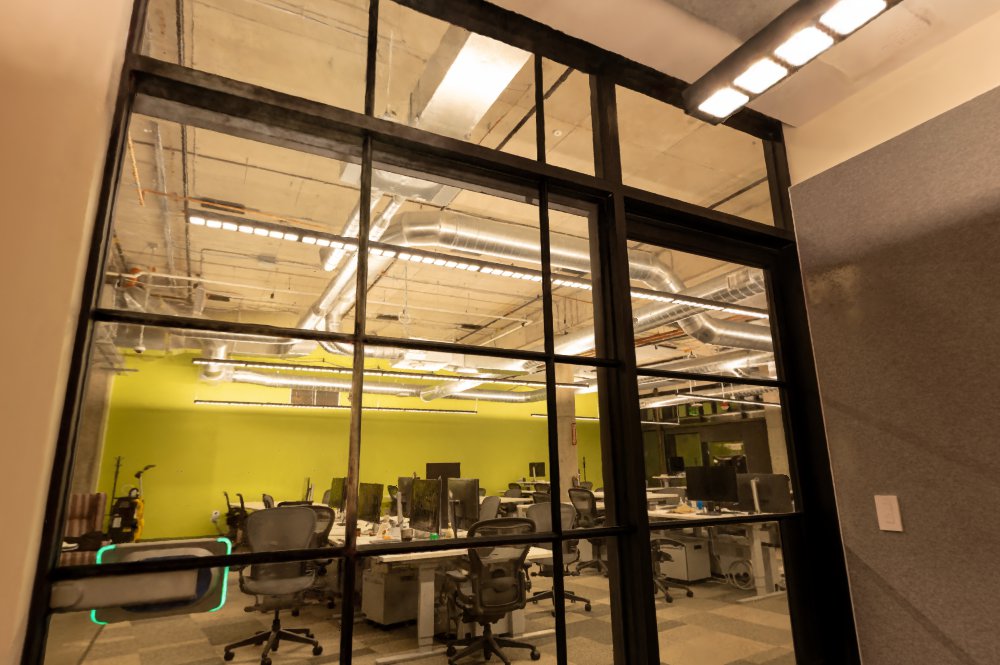}}\\%
  \fbox{\includegraphics[trim=0 0 0 0, clip=true, width=\figwidtfComparisonIndoor]{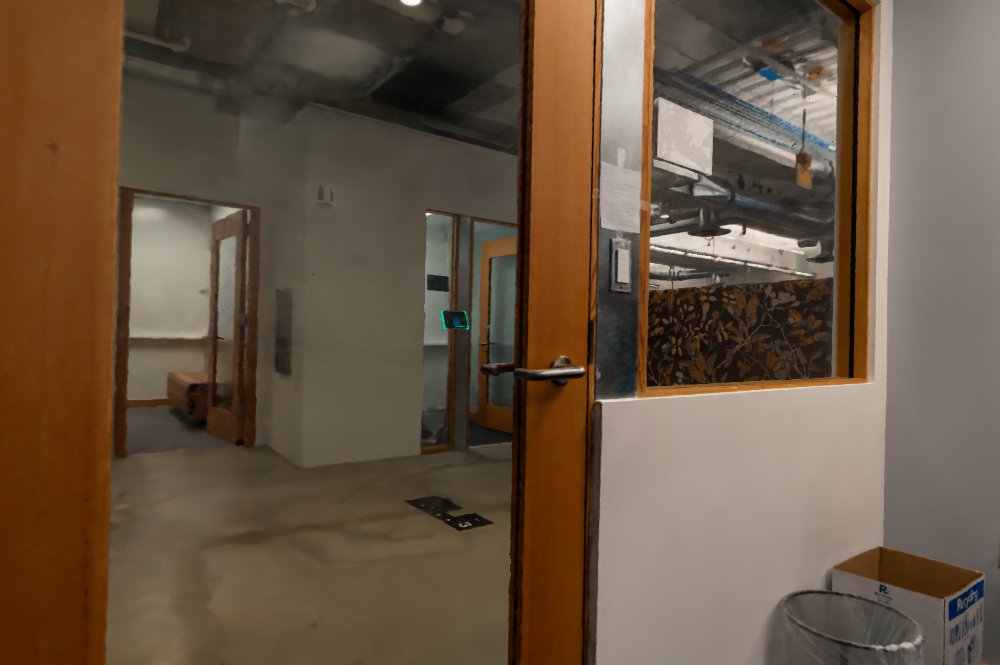}}\\%
  \fbox{\includegraphics[trim=0 0 0 0, clip=true, width=\figwidtfComparisonIndoor]{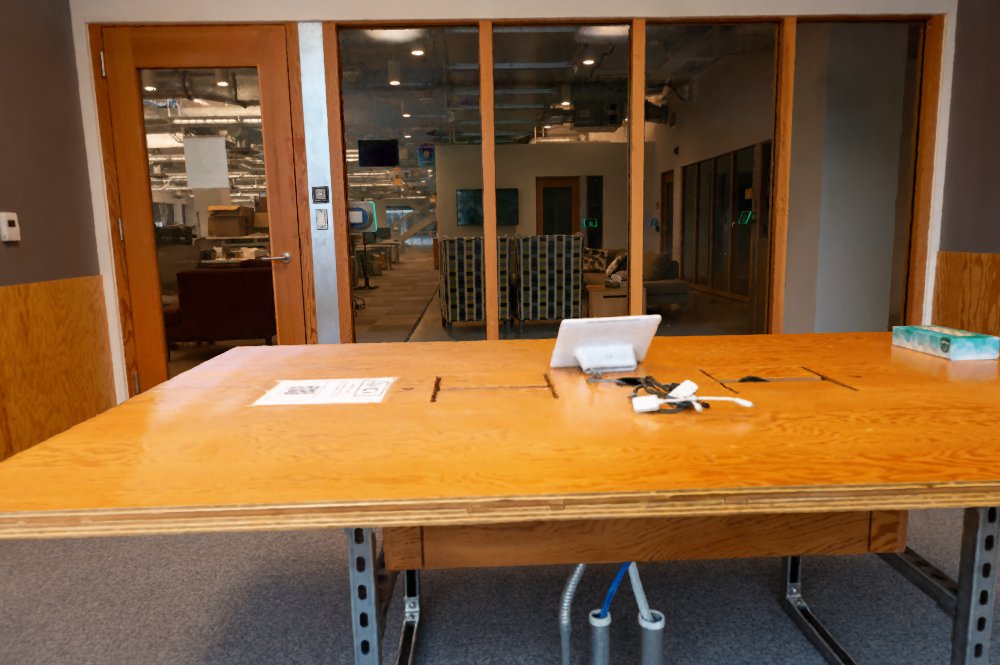}}\\%
  % \fbox{\includegraphics[trim=0 0 0 0, clip=true, width=\figwidtfComparisonIndoor]{fig/comparison/game_room/Ours_3.jpg}}\\%
  \footnotesize Ours}%
% left, bottom, right and top
\vspace{\figcapmargin}
\caption{\textbf{Reflection-removal comparison on the inside-val split.}
We qualitatively evaluate the reflection-removal performance of each method using held-out views captured \emph{inside} the rooms.
}
\label{fig:reflection_removal_comparison_indoor}
\end{figure}

\topic{Reflection-free reconstruction.}
We qualitatively assess the capacity of our reflection-aware NeRF for reflection-free reconstruction using the \emph{inside-val} split in \figref{decomposition}.
This process involves rendering along the primary ray to produce a reflection-free image and along the reflected ray to capture the source of the reflection.
The final image is composed by blending these two images using the predicted attenuation and plane intersection transmittance as in Eq. (\ref{eq:rendering_composed}). 
The results demonstrate our method's capability in synthesizing \emph{reflection-free} views.
We show in \figref{better_geometry} that this leads to more accurate scene geometry reconstruction.
We also compare our method's reflection removal performance against NeRFReN \cite{Guo_2022_CVPR} and MS-NeRF \cite{Yin_2023_CVPR} on the held-out \emph{outside} views in \tabref{quantitative_reflection_removal} and \figref{reflection_removal_comparison}.
However, as the query viewpoints outside the room are far from the training views (inside the room), artifacts are therefore visible.
Therefore, we also conduct a qualitative comparison on the held-out \emph{inside-val} views in \figref{reflection_removal_comparison_indoor} (no reflection-free ground truth is available).
We used the official code of MS-NeRF for scene decomposition.
Since the decomposition is unsupervised, we had to manually select the most suitable decoded images to represent the reflection-free results.
NeRFReN was designed for forward-facing scenes.
To train a NeRFReN model, we constructed a subset of \emph{inside-train}, which only contains forward-facing views.

\begin{figure*}[t]
\newlength\figwidtfComparisonR
\setlength\figwidtfComparisonR{0.196\linewidth}
\centering%
\parbox[t]{\figwidtfComparisonR}{\centering%
  \fbox{\includegraphics[trim=0 0 0 0, clip=true, width=\figwidtfComparisonR]{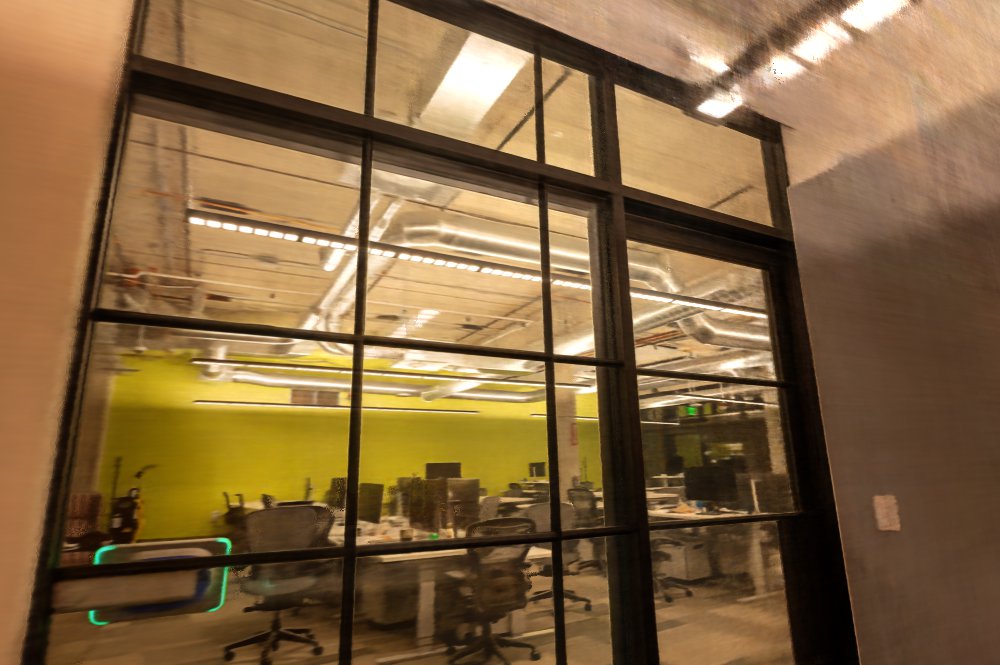}}\\%
  \fbox{\includegraphics[trim=0 0 0 0, clip=true, width=\figwidtfComparisonR]{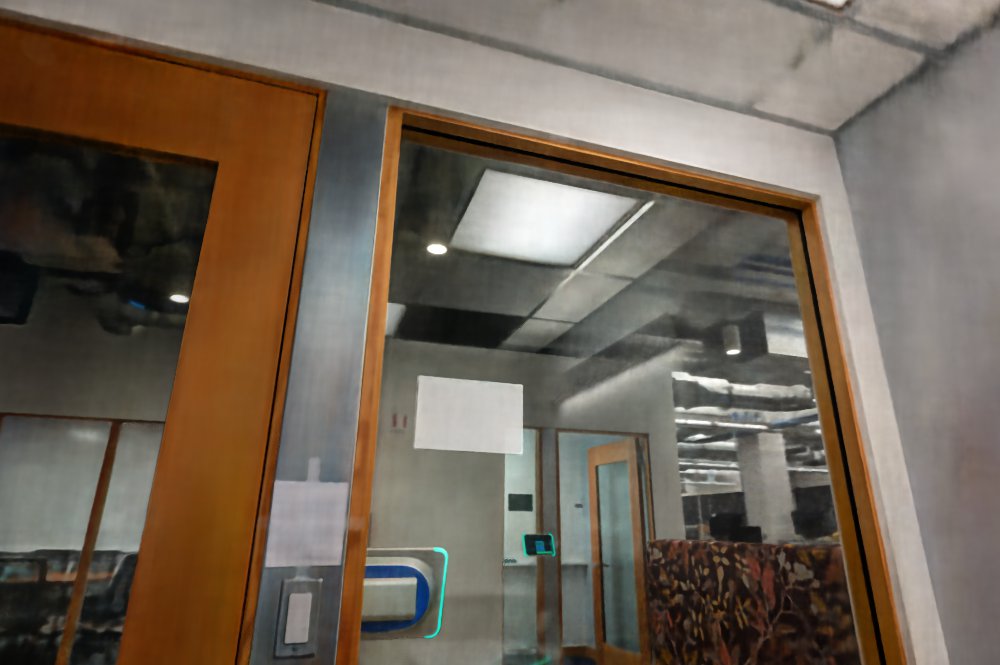}}\\%
  \fbox{\includegraphics[trim=0 0 0 0, clip=true, width=\figwidtfComparisonR]{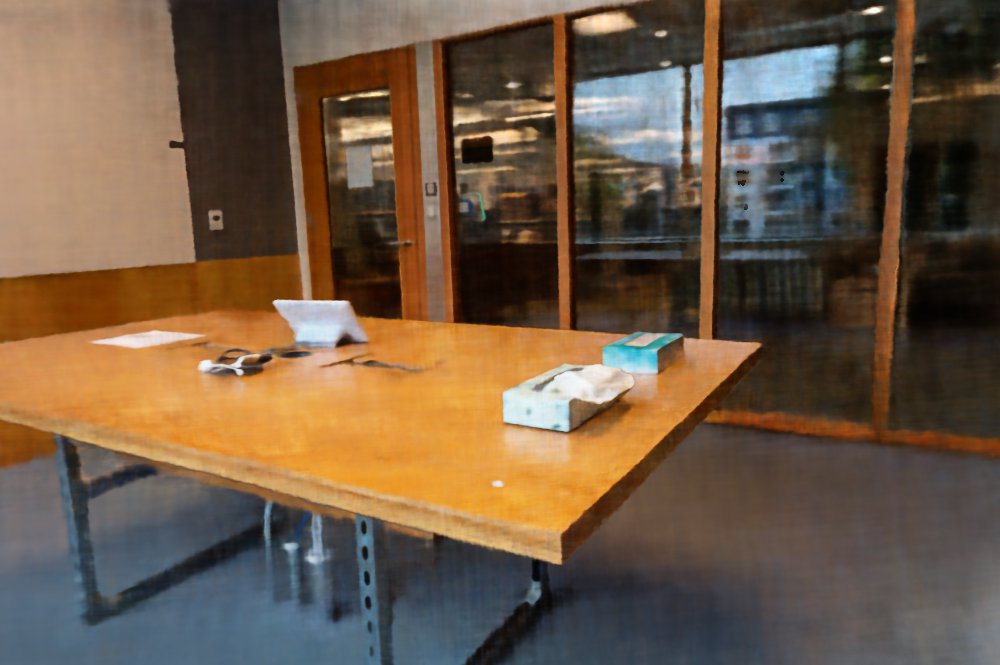}}\\%
  \fbox{\includegraphics[trim=0 0 0 0, clip=true, width=\figwidtfComparisonR]{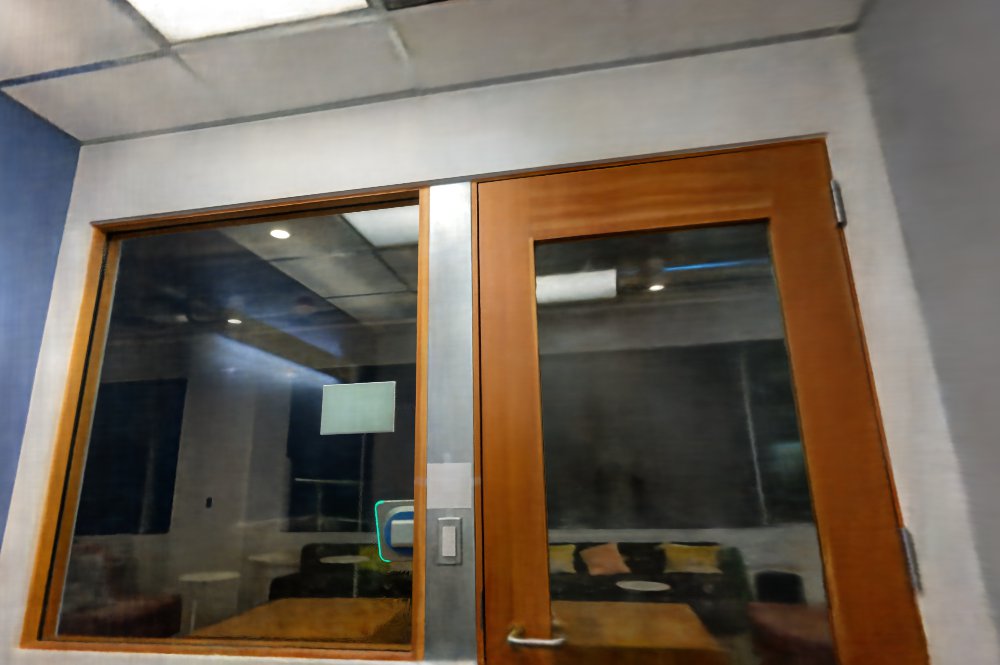}}\\%
  \fbox{\includegraphics[trim=0 0 0 0, clip=true, width=\figwidtfComparisonR]{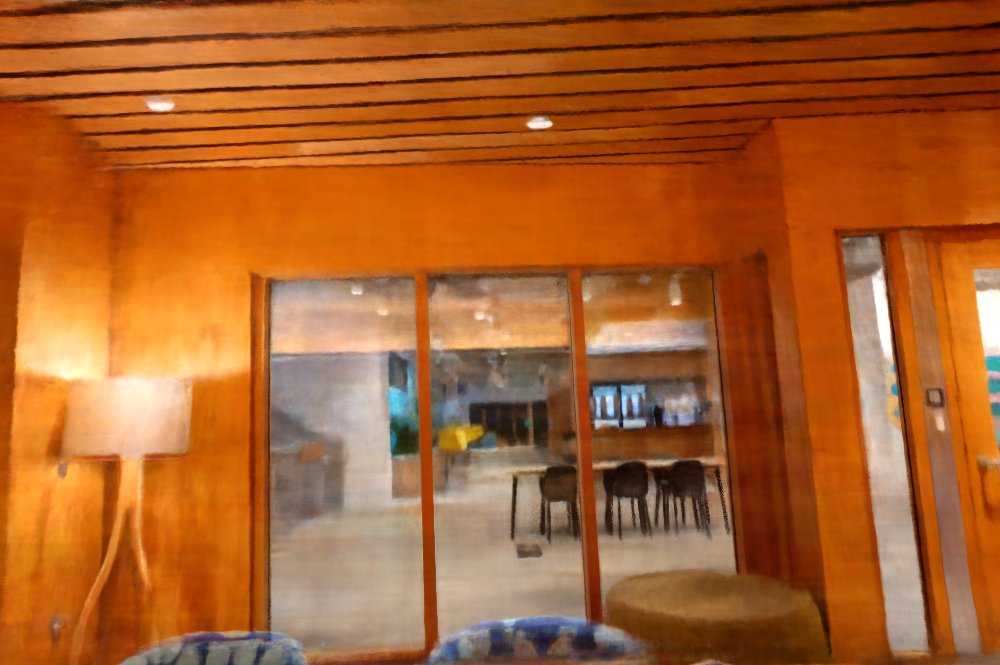}}\\%
  \fbox{\includegraphics[trim=0 0 0 0, clip=true, width=\figwidtfComparisonR]{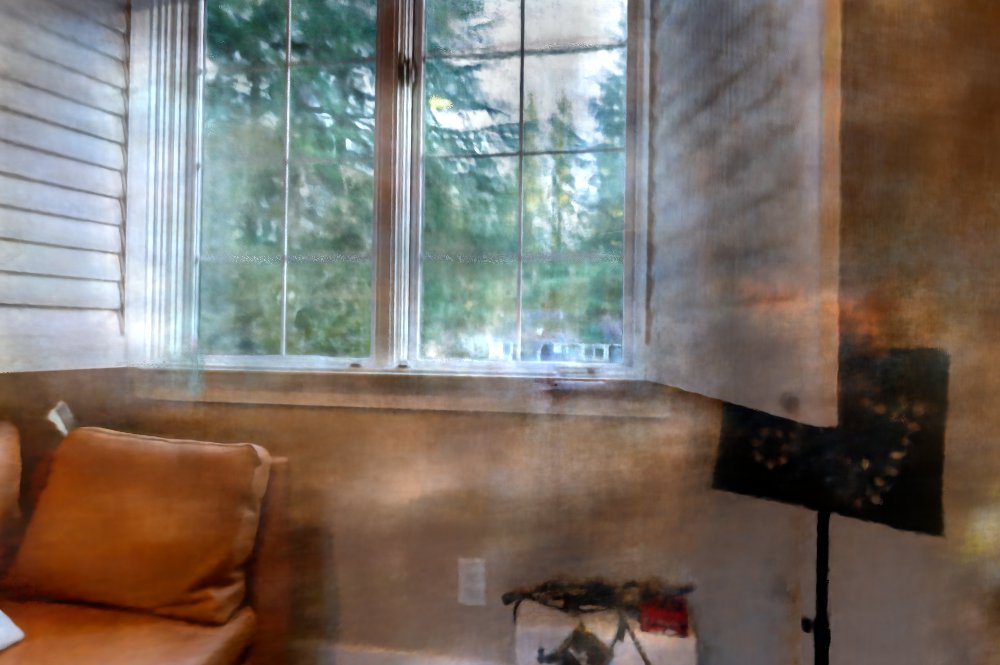}}\\%
   \small NeRFReN~\cite{Guo_2022_CVPR}}%
\hfill%
\parbox[t]{\figwidtfComparisonR}{\centering%
  \fbox{\includegraphics[trim=0 0 0 0, clip=true, width=\figwidtfComparisonR]{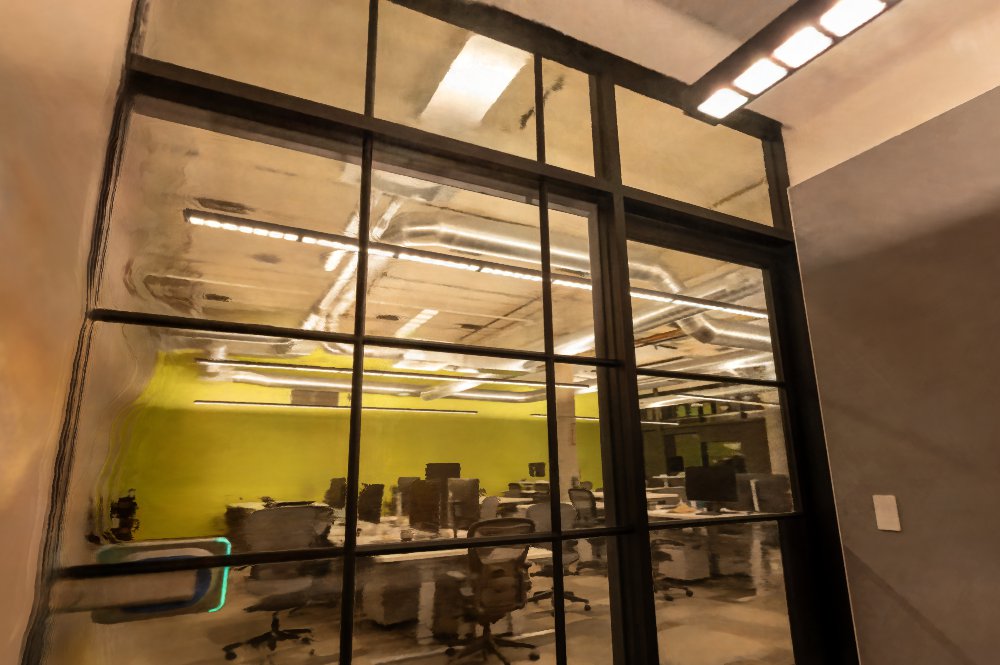}}\\%
  \fbox{\includegraphics[trim=0 0 0 0, clip=true, width=\figwidtfComparisonR]{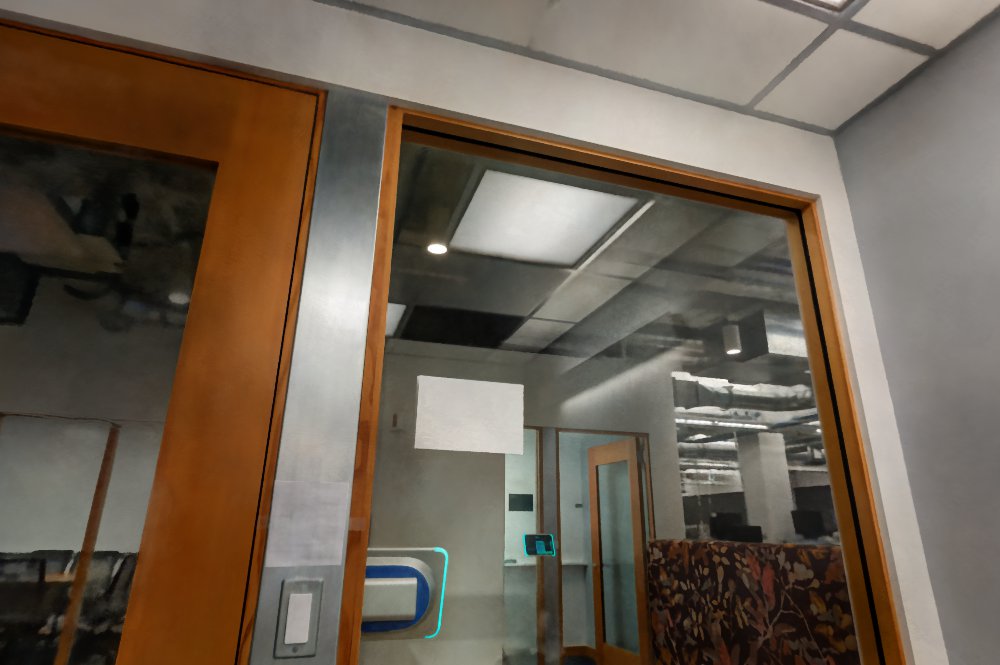}}\\%
  \fbox{\includegraphics[trim=0 0 0 0, clip=true, width=\figwidtfComparisonR]{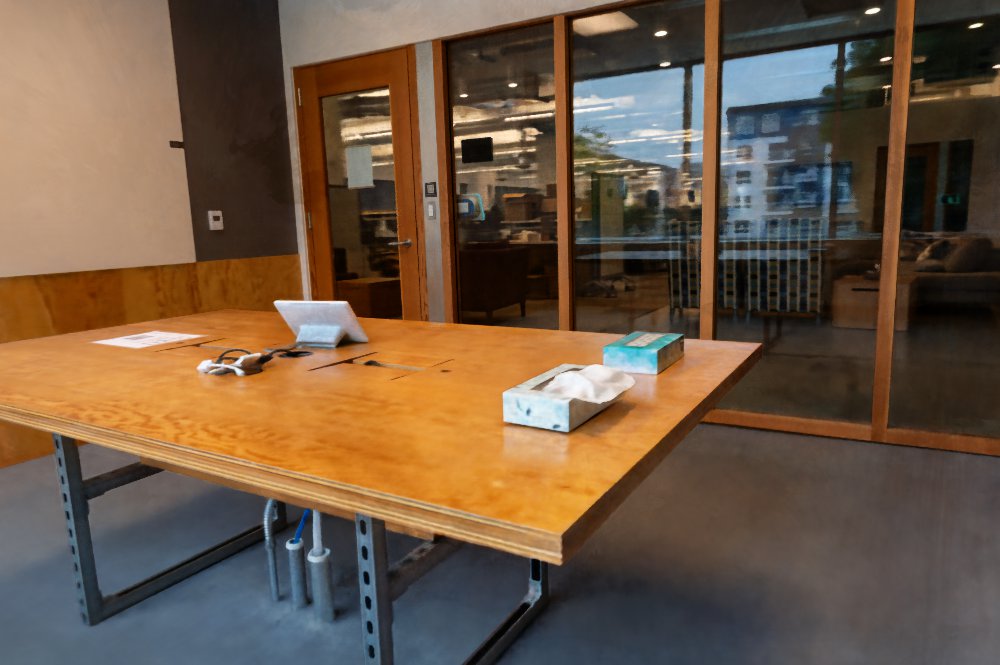}}\\%
  \fbox{\includegraphics[trim=0 0 0 0, clip=true, width=\figwidtfComparisonR]{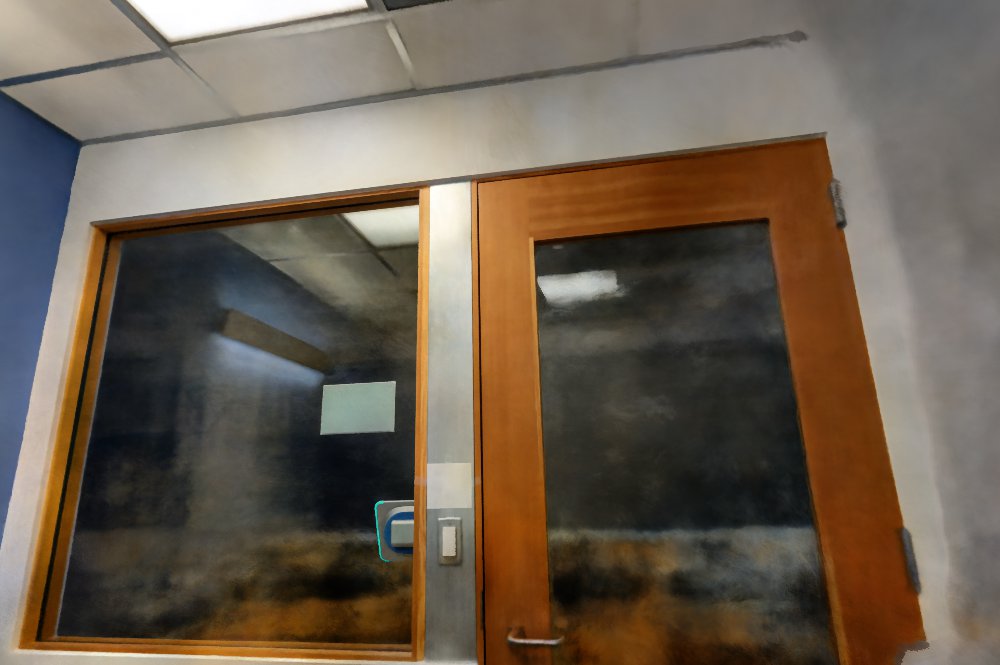}}\\%
  \fbox{\includegraphics[trim=0 0 0 0, clip=true, width=\figwidtfComparisonR]{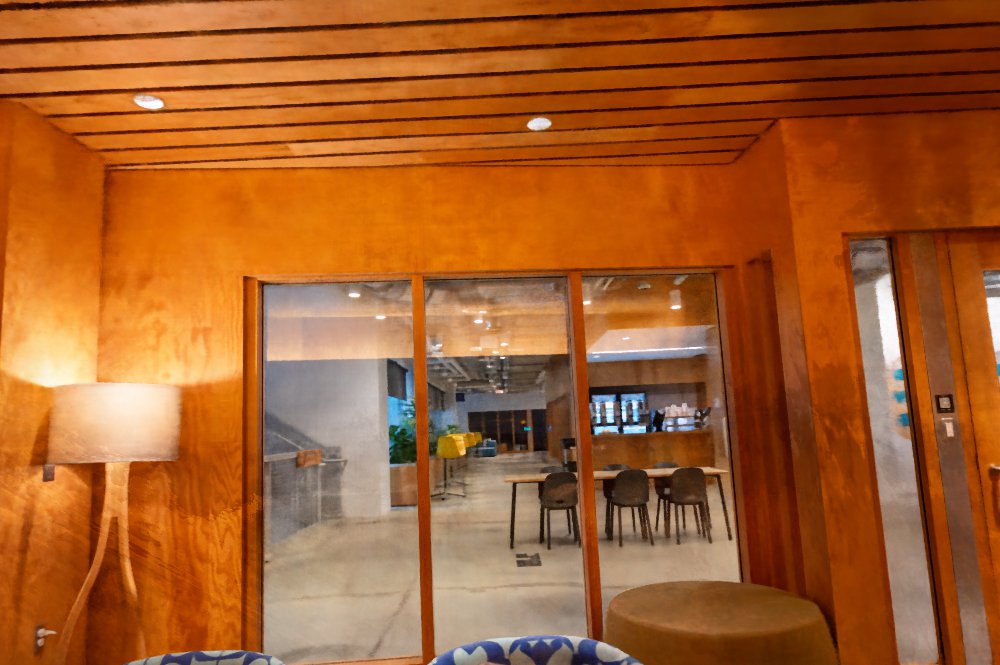}}\\%
  \fbox{\includegraphics[trim=0 0 0 0, clip=true, width=\figwidtfComparisonR]{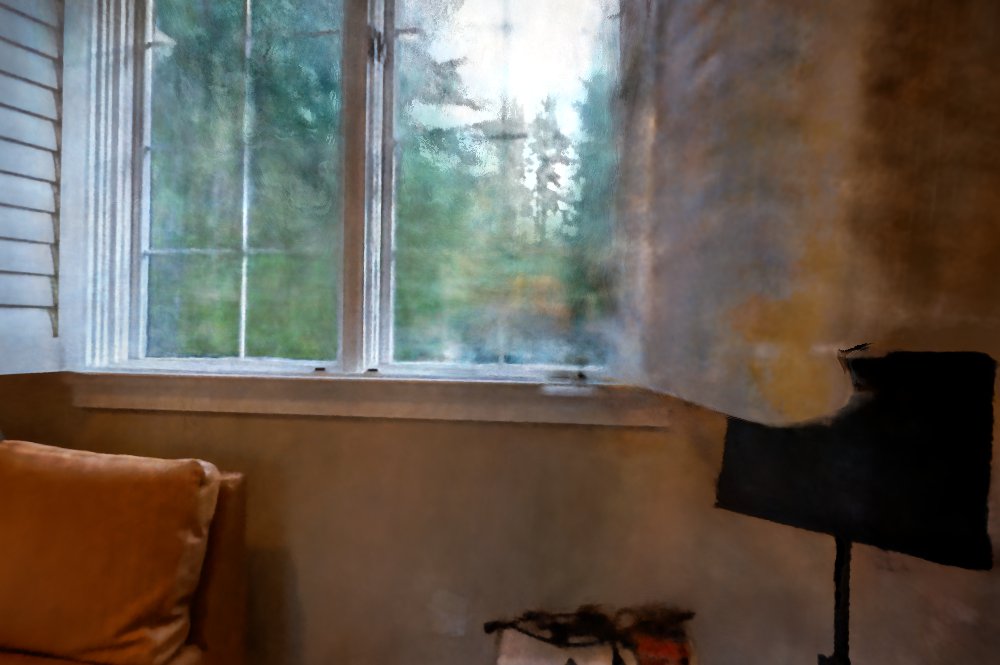}}\\%
   \small MS-NeRF~\cite{Yin_2023_CVPR}}%
\hfill%
\parbox[t]{\figwidtfComparisonR}{\centering%
  \fbox{\includegraphics[trim=0 0 0 0, clip=true, width=\figwidtfComparisonR]{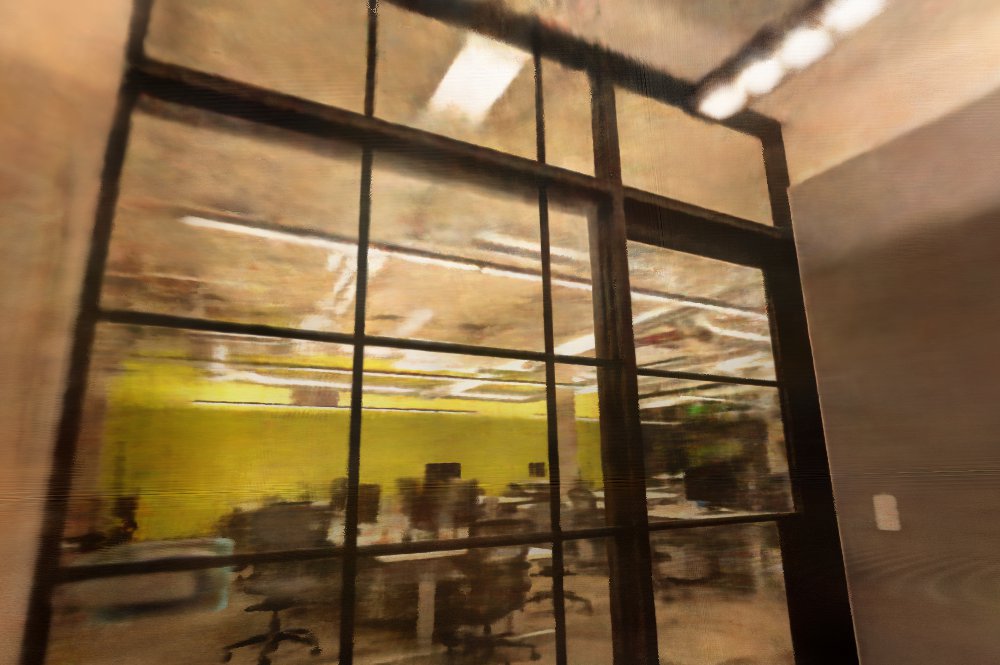}}\\%
  \fbox{\includegraphics[trim=0 0 0 0, clip=true, width=\figwidtfComparisonR]{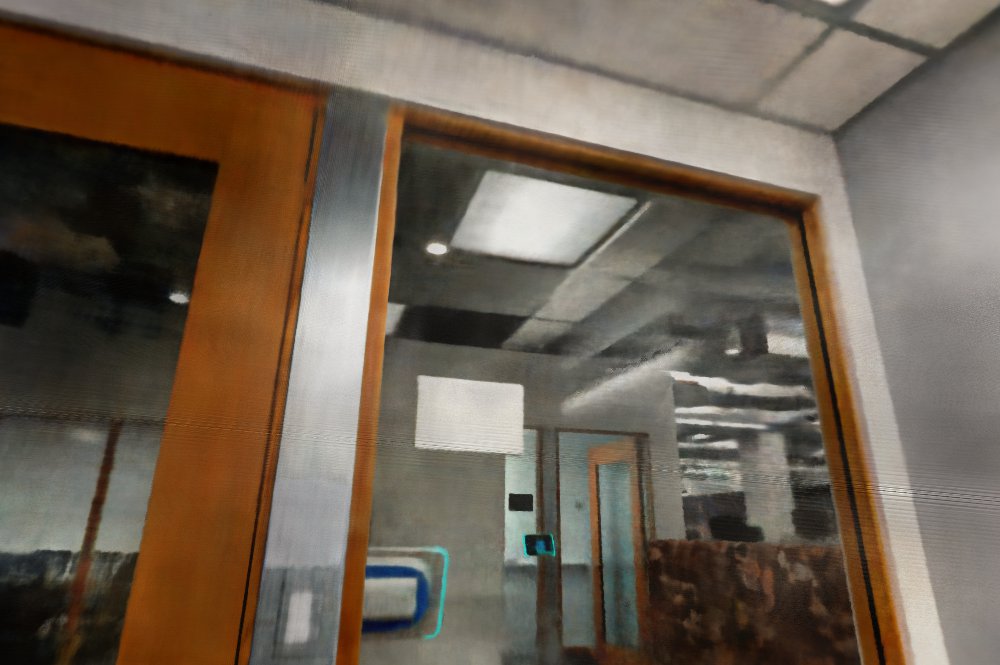}}\\%
  \fbox{\includegraphics[trim=0 0 0 0, clip=true, width=\figwidtfComparisonR]{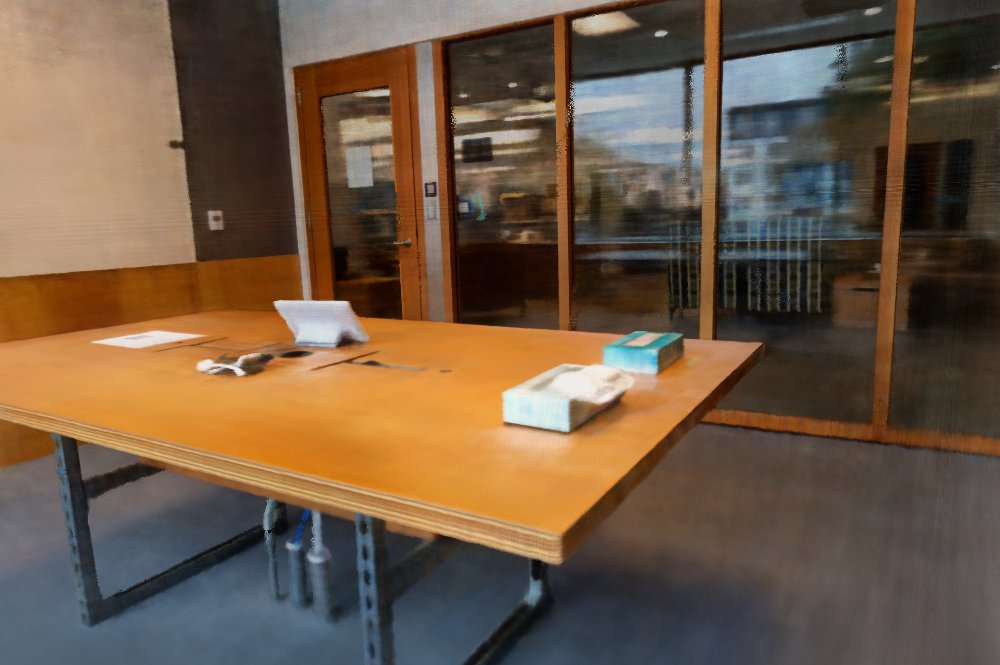}}\\%
  \fbox{\includegraphics[trim=0 0 0 0, clip=true, width=\figwidtfComparisonR]{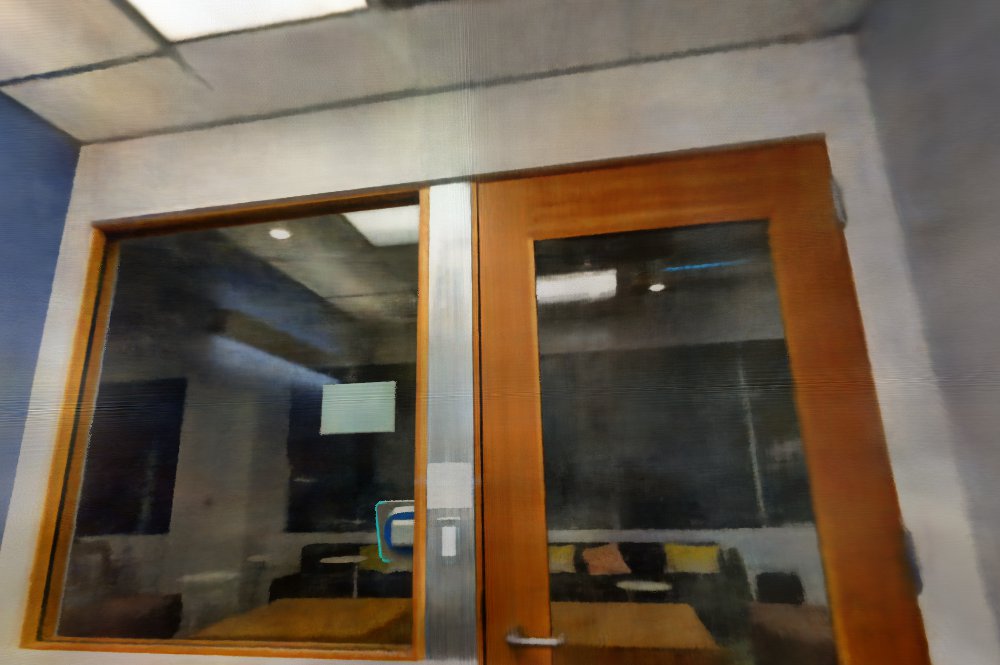}}\\%
  \fbox{\includegraphics[trim=0 0 0 0, clip=true, width=\figwidtfComparisonR]{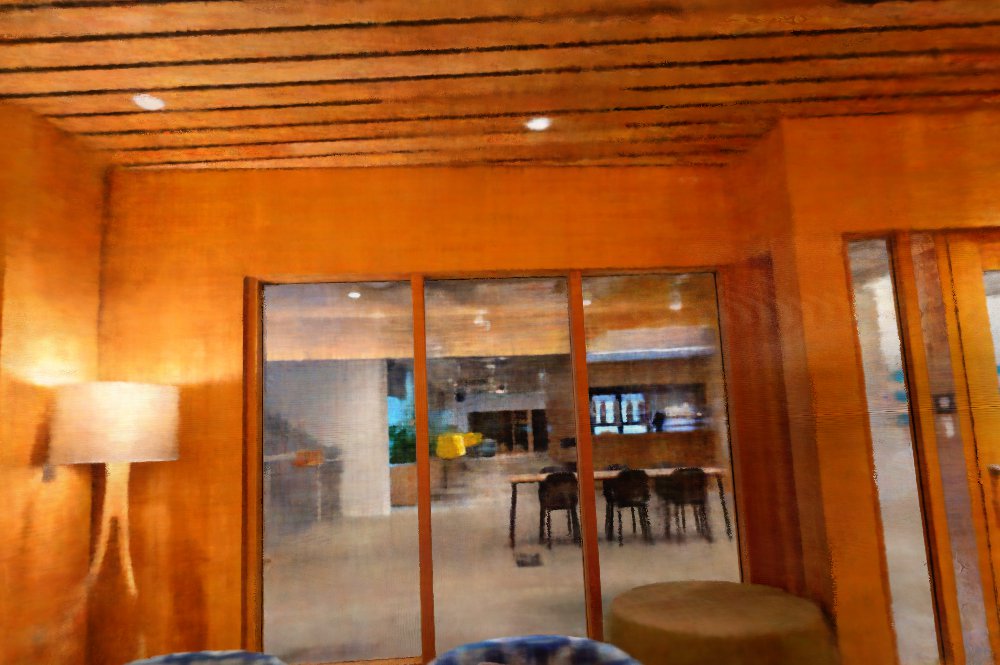}}\\%
  \fbox{\includegraphics[trim=0 0 0 0, clip=true, width=\figwidtfComparisonR]{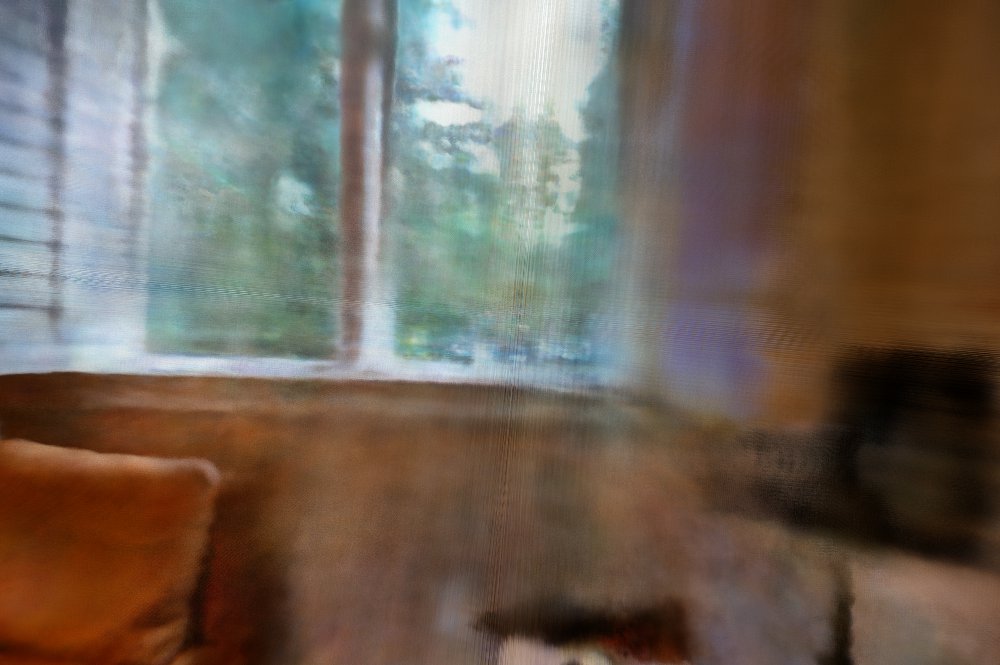}}\\%
   \small Ref-NeRF~\cite{verbin2022refnerf}}%
\hfill%
\parbox[t]{\figwidtfComparisonR}{\centering%
  \fbox{\includegraphics[trim=0 0 0 0, clip=true, width=\figwidtfComparisonR]{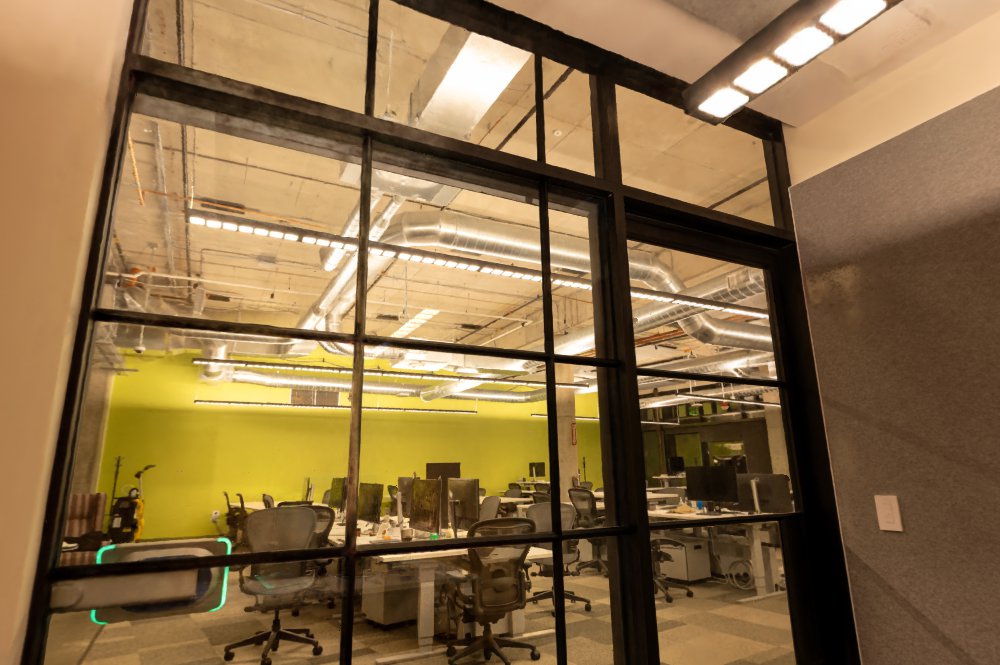}}\\%
  \fbox{\includegraphics[trim=0 0 0 0, clip=true, width=\figwidtfComparisonR]{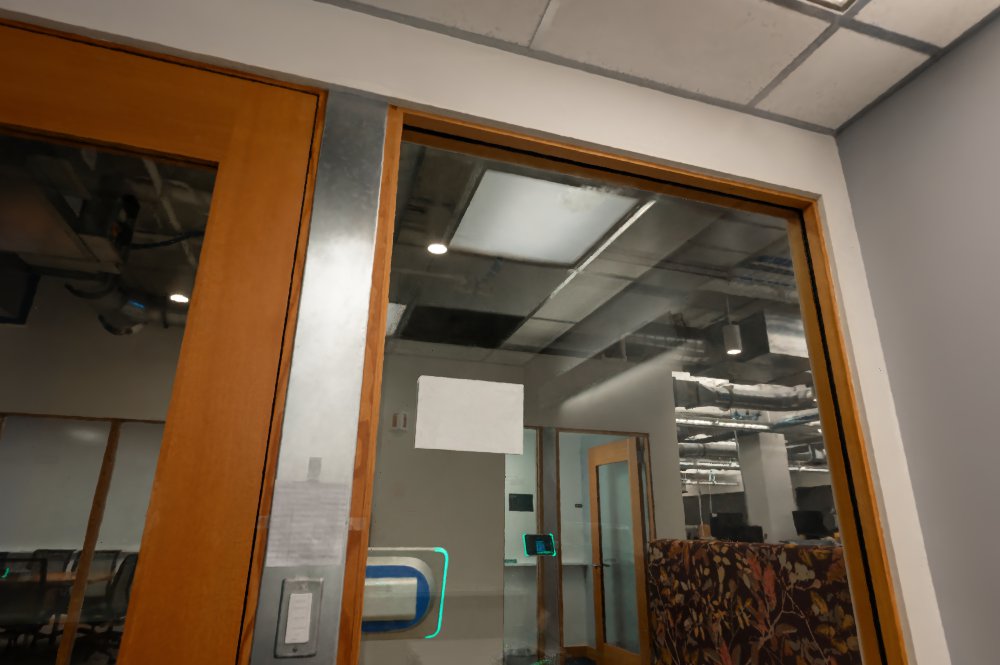}}\\%
  \fbox{\includegraphics[trim=0 0 0 0, clip=true, width=\figwidtfComparisonR]{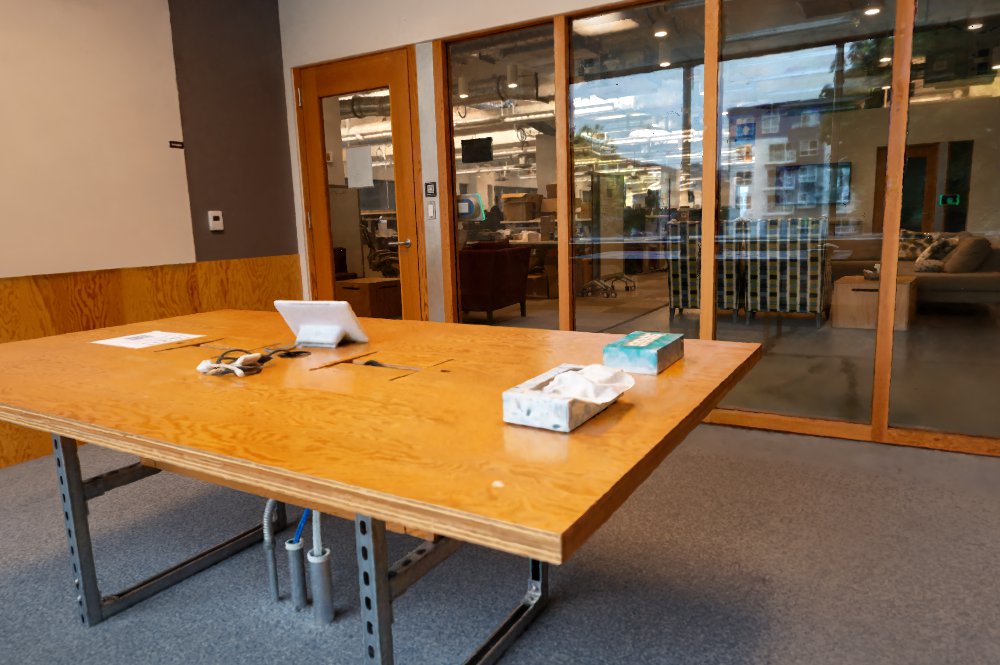}}\\%
  \fbox{\includegraphics[trim=0 0 0 0, clip=true, width=\figwidtfComparisonR]{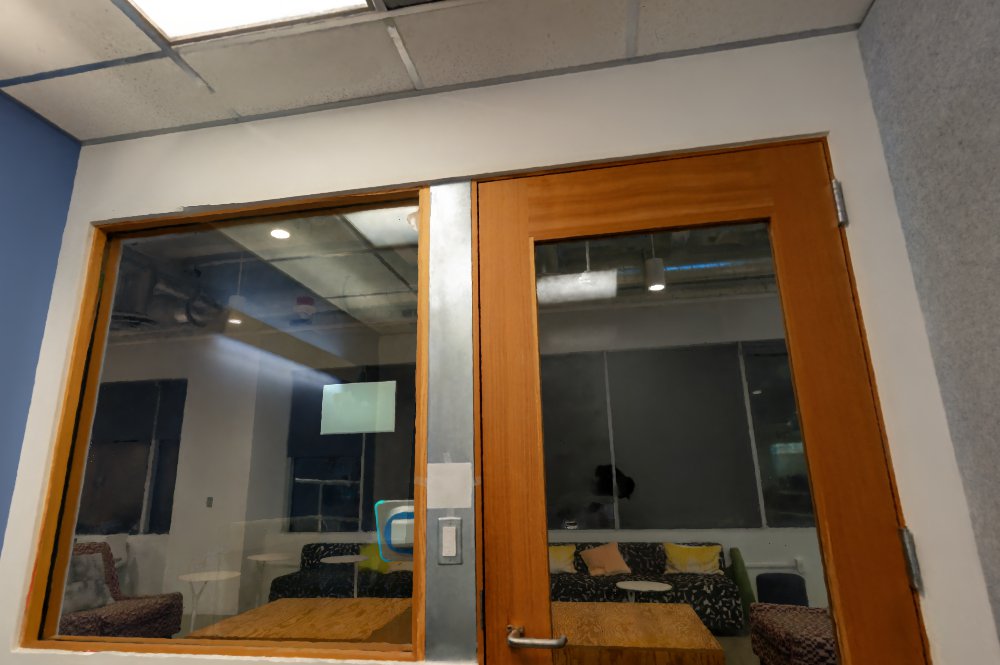}}\\%
  \fbox{\includegraphics[trim=0 0 0 0, clip=true, width=\figwidtfComparisonR]{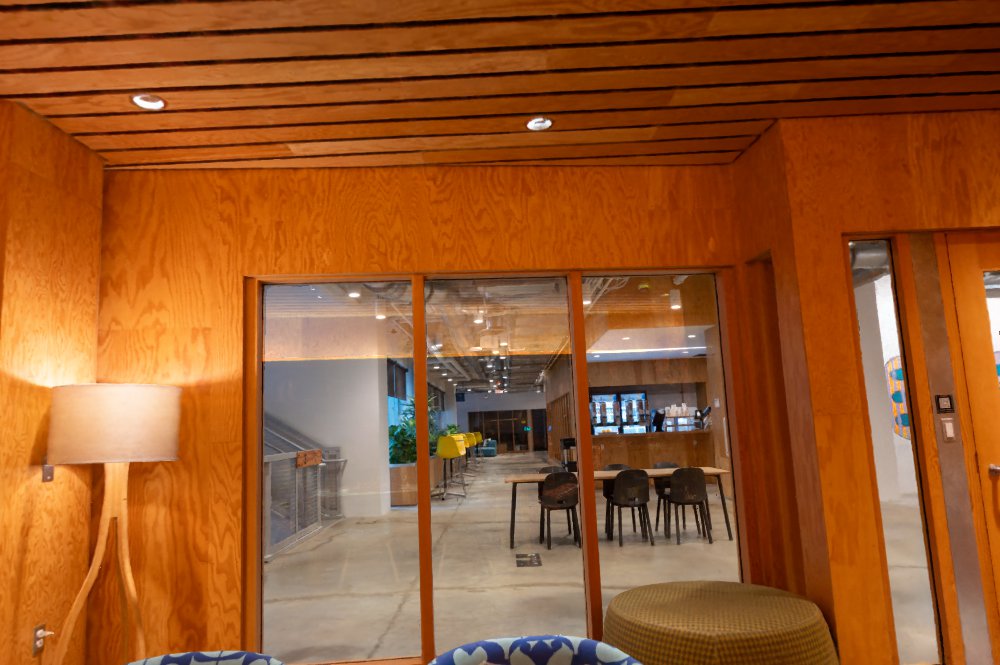}}\\%
  \fbox{\includegraphics[trim=0 0 0 0, clip=true, width=\figwidtfComparisonR]{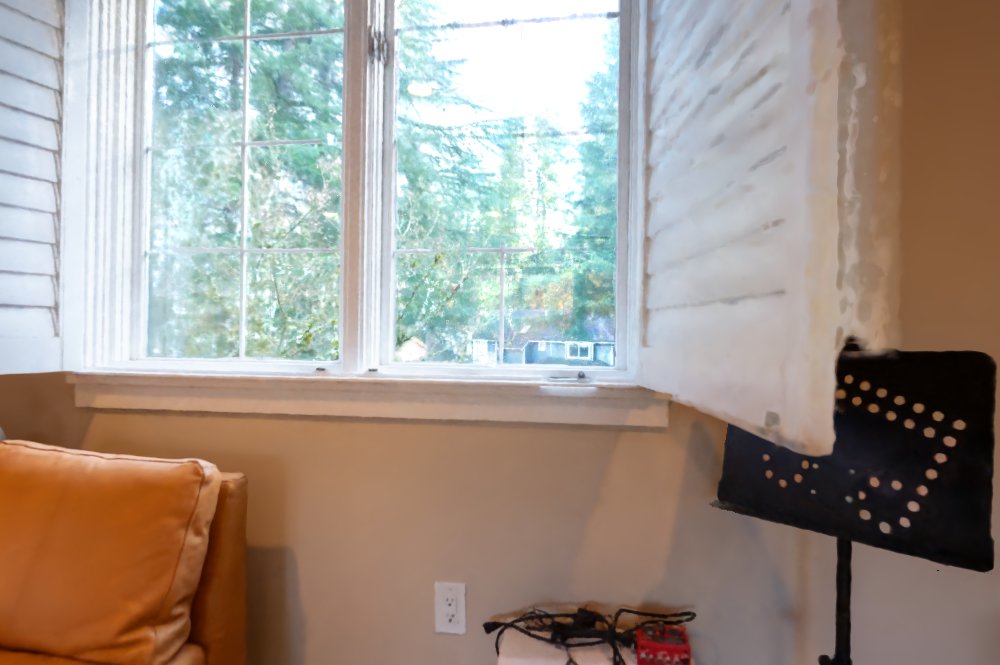}}\\%
   \small Ours}%
\hfill%
\parbox[t]{\figwidtfComparisonR}{\centering%
  \fbox{\includegraphics[trim=0 0 0 0, clip=true, width=\figwidtfComparisonR]{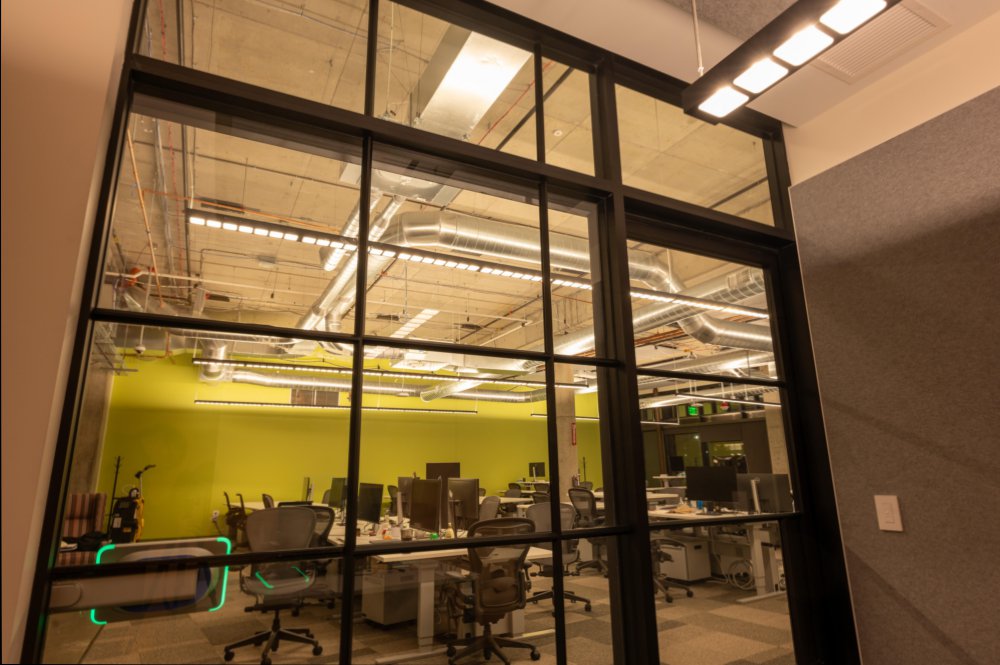}}\\%
  \fbox{\includegraphics[trim=0 0 0 0, clip=true, width=\figwidtfComparisonR]{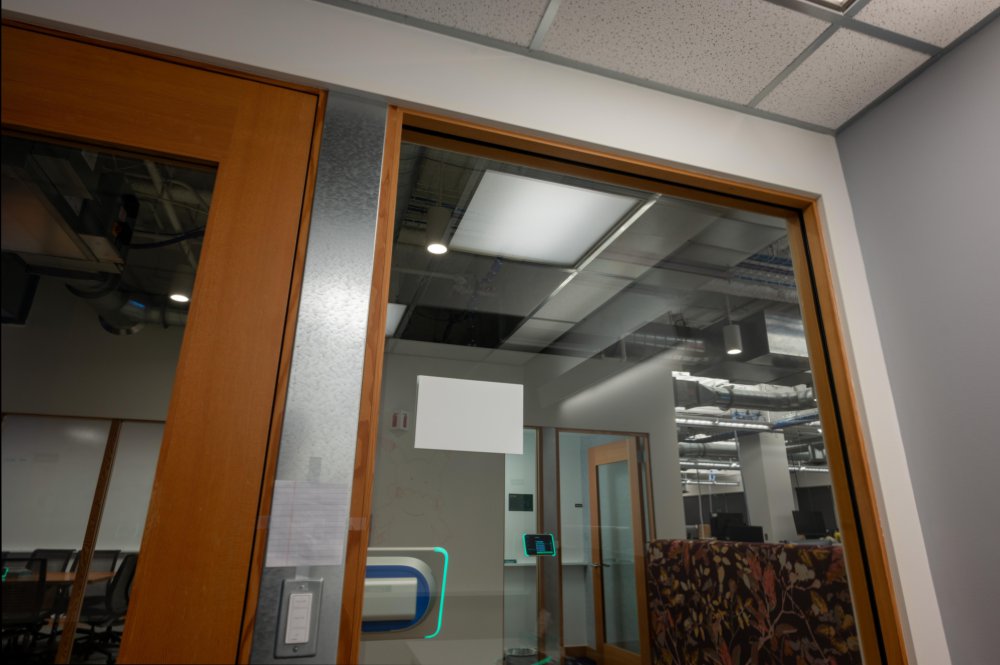}}\\%
  \fbox{\includegraphics[trim=0 0 0 0, clip=true, width=\figwidtfComparisonR]{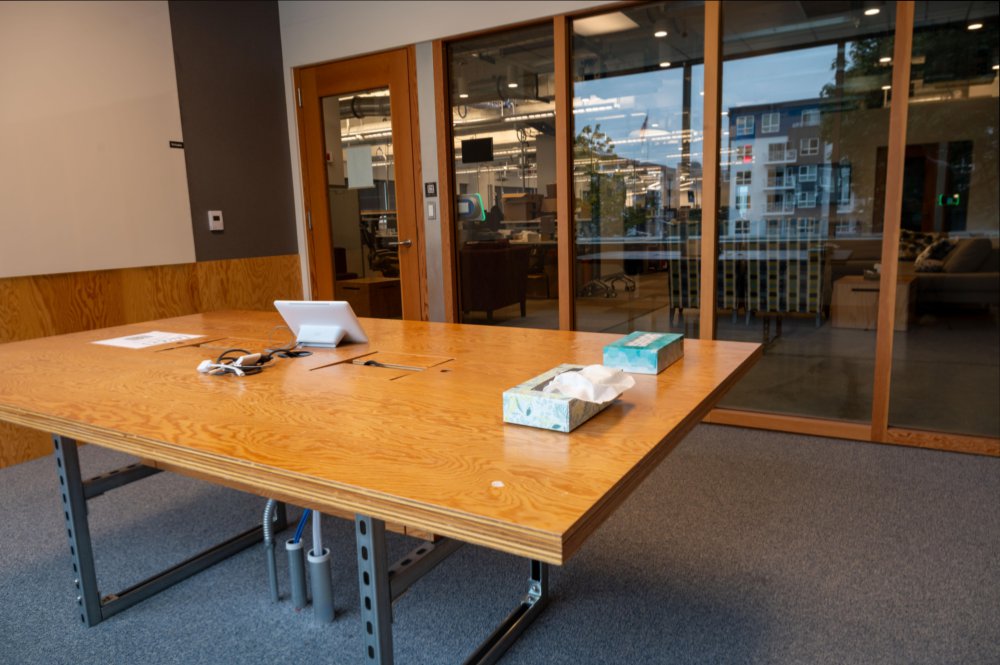}}\\%
  \fbox{\includegraphics[trim=0 0 0 0, clip=true, width=\figwidtfComparisonR]{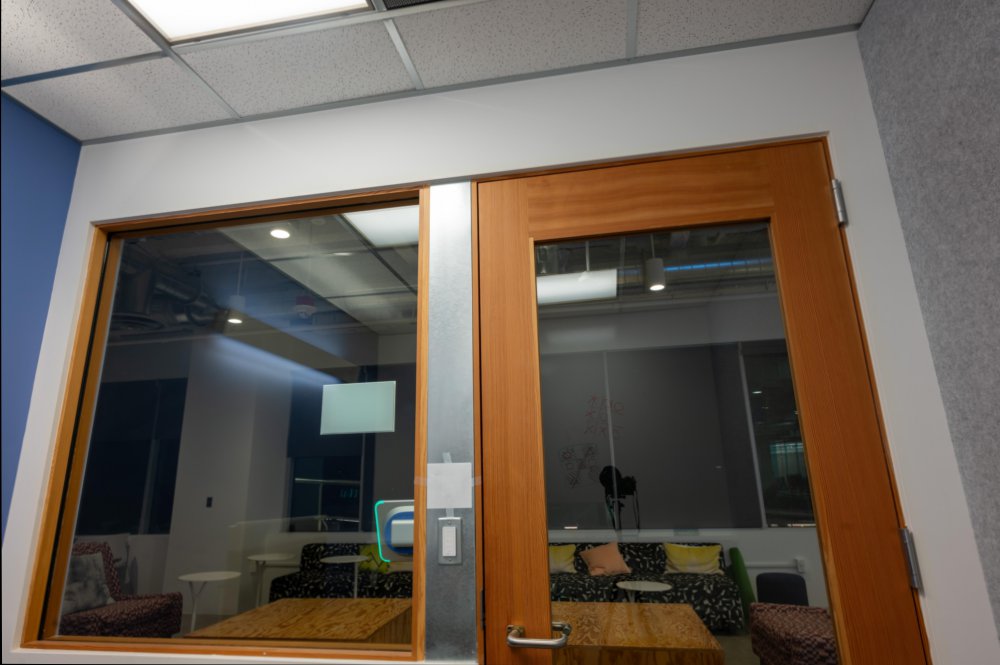}}\\%
  \fbox{\includegraphics[trim=0 0 0 0, clip=true, width=\figwidtfComparisonR]{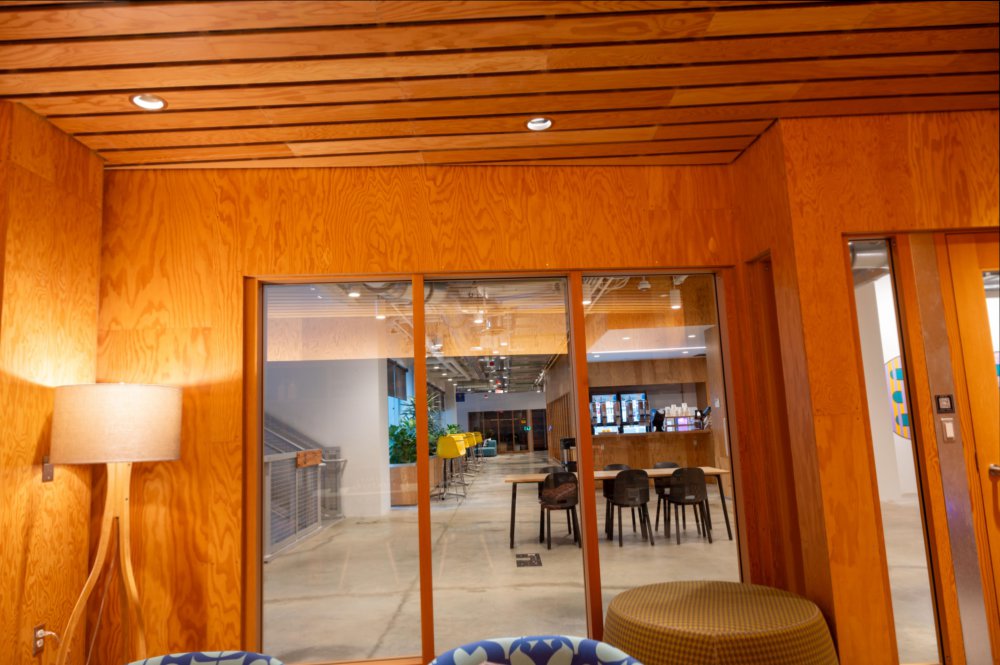}}\\%
  \fbox{\includegraphics[trim=0 0 0 0, clip=true, width=\figwidtfComparisonR]{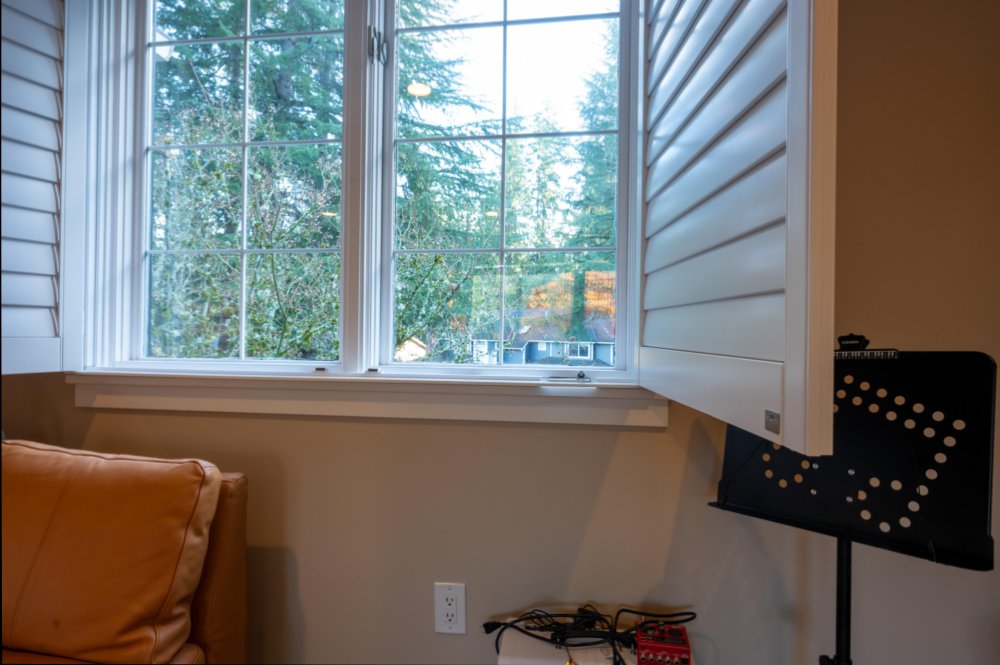}}\\%
   \small GT}%
% left, bottom, right and top
\vspace{\figcapmargin}
\caption{\textbf{Reconstruction comparisons.}
Owing to our method's explicit modeling of reflections, our results exhibit sharper reflections with fewer artifacts compared to other methods.
% \changil{Yes! This should go to Nethe teaser. So, essentially, instead of listing all result images there, you show your decomposition AND all the benefits like better depth, sharper reconstruction, reflection removal example, in comparisons to those of the baseline methods!}
}
\label{fig:reconstruction_comparison}
\end{figure*}

\topic{Improved rendering quality.}
We benchmark the reconstruction quality of our method against several baseline methods.
The comparisons are detailed in \tabref{quantitative_reconstruction} and \figref{reconstruction_comparison}.
Our method, with its explicit modeling of reflections, achieves results with sharper reflections and noticeably fewer artifacts.
Although our results are much sharper and achieve better scores on SSIM and LPIPS, they are darker due to attenuation, which significantly impacts pixel-wise metrics like PSNR.
% Please refer to the supplementary video for comparison.

% \tiny	F-tiny.jpg
% \scriptsize	F-scriptsize.jpg
% \footnotesize	F-footnotesize.jpg
% \small	F-small.jpg
% \normalsize	F-normalsize.jpg
% \large	F-large.jpg
% \Large	F-large2.jpg
% \LARGE	F-large3.jpg
% \huge	F-huge.jpg
% \Huge	F-huge2.jpg

\subsection{Non-transparent Plane}
Our method can handle more than just transparent planes; it can also process opaque surfaces.
We demonstrate our method on an opaque countertop in \figref{countertop}.

\newlength\figwidtfCountertop
\setlength\figwidtfCountertop{0.244\linewidth}
\begin{figure}[t]
\centering%
\parbox[t]{\figwidtfCountertop}{\centering%
  \fbox{\includegraphics[trim=0 0 300 350, clip=true, width=\figwidtfCountertop]{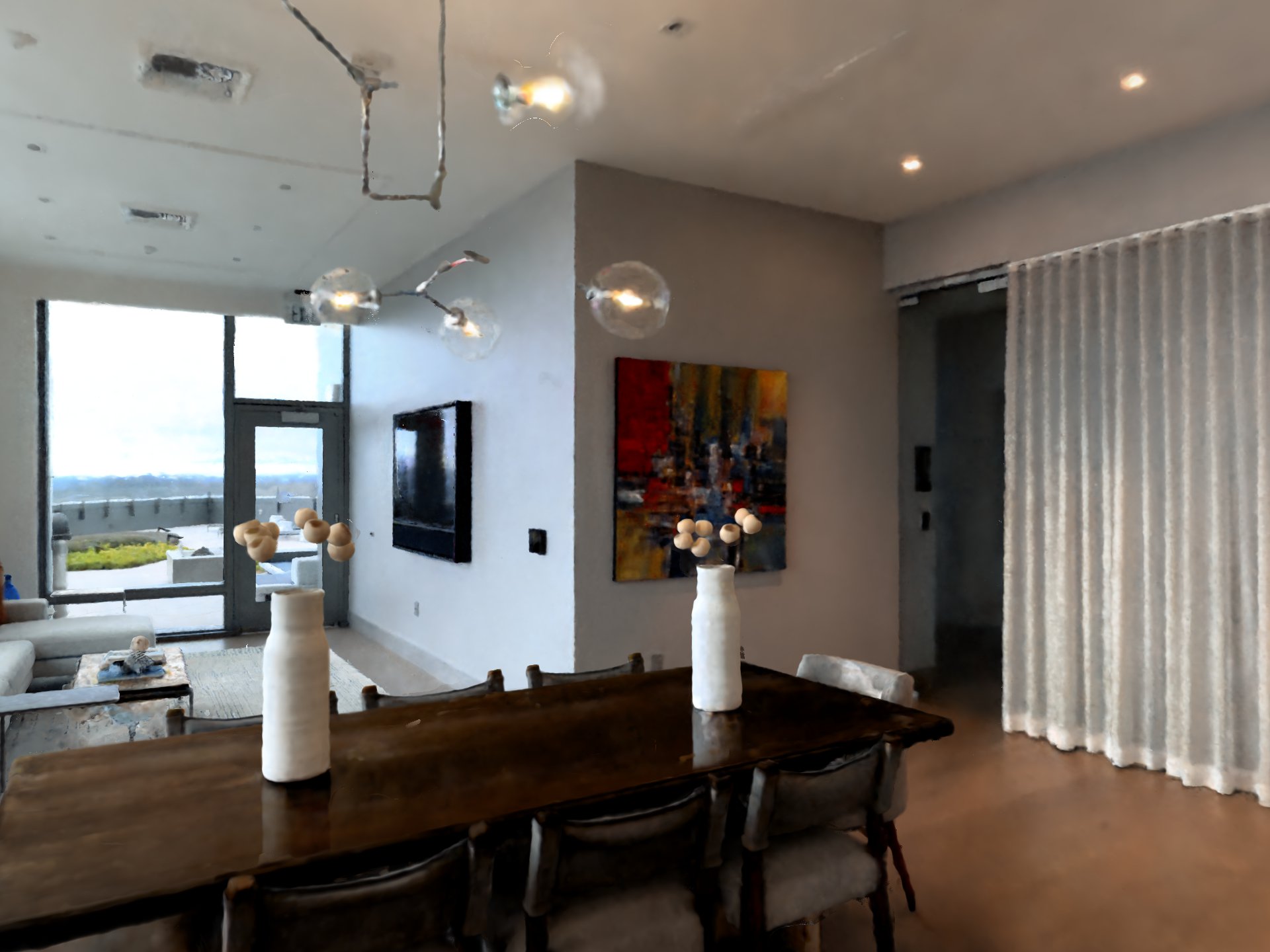}}\\%
   \footnotesize Primary rendering}%
\hfill%
\parbox[t]{\figwidtfCountertop}{\centering%
  \fbox{\includegraphics[trim=0 0 300 350, clip=true, width=\figwidtfCountertop]{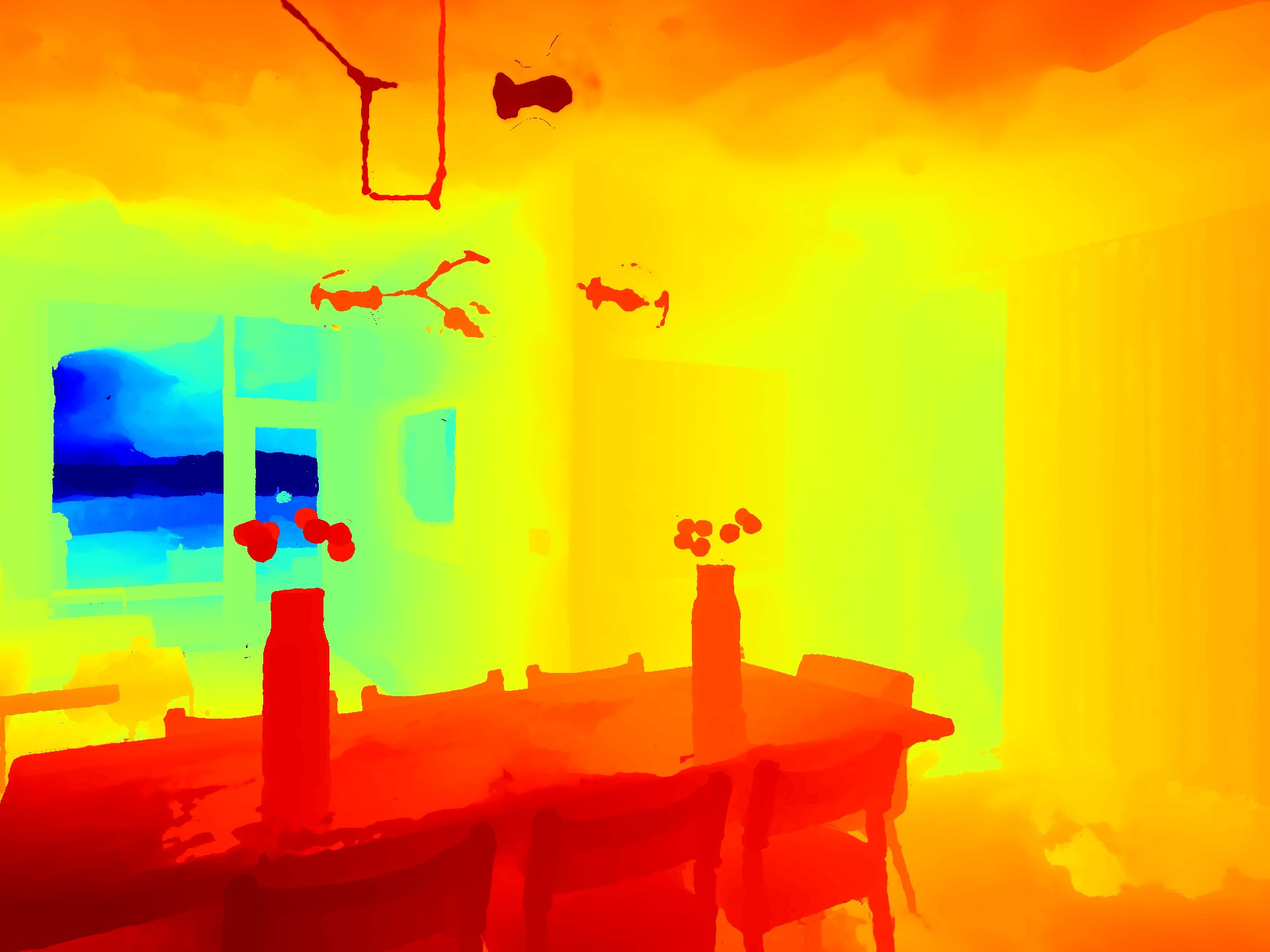}}\\%
   \footnotesize Primary depth}%
\hfill%
\parbox[t]{\figwidtfCountertop}{\centering%
  \fbox{\includegraphics[trim=0 0 300 350, clip=true, width=\figwidtfCountertop]{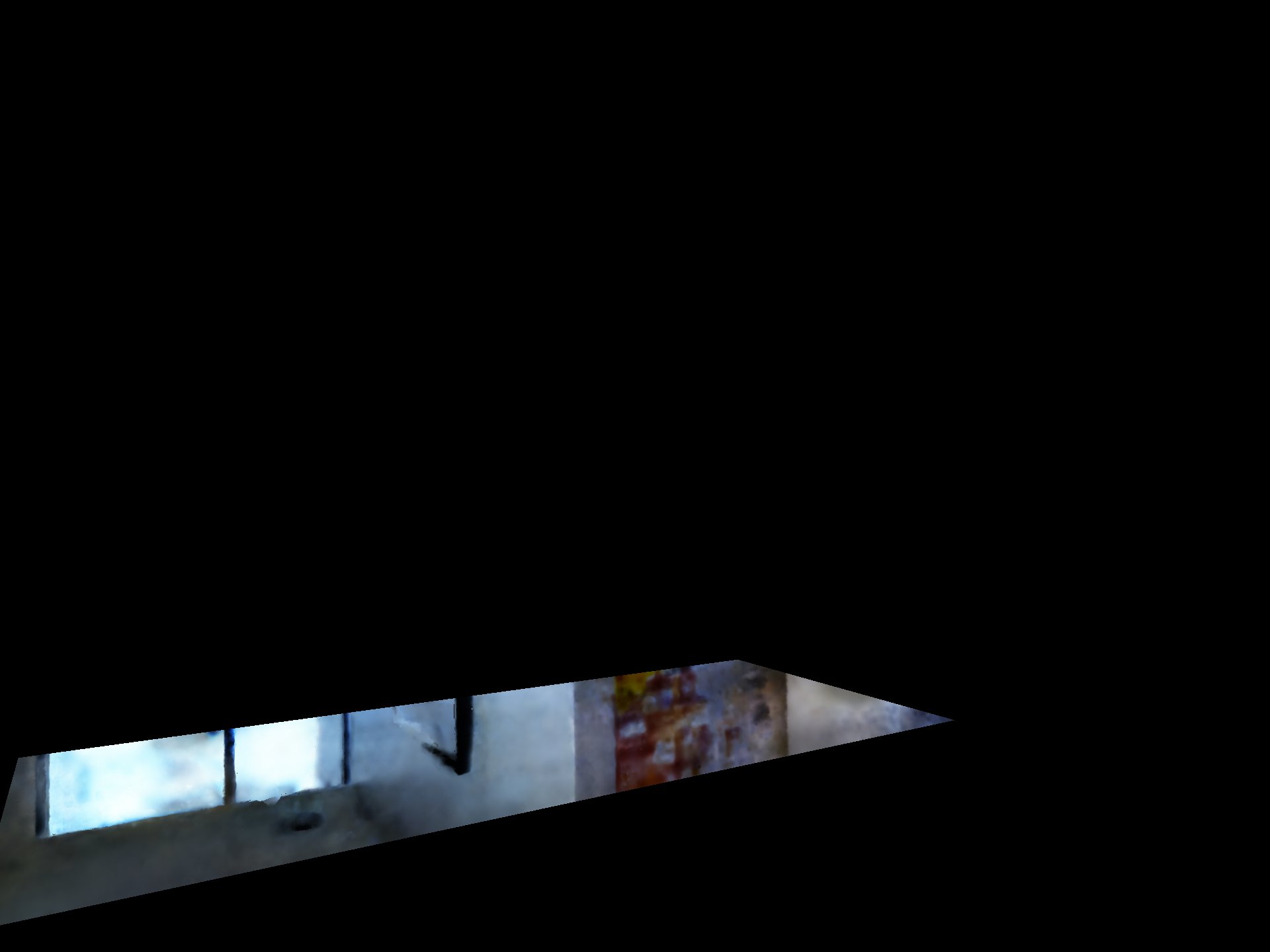}}\\%
   \footnotesize Reflection rendering}%
\hfill%
\parbox[t]{\figwidtfCountertop}{\centering%
  \fbox{\includegraphics[trim=0 0 300 350, clip=true, width=\figwidtfCountertop]{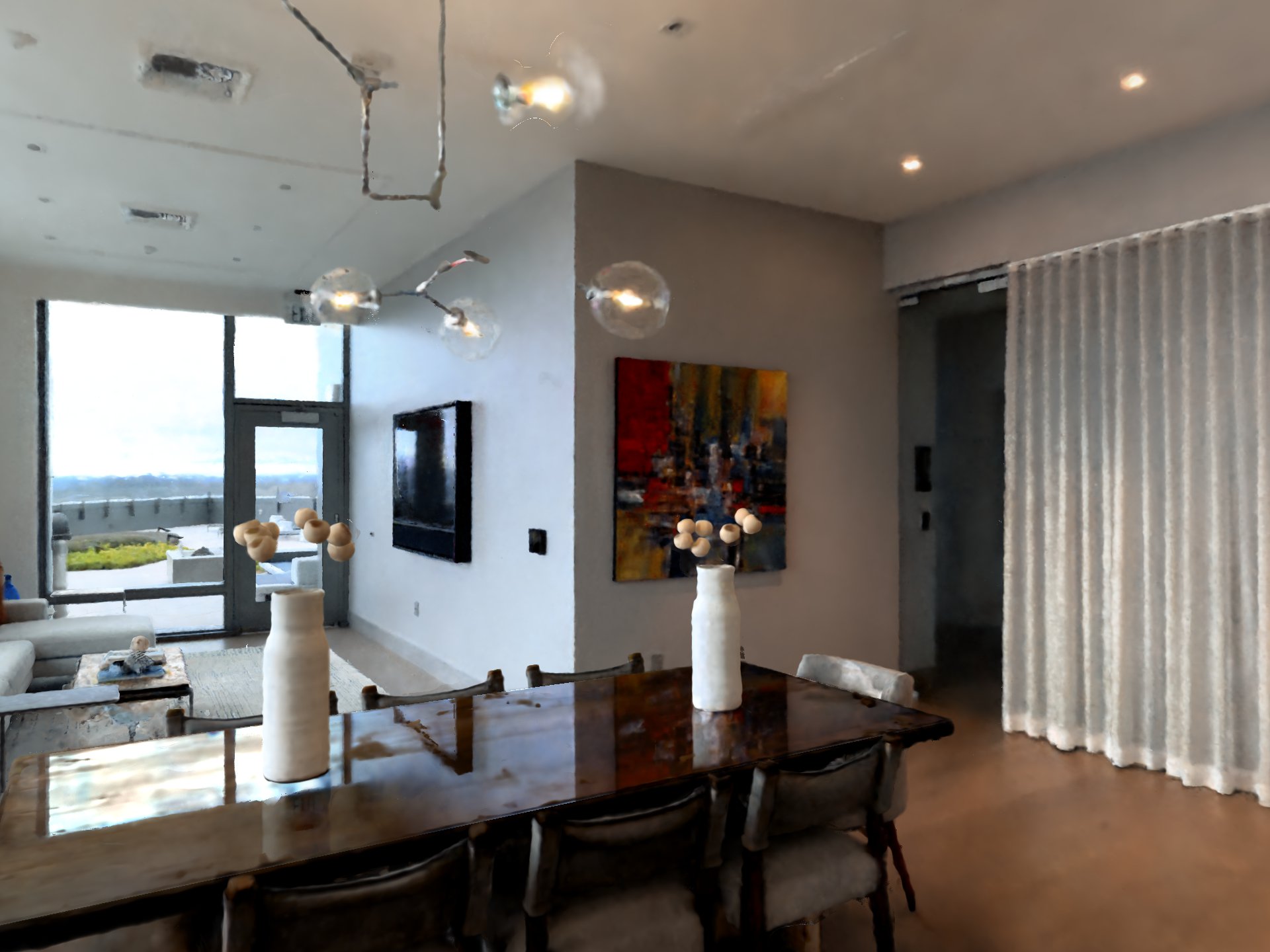}}\\%
   \footnotesize Composed rendering}%
% left, bottom, right and top
\vspace{\figcapmargin}
\caption{\textbf{Non-transparent Plane.}
We showcase our method’s ability to handle non-transparent planes.
}
\label{fig:countertop}
\end{figure}

% \tiny	F-tiny.jpg
% \scriptsize	F-scriptsize.jpg
% \footnotesize	F-footnotesize.jpg
% \small	F-small.jpg
% \normalsize	F-normalsize.jpg
% \large	F-large.jpg
% \Large	F-large2.jpg
% \LARGE	F-large3.jpg
% \huge	F-huge.jpg
% \Huge	F-huge2.jpg

\subsection{Ablation Study}
% We conduct a quantitative analysis to underscore the significance of each key component in our method in \tabref{ablation}.
We quantitatively ablate each key component on challenging held-out \emph{outside} views that deviate significantly from the training views in \tabref{ablation}.
The joint refinement of planes is a critical component of our framework.
As illustrated in \figref{plane_refinement}, there is a noticeable spatial misalignment between the rendered reflection and the observed image prior to refinement.
Joint optimization resolves this issue, achieving better alignment between our rendered reflections and the observed images.
Our sparse edge regularization is critical in preventing the duplication of objects.
Without this regularization, as shown in \figref{effectiveness_edge_loss}, the network tends to introduce false geometry along the primary ray, situated at the same depth as the actual object along the reflected ray.
Our regularization approach effectively mitigates this issue.
Note that the margin appears relatively small because the overall number is lower due to significant view extrapolation. 
The implementation of training scheduling enhances the visual quality of the results. 
This approach fine-tunes the training process, leading to more refined and visually appealing results.

\begin{table}[h]
    % \scriptsize
    \caption{
    \textbf{Ablation study.}
    We report PSNR, SSIM and LPIPS on the Game Room sequence.
    }
    \label{tab:ablation}
    \centering
    \resizebox{0.8\linewidth}{!} 
    {
    \begin{tabular}{l | ccc}
    \toprule
    & PSNR $\uparrow$ & SSIM $\uparrow$ & LPIPS $\downarrow$ \\
    \midrule
    Ours w/o jointly plane refinement        & 19.38 & 0.8123 & 0.138 \\
    Ours w/o training scheduling             & 20.01 & 0.7924 & 0.154 \\
    Ours w/o $\mathcal{L}_\textit{edge}$     & 20.18 & 0.8144 & 0.141 \\
    Ours                                     & \textbf{20.28} & \textbf{0.8216} & \textbf{0.129} \\
    \bottomrule
    \end{tabular}
    }
\end{table}

\subsection{Limitations}
\begin{figure}[t]
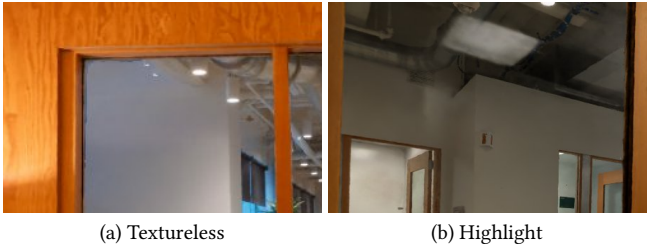

\newlength\figwidthfailure
\setlength\figwidthfailure{0.495\linewidth}
\parbox[t]{\figwidthfailure}{\centering%
  \includegraphics[trim=200 260 550 240, clip=true, width=\figwidthfailure]{fig/results/game_room/rgb_000001.jpg}\\
	\small (a) Textureless}%
\hfill
\parbox[t]{\figwidthfailure}{\centering%
  \includegraphics[trim=100 368 450 0, clip=true, width=\figwidthfailure]{fig/results/meeting_room2/rgb_000000.jpg}\\
	\small (b) Highlight}%
% left, bottom, right and top
\vspace{\figcapmargin}
\caption{\textbf{Failure cases.}
(a) Despite the implementation of sparse edge regularization, our method may be less effective for textureless, edge-blurry reflections.
(b) Inaccuracies in predicted attenuation, especially for highlights, result in erroneous illumination effects.
}
\label{fig:failure}
\end{figure}

In this paper, we focus on glass-material planar reflectors (e.g., windows) because they are ubiquitous in indoor scenes and challenging to model with existing radiance fields.
We assume perfect planar reflection and do not model nonlinear distortion.
We acknowledge that a highly curved surface or an imperfect surface may invalidate our assumptions and lead to degradation in performance. We plan to address these in future work.
In \figref{failure}, we present some instances where our approach did not yield the desired results.
Our method may be less effective in textureless regions, edge-blurry areas, and regions with highlights.

% left, bottom, right and top

% 1000 - 750 = 250
% 665 - 500 = 165
% Despite significant advancements, achieving reflection-free view synthesis continues to be challenging, primarily due to the inherently ill-posed nature of transmittance-reflection separation. 
% This complexity is compounded by the fact that during training, the model is not exposed to any reflection-free data, which could provide clear guidance on differentiating between reflected and non-reflected components.
% In \figref{failure}, we present some instances where our approach did not yield the desired results.
% \input{5_conclusion}
% Bibliography
\bibliographystyle{ACM-Reference-Format}
\bibliography{main}
\clearpage

\end{document}